\documentclass[10pt,journal,compsoc]{IEEEtran}
\newcommand{\final}{1}
\RequirePackage[l2tabu, orthodox]{nag}

\usepackage{times}
\usepackage[LY1]{fontenc}

\usepackage{color}
\usepackage{ifthen}
\usepackage{float}
\usepackage{alltt}
\usepackage{newlfont}
\usepackage{wrapfig}
\usepackage{epsfig}
\usepackage{subcaption}
\usepackage{graphicx}
\usepackage{booktabs}
\usepackage{multirow}
\usepackage{makecell}
\usepackage{array}
\usepackage{comment}
\usepackage{amsmath}
\usepackage{amssymb}
\usepackage{amsthm}
\usepackage{bm}
\usepackage{textcomp}
\usepackage{mathrsfs}
\usepackage{url}
\usepackage{gensymb}
\usepackage[symbol]{footmisc}
\usepackage{placeins}
\usepackage[normalem]{ulem}
\useunder{\uline}{\ul}{}

\usepackage[pagebackref=true,breaklinks=true,colorlinks,bookmarks=false]{hyperref}
\usepackage[capitalise]{cleveref}

\newcommand{\nothing}[1]{}

\definecolor{DeltaColor}{rgb}{0.039,0.73,0.71}
\definecolor{SetaColor}{rgb}{0.867, 0.0235, 0.376}
\definecolor{SigmaColor}{rgb}{0.98,0.45,0.0}
\definecolor{RedColor}{rgb}{0.8,0,0}
\definecolor{AlphaColor}{rgb}{0,0,0.8}
\definecolor{BetaColor}{rgb}{0.8,0,0.8}
\definecolor{GammaColor}{rgb}{0.5,0,0.7}
\definecolor{EpsilonColor}{rgb}{0.353,0.725,0.906}
\definecolor{TauColor}{rgb}{0.423,0.235,0.192}

\definecolor{figred}{rgb}{1,0,0}
\definecolor{figgreen}{rgb}{0,0.6,0}
\definecolor{figblue}{rgb}{0,0,1}
\definecolor{figpink}{rgb}{1,0.63,0.63}

\newcommand{\yuxin}[1]{{\color{SigmaColor} Yuxin: #1 $\qed$}}
\newcommand{\kui}[1]{{\color{GammaColor} Kui: #1 $\qed$}}

\newcommand{\warning}[1]{{\it\color{red} #1}}
\newcommand{\note}[1]{{\it\color{blue} #1}}

\definecolor{AudioColor}{rgb}{0.56,0.34,0.62}

\definecolor{DeadlineColor}{rgb}{0.9,0.4,0} 
\newcommand{\deadline}[1]{{\bf\color{DeadlineColor} ETA: #1}}

\newcommand{\eg}{{\textit{e.g.} }}

\newcommand{\etal}{{\textit{et al.} }}

\newcolumntype{C}[1]{>{\centering}m{#1}}

\ifthenelse{\equal{\final}{1}}
{
\renewcommand{\yuxin}[1]{}
\renewcommand{\kui}[1]{}
\renewcommand{\warning}[1]{}
\renewcommand{\note}[1]{}
\renewcommand{\deadline}[1]{}

}
{}

\newcounter{pccount}
\floatstyle{plain}

\newcommand{\filename}[1]{\url{#1}}
\newcommand{\foldername}[1]{\url{#1}}

\DeclareTextCompositeCommand{\k}{LY1}{a}
  {\oalign{a\crcr\noalign{\kern-.27ex}\hidewidth\char7}}
\DeclareTextCompositeCommand{\k}{LY1}{e}
  {\oalign{e\crcr\noalign{\kern-.27ex}\hidewidth\char7\hidewidth}}
\DeclareTextCompositeCommand{\k}{LY1}{E}
  {\oalign{E\crcr\hidewidth\char7\hidewidth}}

\begin{document}

%
\title{Surface Reconstruction from Point Clouds: A Survey and a Benchmark}

\author{
        Zhangjin~Huang*,
        Yuxin~Wen*,
        Zihao~Wang,
        Jinjuan~Ren,
        and~Kui~Jia
\IEEEcompsocitemizethanks{
\IEEEcompsocthanksitem Z. Huang, Y. Wen, Z. Wang and K. Jia are with the School of Electronic and Information Engineering, South China University of Technology, Guangzhou, China. \protect\\
E-mail: \{eehuangzhangjin, wen.yuxin, eezihaowang\}@mail.scut.edu.cn, kuijia@scut.edu.cn
\IEEEcompsocthanksitem J. Ren is with the University of Macau, Macau, China. \protect\\
E-mails: jinjuanren@um.edu.mo
\IEEEcompsocthanksitem Corresponding author: Kui Jia.}
\thanks{* indicates equal contribution.}
}

\IEEEtitleabstractindextext{
\begin{abstract}
Reconstruction of a continuous surface of two-dimensional manifold from its raw, discrete point cloud observation is a long-standing problem in computer vision and graphics research. The problem is technically ill-posed, and becomes more difficult considering that various sensing imperfections would appear in the point clouds obtained by practical depth scanning.
In literature, a rich set of methods has been proposed, and reviews of existing methods are also provided. However, existing reviews are short of thorough investigations on a common benchmark.
The present paper aims to review and benchmark existing methods in the new era of deep learning surface reconstruction. To this end, we contribute a large-scale benchmarking dataset consisting of both synthetic and real-scanned data; the benchmark includes object- and scene-level surfaces and takes into account various sensing imperfections that are commonly encountered in practical depth scanning. We conduct thorough empirical studies by comparing existing methods on the constructed benchmark, and pay special attention on robustness of existing methods against various scanning imperfections; we also study how different methods generalize in terms of reconstructing complex surface shapes. Our studies help identify the best conditions under which different methods work, and suggest some empirical findings. For example, while deep learning methods are increasingly popular in the research community, our systematic studies suggest that, surprisingly, a few classical methods perform even better in terms of both robustness and generalization; our studies also suggest that the practical challenges of \emph{misalignment of point sets from multi-view scanning}, \emph{missing of surface points}, and \emph{point outliers} remain unsolved by all the existing surface reconstruction methods. We expect that the benchmark and our studies would be valuable both for practitioners and as a guidance for new innovations in future research. We make the benchmark publicly accessible at \url{https://Gorilla-Lab-SCUT.github.io/SurfaceReconstructionBenchmark}.
\end{abstract}
\begin{IEEEkeywords}
Surface reconstruction, surface modeling, point cloud, benchmarking dataset, literature survey, deep learning.
\end{IEEEkeywords}
}

\maketitle
\IEEEdisplaynontitleabstractindextext
\IEEEpeerreviewmaketitle
\ifCLASSOPTIONcompsoc
\IEEEraisesectionheading{\section{Introduction}\label{sec:intro}}
\else

\section{Introduction}
\label{sec:intro}
\fi

\begin{figure*}[htbp]
    \centering
    \includegraphics[width=0.95\textwidth]{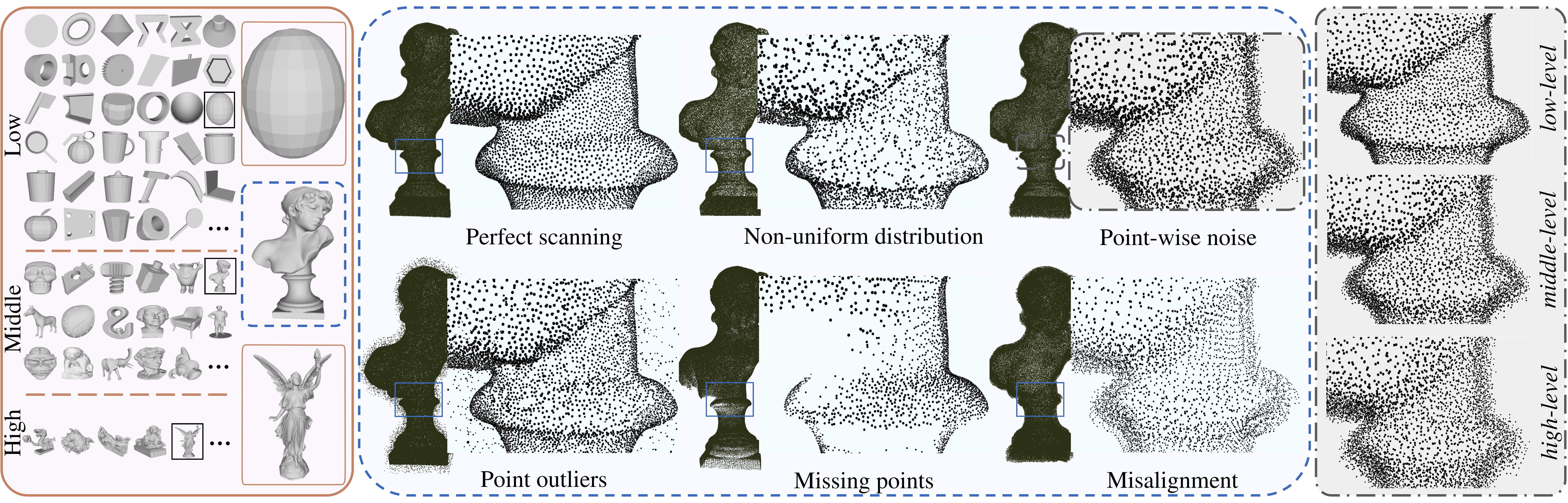}
    \vspace{-0.3cm}
    \caption{An illustration of different surface complexities and scanning challenges included in our contributed benchmark. From left to right: example object surfaces of low, middle, and high complexities, the five challenges possibly encountered in practical surface scanning, and examples of different severity levels for the challenge of noisy scanning.
    }
    \vspace{-0.5cm}
    \label{fig:teaser}
\end{figure*}

\IEEEPARstart{M}{odeling} and reconstruction of object or scene surfaces is a fundamental problem in computer vision and graphics research.
Its applications range from virtual/augmented reality, computer animation, to computer-aided design and robotics.
Given that the mathematical nature of a surface shape is a continuous 2D manifold embedded in the 3D Euclidean space, different approximations are usually adopted when capturing, transmitting, and storing surface shapes, where the prominent examples include point clouds, polygon meshes, and quantized volumes.
In this work, we are particularly interested in reconstructing a continuous surface from its discrete approximation of point cloud, since many depth sensors (e.g., those based on multi-view stereo, structured light, or time-of-flight measurements) produce point clouds (or equivalently, the depth maps) as their original forms of data acquisition, and surface reconstructions from the obtained point clouds are subsequently demanded for various downstream applications.

The problem is technically ill-posed --- infinitely many solutions of the underlying, continuous surface may exist given an observed set of discrete points.
The challenges become even severer considering that various sensing imperfections would appear during the data acquisition process; the captured point clouds could be noisy and distributed in a non-uniform manner, and they could contain outliers and/or cover less on some surface areas; when point clouds are captured at multiple views, they could be subject to less accurate alignments.
All these issues pose the classical problem of surface reconstruction from point clouds as a long-standing challenge that draws continuous efforts from the research community.

In literature, a rich set of methods has been proposed for the focused studies; depending on the types of data imperfection they assume, these methods leverage various priors of surface geometry to combat the otherwise ill-posed problem.

While comprehensive reviews of these methods are given in \cite{bolle1991threedimensionalsurfacereconstructionmethods,lim2014surfacereconstructiontechniquesareview,berger2017survey,surveyonsurfacerec}, these reviews are short of investigations and analyses on a common benchmark that could distinguish existing methods when they cope with the aforementioned data imperfections.
In the meanwhile, the field has witnessed a recent surge of deep learning surface reconstruction, where models of deep networks are learned and employed to either decode surface shapes from point clouds explicitly \cite{atlasnet,luo2021deepdt,pointtrinet}, or generate implicit fields whose zero-level iso-surfaces can be extracted as the results of surface reconstruction \cite{deepsdf,occupancy,LIG}.
It is thus desirable to benchmark both the classical and the more recent, deep learning solutions in order to understand their respective strengths and limitations; such investigations would be no doubt valuable for use of the appropriate methods by practitioners, and also as a guidance to new innovations in future research.

The present paper aims to provide a comprehensive review and benchmark existing methods in the new era of deep learning surface reconstruction.
We organize our review by categorizing existing methods according to what priors of surface geometry they have used to regularize their reconstructions, where we include the more recent priors of deep models and deep learning, in addition to the classical, optimization-based ones.
One of our key contributions is a large-scale benchmarking dataset consisting of both synthetic and real-scanned data (cf. \cref{fig:synthetic_pipeline}, \cref{fig:synthetic_pipeline_scene}, and \cref{fig:scan_obj} for illustrations on how the benchmark is constructed). The benchmark includes object- and scene-level surfaces and takes into account various sensing imperfections that are commonly contained in the point clouds obtained by practical 3D scanning, such as point-wise noise, non-uniform distribution of surface points, point outliers, missing of surface points, and misalignment of point sets from multi-view scanning; \cref{fig:teaser} gives an illustration of these scanning imperfections.
We conduct thorough empirical studies on the constructed benchmark, by comparing existing methods in terms of their capabilities to reconstruct surfaces from observed point clouds. We pay special attention on robustness of existing methods against various scanning imperfections; we also study how different methods generalize in terms of reconstructing complex surface shapes.
Our thorough studies help identify the strengths and limitations of existing methods from multiple perspectives. We summarize a few important findings as follows.
\begin{itemize}
    \item While many challenges of surface reconstruction from point clouds can be more or less tackled by existing methods, those of \emph{misalignment}, \emph{missing points}, and \emph{outliers} have been less addressed and remain unsolved.
    \item Deep learning solutions have shown great promise recently for surface modeling and reconstruction; however, our systematic studies suggest that they struggle in generalizing to reconstruction of complex shapes. It is surprising that some classical methods perform even better in terms of both generalization and robustness.
    \item The use of surface normals is a key to the success of surface reconstruction from raw, observed point clouds, even when the surface normals are estimated less accurately; in many cases, the reconstruction result improves as long as the interior/exterior of the surface can be identified in the 3D space.
    \item There exist inconsistencies between different evaluation metrics, and in many cases, good quantitative results are not always concordant with the visually pleasant ones. This suggests that more foundation studies are demanded to better benchmark different methods and advance the field.
\end{itemize}

\subsection{Related Works}

In this section, we give a summary of existing literatures on surface reconstruction from either point clouds or other observations. We also summarize existing datasets and benchmarks that have served for advancing the field.

\vspace{0.1cm}
\noindent\textbf{Surface Reconstruction from Point Clouds}
There exists a rich set of existing methods studying surface reconstruction from point clouds. These methods are reviewed in \cite{bolle1991threedimensionalsurfacereconstructionmethods,lim2014surfacereconstructiontechniquesareview,berger2017survey,surveyonsurfacerec}. Earlier reviews of \cite{bolle1991threedimensionalsurfacereconstructionmethods} and \cite{lim2014surfacereconstructiontechniquesareview} organize existing methods according to what functions of surface representation they use, e.g., implicit or explicit functions. More recently, existing methods are categorized in \cite{surveyonsurfacerec} based on the difference of used techniques, including interpolation and approximation techniques, learning-based techniques, and soft computing techniques. Our organization of review in the present paper is more similar to that in \cite{berger2017survey}, which also organizes existing methods based on the used geometry priors. Compared with \cite{berger2017survey}, our review is more comprehensive and includes the recent methods of deep learning surface reconstruction. In addition, we also compare existing methods empirically on a common benchmark, which helps identify the respective strengths and limitations of existing methods.

\vspace{0.1cm}
\noindent\textbf{More General Surface Modeling and Reconstruction}
Surface reconstruction can be achieved from other raw observations as well, such as single- or multi-view images, motion, and/or illumination and shading.
Literature reviews on the traditional methods of multi-view image reconstruction are provided in
\cite{slabaugh2001multiviewstereosurvey2}, \cite{remondino2006imagemodelingsurvey}, and \cite{seitz2006multiviewstereosurvey_and_benchmarks}. A more recent survey is given in \cite{han2019imagebasedsurvey} on multi-view image reconstruction with deep learning. Fahim \emph{et al.} \cite{fahim2021singleviewsurvey} focus on a more challenging setting of deep learning surface reconstruction from as few as a single image.
Zhu \emph{et al.} \cite{zhu2017imagemodelingsurvey2} provide a more comprehensive survey of 3D modeling methods, including multi-view 3D reconstruction, structure from motion, and shape from shading.
Different from the above reviews, the present work focuses on surface reconstruction from raw, observed point clouds, considering that depth sensors are increasingly popularly deployed in either portable or fixed devices.

\vspace{0.1cm}
\noindent\textbf{Datasets and Benchmarks}
Existing datasets that support surface reconstruction studies are based on synthetic or real-scanned data; they may include object- and/or scene-level surfaces. For synthetic datasets, surface meshes are usually provided from which point clouds can be sampled. For example, the ShapeNet \cite{shapenet} and ModelNet \cite{modelnet40} are two commonly used synthetic datasets consisting of simple, object-level shapes. More complex synthetic object surfaces are provided in the datasets of 3DNet \cite{3dnetdata}, ABC \cite{abcdata}, Thingi10k \cite{thingi10k}, and Three D Scans \cite{Threedscans}. The datasets of SceneNet \cite{scenenet} and 3D-FRONT \cite{3dfrontdata} provides synthetic, scene-level surfaces. In the meanwhile, there exist datasets of real-scanned, object-level surfaces \cite{bigbird,ycb,kit} and those of real-scanned, scene-level surfaces \cite{replicadata,matterport3d}; however, due to the lack of high-precision scanning, their reconstruction ground truths are usually obtained by appropriate surface reconstruction algorithms, which jeopardizes their roles for benchmarking different methods.
Most of the above datasets do not consider sensing imperfections that may appear in practically scanned point clouds, except for \cite{berger2013benchmark} that uses virtual scanning to simulate point cloud imperfections; however, the dataset \cite{berger2013benchmark} is relatively small, with only eight instances of object surfaces.
In contrast, our contributed benchmarking dataset is more comprehensive, including both synthetic and real-scanned data, and covering both object- and scene-level surfaces; we intentionally inject various sensing imperfections into point cloud data of the dataset, including \emph{point-wise noise}, \emph{non-uniform distribution of surface points}, \emph{point outliers}, \emph{missing of surface points}, and \emph{misalignment among point sets from multi-view scanning}. We expect our benchmark would facilitate more thorough studies in future research.

\subsection{Contributions}

As stated in \cref{sec:intro}, with the surge of deep learning surface reconstruction, the present paper aims to provide a comprehensive review of exiting methods in the new era, and study their respective advantages and disadvantages when reconstructing object- or scene-level surfaces from raw, observed point clouds. To this end, we contribute a large-scale benchmark consisting of both synthetic and real-scanned data. We use the constructed benchmark for systematic studies of existing methods, focusing on the robustness of these methods against various data imperfections, and also on how existing methods generalize in terms of reconstructing complex surface shapes. We summarize our key contributions as follows.
\begin{itemize}
\item We provide a comprehensive review of existing surface reconstruction methods, by bridging together the classical, optimization-based methods with the more recent, deep learning-based ones, where we categorize these methods according to what priors of surface geometry they have used to regularize their solutions.
\item We contribute a large-scale benchmarking dataset consisting of both synthetic and real-scanned data. The point cloud data in the benchmark have various sensing imperfections that are commonly encountered in practical 3D scanning processes; these imperfections are intentionally included to benchmark the robustness of existing methods.
\item We compare existing methods by conducting thorough empirical studies on the constructed benchmark. Our studies help identify the strengths and limitations of existing methods, which are valuable both for choices of appropriate methods by practitioners and for guiding the directions of new innovations in future research.
\end{itemize}

\subsection{Paper Organization}

The paper is organized as follows.
\cref{sec:problem_statement} gives the formal definition of our studied problem. \cref{sec:review} organizes and reviews existing methods based on priors of surface geometry that they have used to regularize the reconstructions. We present our contributed large-scale benchmark in \cref{sec:Challenges}, where we give details about how we construct the benchmark and also the benchmark statistics; we make the benchmark, our construction manner of the benchmark, and also implementation codes of representative methods publicly accessible at \url{https://Gorilla-Lab-SCUT.github.io/SurfaceReconstructionBenchmark}. Experimental setups of our empirical studies are given in \cref{sec:processing}, before results, analyses, and important insights are presented in \cref{sec:experiments}. We finally draw the paper conclusion in \cref{sec:discuss}.

\section{Problem Statement}\label{sec:problem_statement}

Consider a discrete point set $\mathcal{P}$ that may be obtained by scanning an object or scene surface using some 3D sensing devices; each $\bm{p} \in \mathbb{R}^3$ of its contained points collects the coordinates in the Euclidean space.
Our goal of interest is to recover its underlying, continuous surface $\mathcal{S}^{*}$ from which the points $\{ \bm{p} \in \mathcal{P} \}$ are practically observed.
Given that recovering the continuous $\mathcal{S}^{*}$ from the discrete $\mathcal{P}$ is an ill-posed problem, an appropriate regularization must be imposed in order to recover a geometry-aware approximation $\mathcal{S}$, e.g., a smooth and/or fair surface \cite{polygonmeshprocessing}.
This formally amounts to solving the following regularized optimization
\begin{equation}\label{EqnReguObj}
    \min_{\mathcal{S}} L(\mathcal{S}; \mathcal{P}) + \lambda R(\mathcal{S}) ,
\end{equation}
where $L$ is a loss term for data fidelity to the observed $\mathcal{P}$, $R$ is a regularizer that constrains the solution with a certain prior of surface geometry, and $\lambda$ is a scalar penalty.
Note that the objective (\ref{EqnReguObj}) is only in an abstract form, since it is difficult to define both $L$ and $R$ directly on $\mathcal{S}$.
In practice, one may represent a surface either explicitly as a parametric mapping $\bm{f}: \Omega^2 \rightarrow \mathcal{S}$, where $\Omega^2 \subset \mathbb{R}^2$ denotes the 2D domain
and $\mathcal{S} = \{ \bm{f}(\bm{x}) \in \mathbb{R}^3 \}$ with $\bm{x} \in \Omega^2$, or implicitly as the zero-level set of an implicit function $F: \mathbb{R}^3 \rightarrow \mathbb{R}$, i.e., $\mathcal{S} = \{\bm{q} \in \mathbb{R}^3 | F(\bm{q}) =0 \}$.
Correspondingly, one can instantiate the objective (\ref{EqnReguObj}) as the following one that pursues $\mathcal{S}$ by optimizing an explicit mapping
\begin{equation}\label{EqnReguObjExp}
    \min_{\bm{f}  \in \mathcal{H}_{\bm{f}} } L^{\textrm{\tiny exp}}(\bm{f}; \mathcal{P}) + \lambda R^{\textrm{\tiny exp}}(\bm{f}) ,
\end{equation}
where $L^{\textrm{\tiny exp}}$ and $R^{\textrm{\tiny exp}}$ are respectively the instantiated loss function and regularizer, and $\bm{f}$ is optionally constrained in a hypothesis space $\mathcal{H}_{\bm{f}}$ (e.g., by choosing $\bm{f}$ as a neural network).
Alternatively, one may instantiate the objective (\ref{EqnReguObj}) as the following one that pursues an implicit representation of $\mathcal{S}$
\begin{equation}\label{EqnReguObjImp}
    \min_{F \in \mathcal{H}_F} L^{\textrm{\tiny imp}}(F; \mathcal{P}) + \lambda R^{\textrm{\tiny imp}}(F)  \ \mathrm{s.t.} \ F(\bm{q}) = 0 \ \forall \ \bm{q} \in \mathcal{S} ,
\end{equation}
where $L^{\textrm{\tiny imp}}$ and $R^{\textrm{\tiny imp}}$ are again the instantiated functions, and $\mathcal{H}_F$ denotes a hypothesis space that optionally constrains the implicit function $F$.
We present the subsequent sections based on the notations defined in \cref{table:notation}, unless specified otherwise.

\begin{table}[hptb]
    \caption{Math notations.
    }
    \vspace{-0.1in}
    \label{table:notation}
    \centering
    \begin{tabular}{l p{0.70\columnwidth}}
    \hline
    \textbf{Notation} & \textbf{Description} \\
    \hline
    $\mathcal{S}^{*}$ & An underlying surface to be recovered; $\mathcal{S}^{*}\subset\mathbb{R}^3$. \\
    \hline
    $\mathcal{S}$ & A reconstructed surface; $\mathcal{S}\subset\mathbb{R}^3$. \\
    \hline
    $\mathcal{G}_{\mathcal{S}}$ & A triangular mesh representation of surface $\mathcal{S}$. A mesh with $n_{\mathcal{G}}$ faces is collectively written as $\mathcal{G}_{\mathcal{S}}\triangleq \{\mathcal{T}_i\}_{i=1}^{n_{\mathcal{G}}}$, where each face $\mathcal{T}$ is specified by $\{\bm{v}_1,\bm{v}_2,\bm{v}_3\}$ containing three vertices; we also write as $\bm{e}_{ij}$ for the edge connecting vertices $\bm{v}_i$ and $\bm{v}_j$. \\
    \hline
    $\mathcal{P}$ & A set of $n_{\mathcal{P}}$ discrete points $\{\bm{p}_i \in \mathbb{R}^3\}_{i=1}^{n_{\mathcal{P}}}$, representing the practical sampling of an underlying surface $\mathcal{S}^*$. \\
    \hline
    $\bm{n}_{\bm{p}}$ & An estimated, oriented surface normal defined at a surface point $\bm{p}$; $\bm{n}_{\bm{p}} \in \mathbb{R}^3$. \\
    \hline
    $\mathcal{N}(\bm{p})$ & A local neighborhood of points centered at $\bm{p}$. \\
    \hline
    $\Omega$ & A domain of subset space, \eg, $\Omega^k \subset \mathbb{R}^k$. \\
    \hline
    $\mathcal{C}^k$ & The smoothness of a function, where $k$ is the number of continuous derivatives the function has over some domain.  \\
    \hline
    $d(\bm{p})$ & A signed or unsigned distance field value at a point $\bm{p}$. \\
    \hline
    $\bm{f}_{\bm{\theta}}$ or $F_{\bm{\theta}}$ & An explicit or implicit model of surface reconstruction parameterized by $\bm{\theta}$ (e.g., a neural network); the parameters are also denoted as $\bm{\theta}_{\bm{f}}$ or $\bm{\theta}_F$. \\
    \hline
    $\nabla_{x}\bm{f}_{\bm{\theta}}$ or $\nabla_{x}F_{\bm{\theta}}$ & Model derivative with respect to $x$. \\
    \hline
    $L(\mathcal{S}; \mathcal{P})$ & Loss function of a reconstructed surface $\mathcal{S}$ for data fidelity to an observed $\mathcal{P}$. \\
    \hline
    $R(\mathcal{S})$ & Regularizer imposed on a reconstructed surface $\mathcal{S}$. \\
    \hline
    $\bm{K}$ & Extrinsic matrix of a camera. \\
    \hline
    \end{tabular}
\end{table}

\section{Surface Reconstruction with a Categorization of Geometric Priors}\label{sec:review}

Given an observed point set $\mathcal{P}$, a reconstructed surface $\mathcal{S}$ should be close to $\mathcal{P}$ under some distance metric; this is guaranteed by the first term $L(\mathcal{S}; \mathcal{P})$ of data fidelity in the abstract objective (\ref{EqnReguObj}).
\cref{sec:problem_statement} also suggests that $L(\mathcal{S}; \mathcal{P})$ can be instantiated either explicitly or implicitly. The explicit form is generally written as
\begin{equation}\label{eq:L_first_order_ex}
    \begin{split}
        L^{\textrm{\tiny exp}}(\bm{f}; \mathcal{P}) = \frac{1}{n_{\mathcal{P}}} \sum_{\bm{p}\in\mathcal{P}} \min_{\bm{x} \in \Omega^2} \|\bm{f}(\bm{x}) - \bm{p}\|_{\ell} ,
    \end{split}
\end{equation}
where $\| \cdot \|_{\ell}$ denotes a proper norm of distance, with $\ell$ typically set as $1$ or $2$; the above term (\ref{eq:L_first_order_ex}) constrains the learning of mapping function $\bm{f}  \in \mathcal{H}_{\bm{f}}$. In practice, one may sample a fixed set $\{ \bm{x} \in \Omega^2 \}$ instead of optimizing over the whole domain $\Omega^2$, which gives variants of \cref{eq:L_first_order_ex} based on point-set distances, such as Chamfer or Hausdorff distances.
An implicit form of $L(\mathcal{S}; \mathcal{P})$ is generally written as
\begin{equation}\label{eq:L_first_order_im}
    \begin{split}
        L^{\textrm{\tiny imp}}(F; \mathcal{P}) = \frac{1}{n_{\mathcal{P}}} \sum_{\bm{p}\in\mathcal{P}}\| F(\bm{p}) \|_{\ell} ,
    \end{split}
\end{equation}
which learns the implicit function $F \in \mathcal{H}_F$ by minimizing a proper norm of $F(\bm{p})$ for any $\bm{p} \in \mathcal{P}$. Advanced versions of the implicit data fidelity loss exist, e.g.,
\begin{equation}\label{eq:L_higher_order_im}
    \begin{split}
        L^{\textrm{\tiny imp++}}(F; \mathcal{P}) = \alpha_1 \mathbb{E}_{\bm{q} \in \mathbb{R}^3} \| F(\bm{q}) - d(\bm{q}; \mathcal{P}) \|_{\ell_1} + \\
        \alpha_2 \mathbb{E}_{\bm{q} \in \mathbb{R}^3} \| \nabla_{\bm{q}} F(\bm{q}) - \bm{n}(\bm{q}; \mathcal{P}) \|_{\ell_2} + \cdots ,
    \end{split}
\end{equation}
where when $F$ models a Signed Distance Function (SDF) \cite{deepsdf}, $d(\bm{q}; \mathcal{P})$ denotes the signed distance between any space point $\bm{q} \in \mathbb{R}^3$ and the observed point set $\mathcal{P}$, which vanishes when $\bm{q}$ hits any $\bm{p} \in \mathcal{P}$,
and when $F$ models an Occupancy Field (OF) \cite{occupancy}, $d(\bm{q}; \mathcal{P}) \in \{0, 1\}$ depending on whether $\bm{q}$ is inside or outside the surface $\mathcal{S}$, which is practically estimated by comparing $\bm{q}$ with the observed $\mathcal{P}$;
$\bm{n}(\bm{q}; \mathcal{P})$ denotes the normal at point $\bm{q}$, which, when $\bm{q}$ hits some $\bm{p} \in \mathcal{P}$, can be estimated by computing the local tangent plane of $\mathcal{P}$ at $\bm{p}$, and $\| \bm{n}(\bm{q}; \mathcal{P}) \|_2 = 1$ otherwise (when $F$ models an SDF, the second term in \cref{eq:L_higher_order_im} is correspondingly written as $\mathbb{E}_{\bm{q} \in \mathbb{R}^3} | \| \nabla_{\bm{q}} F(\bm{q}) \|_2 - 1 |$); one may use $\{\alpha_1, \alpha_2, \cdots\}$ to weight or switch on/off different terms in \cref{eq:L_higher_order_im}.

Due to the ill-posed nature of surface reconstruction from $\mathcal{P}$, neither of the data fidelity loss terms (\ref{eq:L_first_order_ex}) or (\ref{eq:L_first_order_im}) is sufficient to reconstruct a geometry-plausible $\mathcal{S}$. In literature, various instantiations of the regularization $R(\mathcal{S})$ in \cref{EqnReguObj} have been proposed, in order to make the problem be better posed. In the remainder of this section, we discuss the essence of existing surface reconstruction methods by categorizing their adopted regularization of geometric priors, including \emph{triangulation-based prior}, \emph{smoothness prior}, \emph{template-based prior}, \emph{modeling prior}, \emph{learning-based prior}, and \emph{hybrid prior}, where we include both classical and the recent, deep learning ones.
We expect our categorization and discussion would foster new innovations by bridging classical surface reconstruction methods with the more recent deep learning solutions.

\subsection{Triangulation-based Prior}
\label{SecPriorTriangula}

A closed surface is continuous and could be locally differentiable up to different orders. When the $\mathcal{S}$ to be recovered from $\mathcal{P}$ is locally differentiable up to the first order, i.e., $\bm{f}$ is locally of $\mathcal{C}^1$, a piecewise linear assumption stands as a good prior for modeling the surface, which gives a \emph{mesh} representation of the surface.
Among various meshing schemes, Delaunay triangulation \cite{delaunaytriangulation_edelsbrunner1994triangulating} is a classical one that approximates $\mathcal{S}$ as a mesh that satisfies
\begin{eqnarray}\label{eq:R_triangular}
\mathcal{G}_{\mathcal{S}} = \{ \mathcal{T}_i \}_{i=1}^{n_{\mathcal{G}}}   \qquad\qquad \qquad\qquad \nonumber \\ \mathrm{s.t.} \ \ \ \bm{v}_1^{\mathcal{T}} \in \mathcal{P}, \bm{v}_2^{\mathcal{T}} \in \mathcal{P}, \bm{v}_3^{\mathcal{T}} \in \mathcal{P} \ \forall \ \mathcal{T} \in \mathcal{G}_{\mathcal{S}} , \nonumber \\ \bm{p} \notin \textrm{CC}(\mathcal{T}) \ \forall \ \bm{p} \in \mathcal{P} / \{\bm{v}_1^{\mathcal{T}}, \bm{v}_2^{\mathcal{T}}, \bm{v}_3^{\mathcal{T}}\}  \ \land \ \forall \ \mathcal{T} \in \mathcal{G}_{\mathcal{S}} ,
\end{eqnarray}
where $\mathcal{T}$ is a triangular face with its three vertices denoted as $\{ \bm{v}_1^{\mathcal{T}}, \bm{v}_2^{\mathcal{T}}, \bm{v}_3^{\mathcal{T}}\}$, $n_{\mathcal{G}}$ is the number of faces, and $\textrm{CC}(\mathcal{T})$ denotes the space enclosed by the circumcircle of $\mathcal{T}$ passing through its three vertices; note that by Delaunay triangulation, the explicit data fidelity \cref{eq:L_first_order_ex} is satisfied simultaneously.

Representative methods \cite{greedydelaunay,BPA} of Delaunay triangulation first generate a set of triangular faces directly from the observed $\mathcal{P}$, and then select the optimal subset from them to generate the final triangular mesh.
Greedy Delaunay (GD) \cite{greedydelaunay} proposes greedy algorithm based on topological constraints to select valid triangles sequentially, where the initial triangles are generated by Delaunay triangulation; Ball-Pivoting Algorithm (BPA) \cite{BPA} uses balls with various radii rolling over the points in $\mathcal{P}$ to generate triangles, where every three points touched by a rolling ball will construct a new triangle if the triangle does not encompass any other points, which can also be regarded as an approximation of Delaunay triangulation.

\subsection{Surface Smoothness Priors}
\label{SecPriorSmoothness}

In more general cases, the surface $\mathcal{S}$ to be recovered is expected to be \emph{smooth} or continuously differentiable up to a certain order \cite{polygonmeshprocessing}. Surface smoothness is usually enforced by the following two manners.

Given that the observed points in $\mathcal{P}$ could be noisy, the first manner smoothes out $\{ \bm{p} \in \mathcal{P} \}$ via local weighted combination when fitting the explicit mapping function $\bm{f}$, resulting in an \emph{regularized} version of \cref{eq:L_first_order_ex} as
\begin{equation}\label{eq:L_first_order_ex_regu}
    \begin{split}
        \min_{\bm{f}} L^{\textrm{\tiny exp}}(\bm{f}; \mathcal{P}) = \frac{1}{n_{\mathcal{P}}} \sum_{\bm{p}\in\mathcal{P}} \min_{\bm{x} \in \Omega^2} \|\bm{f}(\bm{x}) - \hat{\bm{p}}(\bm{p})\|_2^2 \\
        \textrm{s.t.} \ \hat{\bm{p}}(\bm{p}) = \sum_{\bm{p}' \in \mathcal{N}(\bm{p})} \bm{p}' / g(\| \bm{p} - \bm{p}' \|_2) , \qquad
    \end{split}
\end{equation}
where $\mathcal{N}(\bm{p})$ denotes a local neighborhood of the observed $\bm{p}$, containing $\{ \bm{p}' \in \mathcal{P} \}$, and $g(\| \bm{p} - \bm{p}' \|_2)$ denotes a function (e.g., a Radial Basis Function or RBF) whose value is proportional to the distance $\| \bm{p} - \bm{p}' \|_2$. A similar implicit objective exists by regularizing the second-order term in \cref{eq:L_higher_order_im}, giving rise to the method of Poisson Surface Reconstruction (PSR) \cite{kazhdan2006poisson}
\begin{equation}\label{eq:psr}
    \begin{split}
        \min_{F} L^{\textrm{\tiny PSR}}(F; \mathcal{P}) = \mathbb{E}_{\bm{q} \in \mathbb{R}^3} \| \nabla_{\bm{q}} F(\bm{q}) - \hat{\bm{n}}(\bm{q}; \mathcal{P}) \|_2^2 \\
        \textrm{s.t.} \ \hat{\bm{n}}(\bm{q}; \mathcal{P}) = \sum_{\bm{p}' \in \mathcal{N}(\bm{q})} \bm{n}(\bm{p}'; \mathcal{P}) / g(\| \bm{q} - \bm{p}' \|_2) ,
    \end{split}
\end{equation}
where $\mathcal{N}(\bm{q})$ denotes a local neighborhood of a space point $\bm{q} \in \mathbb{R}^3$, containing observed points $\{ \bm{p}' \in \mathcal{P} \}$. PSR constrains the normal field only and usually produces over-smooth results. As a remedy, Screened Poisson Surface Reconstruction (SPSR) \cite{kazhdan2013screened} improves over PSR by incorporating regularized versions of both the first- and second-order terms in \cref{eq:L_higher_order_im}.
More recently, Shape As Points \cite{shapeaspoint} develops a differentiable Poisson solver in a spectral manner, which enables an end-to-end optimization and thus could be integrated into the learning of deep neural networks.

The second manner achieves surface smoothness by constraining the function complexities of $\bm{f}$ or $F$. This can be equivalently achieved by constraining the hypothesis space of $\mathcal{H}_{\bm{f}}$ or $\mathcal{H}_{F}$, e.g., by constraining $\mathcal{H}_{\bm{f}}$ as B-spline functions \cite{automaticbspline,triangularbsplines,parsenet} or NURBS \cite{simpleNURBS,integratedNURBS}. Intuitively, a more complex function is able to fit a surface of complex geometry; but it also tends to be overfitted to the observed $\mathcal{P}$, producing a less smooth surface. For example, using RBFs in \cite{CarrRBF} means that the approximate function is in the form of low-degree polynomials with an interpolation of many basic functions centered at the observed points. The works \cite{Fourier,Wavelet} are similar to \cite{CarrRBF} but approximate their respective implicit field functions using different basis functions. More specifically, Kazhdan \cite{Fourier} firstly computes the Fourier coefficients of its implicit field function with the help of Monte-Carlo approximation of Divergence Theorem, and then uses inverse Fourier transform to obtain the implicit function for extraction of iso-surface; a regular grid is needed to perform fast Fourier transform in \cite{Fourier}, and instead Manson \emph{et al.} \cite{Wavelet} use wavelets, which provide a localized, multi-resolution representation of the implicit function.

Methods such as Point Set Surfaces (PSS) \cite{levinmeshindependentmlsbasis1,abc2001pointsetsurfacemls,SPSS,APSS} combine both of the above smoothing strategies. PSS is derived from Moving Least Squares (MLS) \cite{MLS_Smooth_Denoise,RMLSorMLS_Smooth_Denoise2,IMLS,RIMLS}, whose surface can be defined either by an explicit function with stationary projection operator \cite{levinmeshindependentmlsbasis1,abc2001pointsetsurfacemls,MLS_Smooth_Denoise,RMLSorMLS_Smooth_Denoise2,SPSS} or by an implicit function \cite{IMLS,RIMLS,APSS}. In either case, a weighted combination of spatially-varying low-degree polynomials acts as the most important ingredient to locally approximate the observed points and construct the surfaces.
In Simple Point Set Surfaces (SPSS) \cite{SPSS}, the authors iteratively project all the given points along the normal directions onto the local reference planes, which are defined by a weighted average of the points to be projected and their neighborhood points; then the local reference plane at each evaluation point would give a local orthogonal coordinate system to compute a local bivariate polynomial approximation to the surface.
However, the local reference plane can hardly be a good approximation and sometimes even becomes unstable when the observed points are sparse.
Algebraic Point Set Surface (APSS) \cite{APSS} overcomes this issue by using an algebraic sphere to fit the observed points, which forms an implicit function to represent the algebraic distance between the evaluation point and the fitted sphere; the fitting problem is then solved by a least squares problem between the gradient field of the algebraic sphere and the normals of the observed points.
In Robust Implicit MLS (RIMLS) \cite{RIMLS}, the authors combine kernel regression and statistical robustness with MLS to cope with the limitation that MLS can only reconstruct smooth surfaces.
Other PSS methods share the same principle; more details can be found in \cite{mlssurvey} and \cite{berger2017survey}.

\subsection{Template-based Priors}
Template-based priors assume that a surface could be represented by combination of a group of templates, where the templates could be geometric primitives such as spheres or cubes, or complex ones from an auxiliary dataset. As such, surface reconstruction boils down as the problem of estimating and fitting the correct templates to the observed $\mathcal{P}$. This can be formally written as
\begin{equation}\label{eq:TemplatePrimAndModel}
\min_{\{ w\}, \{ \bm{\theta} \} } D\left( \sum_{i=1}^{|\{\mathcal{M}\}|} w_i \mathcal{M}_i(\bm{\theta}_i) , \mathcal{P} \right) ,
\end{equation}
where $\{\mathcal{M}\}$ denotes the set of predefined templates, and each template $\mathcal{M}$ is parameterized by (the possibly learnable) $\bm{\theta}$ and weighted by $w$; $D(\cdot, \cdot)$ denotes a proper distance between the formed surface and the observed $\mathcal{P}$. By minimizing the fitting error (\ref{eq:TemplatePrimAndModel}), a surface is reconstructed by the determined $\{ w\}$ and/or $\{ \bm{\theta} \}$.

\subsubsection{Geometric Primitives}
Templates of geometric primitives are simple shapes that can be analytically represented by a certain number of parameters, \eg, cuboids, spheres, cylinders, cones, etc.
Random sample consensus (RANSAC) \cite{efficientRANSAC,nan2017polyfitplaneprimitive} is the most commonly used method to solve \cref{eq:TemplatePrimAndModel} that fits the right templates to the observed $\mathcal{P}$, where $D(\cdot, \cdot)$ sums up the element-wise (Euclidean) distances between randomly sampled $\{ \bm{p} \in \mathcal{P} \}$ and their projections onto the fitted templates.
More specifically, Schnabel \emph{et al.} \cite{efficientRANSAC} introduce an efficient RANSAC-based algorithm with a novel sampling strategy and an efficient score evaluation scheme, where the primitive with maximum score is extracted iteratively.
Nan and Wonka \cite{nan2017polyfitplaneprimitive} propose to only extract planar primitives based on RANSAC, obtaining the final surface by minimizing a weighted sum of several energy terms to select the optimal set of faces.

\subsubsection{Retrieval-based Templates}

Given an auxiliary set $\{ \mathcal{M} \}$ of shape models, retrieving-and-deforming methods solve \cref{eq:TemplatePrimAndModel} to find the closest shapes and deform them, via optimization of model parameters $\{ \bm{\theta} \}$, to fit the observed $\mathcal{P}$.
For example, in scene reconstruction \cite{dbretrival1,dbretrival2,dbretrival3,dbretrival4}, the observed scene points are segmented into semantic classes, each of which is then fit with a retrieved shape model followed by rigid or non-rigid deformation; in object reconstruction, Pauly \emph{et al.} \cite{dbretrival6_example-based3dscancompletion} warp and blend multiple retrieved object shapes to conform with the observed points, and Shen \emph{et al.} \cite{dbretrival5_partassembly} retrieve individual object parts to form a surface by part assembly.

\subsection{Modeling Priors}

While constraining the complexity of hypothesis space of $\mathcal{H}_{\bm{f}}$ or $\mathcal{H}_{F}$ would promote reconstruction of smoother surfaces, as discussed in \cref{SecPriorSmoothness}, the choice of $\mathcal{H}_{\bm{f}}$ or $\mathcal{H}_{F}$ itself regularizes the reconstruction given that only specific types of surface can be modeled by the choice. A prominent example is the recent trend of using deep neural networks for geometric modeling and surface reconstruction. We term the geometric priors provided by the respective designs of models themselves as \emph{modeling priors}.

Motivated from deep image prior \cite{ulyanov2018deep}, Deep Geometric Prior (DGP) \cite{dgp} verifies the efficacy of deep networks as a prior for geometric surface modeling, \emph{even when the networks are not trained}.
Latter on, Point2Mesh \cite{point2mesh} and SAIL-S3 \cite{SAIL-S3} extend the global modeling adopted in \cite{dgp} as local ones, where the former constructs its local, implicit functions as a weight-shared MeshCNN \cite{meshcnn} and the later method constructs them as a weight-shared Multi-Layer Perceptron (MLP).
Deep Manifold Prior \cite{deepmanifoldprior} delivers mathematical analyses on such modeling properties for MLP as well as convolutional networks.
There have also been a few works making use of modeling priors while not explicitly mentioning it.
Atzmon \emph{et al.} \cite{controllingneurallevelsets} theoretically prove that MLP with Rectified Linear Units (ReLUs) generate piecewise linear surfaces, and a meshing algorithm of Analytic Marching is also proposed in \cite{AnalyticMarching} that is able to analytically compute the piecewise linear surface mesh from such a network, both of which deliver mathematical analyses on how the structure of MLP itself acts as a  regularizer.
Later on, Deep Manifold Prior \cite{deepmanifoldprior} extends such modeling properties for both MLP and convolutional networks.
Implicit Geometric Regularization (IGR) \cite{gropp2020implicitgeometricregularizationforlearningshape} shows that additional regularization on the gradient of neural network would further encourage the generated surfaces to be smooth.
Sign Agnostic Learning (SAL) \cite{sal} and its variants \cite{sald,LightSAL} study reconstructing a surface from an un-oriented point cloud via a specially initialized neural network, and Davies \emph{et al.} \cite{davies2021effectivenessofweight} adopt the same strategy for surface reconstruction from an oriented point cloud.
Neural Splines \cite{neuralspline} performs reconstruction based on random feature kernels arising
from infinitely-wide shallow ReLU networks and shows that such solutions bias toward reconstruction of smooth surface.

\subsection{Learning-based Priors}
\label{sec:learning_prior}

Given the parametrization of $\bm{\theta}_{\bm{f}}$ for $\bm{f} \in \mathcal{H}_{\bm{f}}$ and $\bm{\theta}_F$ for $F \in \mathcal{H}_{F}$, priors can be learned from an auxiliary set of training shapes as \emph{optimized model parameters}. Note that such learning-based priors are different from modeling priors presented in the preceding section, where priors are provided by the models themselves.
Let $\{\mathcal{P}_i, \mathcal{S}_i^{*} \}_{i=1}^N$ be the training set containing $N$ pairs of observed point sets and their corresponding ground-truth surfaces. By adapting the objective (\ref{EqnReguObjExp}), an explicit surface reconstruction based on a learned prior can be generally written as
\begin{align}
    &\min_{\bm{z}} L^{\textrm{\tiny exp}} \left( \bm{f}(\bm{z}; \tilde{\bm{\theta}}_{\bm{f}}); \mathcal{P} \right) + \lambda R^{\textrm{\tiny exp}} (\bm{z}) \qquad\qquad \label{eq:trn_exp_obj} \\
    \textrm{s.t.} \ \tilde{\bm{\theta}}_{\bm{f}}  = \arg&\min_{\bm{\theta}_{\bm{f}}} \frac{1}{N} \sum_{i=1}^N \tilde{L}^{\textrm{\tiny exp}} \left( \bm{\theta}_{\bm{f}}; \{\mathcal{P}_i, \mathcal{S}^{*}_i\} \right) + \tilde{\lambda} \tilde{R}^{\textrm{\tiny exp}}(\bm{\theta}_{\bm{f}}) , \label{eq:trn_exp_constraint}
\end{align}
where the constraint (\ref{eq:trn_exp_constraint}) learns the prior as the optimized model parameter $\tilde{\bm{\theta}}_{\bm{f}}$, $\tilde{L}^{\textrm{\tiny exp}}$ could be different from $L^{\textrm{\tiny exp}}$, and an objective for smooth and/or fair surface may also be incorporated into $\tilde{R}^{\textrm{\tiny exp}}$ in addition to a simple norm constraint of $\bm{\theta}_{\bm{f}}$; given the learned and then fixed $\tilde{\bm{\theta}}_{\bm{f}}$, the objective (\ref{eq:trn_exp_obj}) fits the model prediction to any observed $\mathcal{P}$ by optimizing a latent code $\bm{z}$, where norm of $\bm{z}$ is usually penalized to prevent overfitting. Learning a prior for implicit surface reconstruction can be similarly written as follows, by adapting the objective (\ref{EqnReguObjImp})
\begin{align}
    \min_{\bm{z}} L^{\textrm{\tiny imp}} \left( F(\bm{z}; \tilde{\bm{\theta}}_F); \mathcal{P} \right) + \lambda R^{\textrm{\tiny imp}} (\bm{z}) \qquad\qquad  \label{eq:trn_imp_obj}  \\
    \textrm{s.t.} \ F(\bm{q}, \bm{z}; \tilde{\bm{\theta}}_F) = 0 \ \forall \ \bm{q} \in \mathcal{S} \qquad \ \textrm{and} \qquad \notag \\
    \tilde{\bm{\theta}}_F  = \arg\min_{\bm{\theta}_F} \frac{1}{N} \sum_{i=1}^N \tilde{L}^{\textrm{\tiny imp}} \left( \bm{\theta}_F; \{\mathcal{P}_i, \mathcal{S}^{*}_i\} \right) + \tilde{\lambda} \tilde{R}^{\textrm{\tiny imp}}(\bm{\theta}_F) . \label{eq:trn_imp_constraint}
\end{align}
The model $\bm{f}$ can also be constructed as an auto-encoder architecture \cite{occupancy,convocc,points2surf}, i.e., $\bm{f} = \bm{f}_{\textrm{\tiny encoder}} \circ \bm{f}_{\textrm{\tiny decoder}}$. In such a case, given the prior $\tilde{\bm{\theta}}_{\bm{f}} = \{ \tilde{\bm{\theta}}_{\bm{f}_{\textrm{\tiny encoder}}}, \tilde{\bm{\theta}}_{\bm{f}_{\textrm{\tiny decoder}}} \}$ already learned by \cref{eq:trn_exp_constraint}, a latent code can be directly obtained as
\begin{equation}\label{eq:test_exp_latentcodecomp}
\bm{z} = \bm{f}_{\textrm{\tiny encoder}} (\mathcal{P}; \tilde{\bm{\theta}}_{\bm{f}_{\textrm{\tiny encoder}}}) ,
\end{equation}
instead of optimizing the objective (\ref{eq:trn_exp_obj}). The above applies to an auto-encoder based implicit function $F = F_{\textrm{\tiny encoder}} \circ F_{\textrm{\tiny decoder}}$ as well, and given the learned prior $\tilde{\bm{\theta}}_F = \{ \tilde{\bm{\theta}}_{F_{\textrm{\tiny encoder}}}, \tilde{\bm{\theta}}_{F_{\textrm{\tiny decoder}}} \}$,  one can compute its latent code directly as
\begin{equation}\label{eq:test_imp_latentcodecomp}
\bm{z} = F_{\textrm{\tiny encoder}} (\mathcal{P}; \tilde{\bm{\theta}}_{F_{\textrm{\tiny encoder}}}) .
\end{equation}
Note that the most recent implicit methods \cite{yang2021deepoptprior,tang2021saconvonet} allow the priors of model parameters to be further optimized when fitting to the observed points, i.e., optimizing over both the latent code and model parameters in \cref{eq:trn_exp_obj} or \cref{eq:trn_imp_obj}, and thus potentially bridge the gap between the learning-based priors and classical ones.

Depending on how the training shapes in $\{\mathcal{P}_i, \mathcal{S}_i^{*} \}_{i=1}^N$ are organized, the priors learned via \cref{eq:trn_exp_constraint} or \cref{eq:trn_imp_constraint} may be either at a global, semantic level or as local, shape primitives. For example, when pairs in $\{\mathcal{P}_i, \mathcal{S}_i^{*} \}_{i=1}^N$ capture surfaces of object instances belonging to a semantic category, the learned priors would encode shape patterns common to this object category. To learn priors for reconstruction of arbitrary surface shapes, one may have to prepare $\{\mathcal{P}_i, \mathcal{S}_i^{*} \}_{i=1}^N$ as those encoding surface patches, and expect a global surface of arbitrary shape can be better reconstructed by providing priors on its local shape primitives.

\subsubsection{Learning Semantic Priors}
\label{sec:learning_prior_semantic}

The recent trend of deep learning surface reconstruction starts from learning deep priors at the semantic, object-level. For example, an explicit method of Deep Marching Cubes \cite{deepmarchingcubes} proposes a novel, differentiable layer of marching cubes, which connects mesh surface generation with the learning of semantic prior via a shape encoding network that generates displaced voxel grids.
The seminal deep learning-based implicit methods \cite{occupancy,chen2019learning,deepsdf} model either SDF or OF via deep neural networks.
Specifically, OccNet \cite{occupancy} and IM-Net \cite{chen2019learning} adopt an auto-encoder structure; after training, the encoder generates the latent code representing the shape via a single forward propagation in \cref{eq:test_imp_latentcodecomp} for any observed point set, and the decoder predicts the probability of occupancy according to the given space point along with the latent code.
Different from OccNet and IM-Net, DeepSDF \cite{deepsdf} adopts the structure of decoder-only model, which optimizes the latent code of the given point clouds via maximum a posteriori in \cref{eq:trn_imp_obj}, and predicts signed distances according to the given space point along with the latent code.
Curriculum DeepSDF \cite{curriculumdeepsdf} improves DeepSDF by introducing a progressive learning strategy to learn local details; in the meanwhile, Yao \etal \cite{3dshapelocalgeometry} implement such learning using Graph Neural Networks, which converts the global, semantic latent code into more sophisticated local ones, before feeding into the decoder.
MeshUDF \cite{meshudf} further extends DeepSDF to reconstruct open surfaces, which predicts unsigned distances of any given space point and generates the surface using their customized marching cubes.
More recently, a few methods \cite{airnets,yan2022shapeformer} adopt networks based on transformer \cite{Attention,pointtransformer} to help reconstruction.
Note also that learning semantic priors enables reconstruction of surfaces from raw observations that are originally of no or less 3D shape information (e.g., as few as a single RGB image \cite{atlasnet,SkeletonNet,Junyi1}), by training encoders that learn latent shape spaces from such observations.

\subsubsection{Learning Priors as Local, Shape Primitives}
To reconstruct a scene surface or surface of an object that cannot be semantically categorized, existing methods resort to modeling and learning local priors of shape primitives either at regular grids that partition the 3D space \cite{ifnet,convocc,chibane2020neuralunsigneddistancefield,deeplocalshapelocalsdf,LIG,ssrnet_scalable3dsrn,takikawa2021neuralgeometriclevelofdetail}, or on local patches along the surface manifold \cite{meshletprior,patchnetspatchsdf,points2surf}.
Among the former methods, Implicit Feature Networks (IF-Net) \cite{ifnet} and Convolutional OccNet \cite{convocc} (ConvOccNet) obtain latent codes for local grids via auto-encoder structure, where IF-Net \cite{ifnet} adopts 3D convolution that convolves each input point with its surrounding points to get the latent code at each local grid, and ConvOccNet \cite{convocc} uses PointNet \cite{pointnet} as its encoder to get the latent code for each point and then encapsulates all the latent codes into a volumetric feature via average pooling.
Later on, Chibane \emph{et al.} \cite{chibane2020neuralunsigneddistancefield} extend IF-Net \cite{ifnet} to support sign-agnostic learning.
Deep Local Shape (DeepLS) \cite{deeplocalshapelocalsdf} divides the whole 3D space into regular voxels, and trains an implicit function whose parameters are shared among different local voxels.
Local Implicit Grid (LIG) \cite{LIG} trains an auto-encoder to extract the latent codes of local voxels during training (via \cref{eq:test_imp_latentcodecomp}), while retaining the voxel-shared decoder with fixed parameters during inference only.
To improve the efficiency of local encoding, Scalable Surface Reconstruction Network (SSRNet) \cite{ssrnet_scalable3dsrn} makes use of octree to partition the whole 3D space, and uses fully-convolutional U-shaped network with skip connections, which is based on modified tangent convolution \cite{tangentconv3d} to get the latent code for each local grid.
Neural Geometric Level of Detail (LOD) \cite{takikawa2021neuralgeometriclevelofdetail} adopts the structure of sparse octree to further improve the efficiency, where latent codes are computed only for local voxels that intersect with the surface.
Ummenhofer \etal \cite{adamulconvkernel} aggregate latent codes for local grids via adaptive grid convolution on multiple levels of the octree input space.
As for the methods based on local surface patches, Badki \emph{et al.} \cite{meshletprior} introduce the concept of meshlets, and train a variational auto-encoder to learn the latent space of pose-disentangled meshlets; by back-propagating the error with respect to the given points and the meshlets, the method updates the meshlets' latent codes and deforms the meshlets to fit the given points at inference time.
PatchNets \cite{patchnetspatchsdf} leverages the structure of implicit auto-decoder to learn across different local surface patches, and the decoder is also trained with an elaborate loss function to ensure the smoothness of the reconstructed patches.
Points2Surf \cite{points2surf} learns features from both local patches and the global surface, and reconstructs the surface with an implicit decoder, where the former takes a local encoder to learn the absolute distance of a queried point from the local surfaces, and the latter learns the interior/exterior of the surface with a global encoder.
POCO \cite{POCO} and \cite{Neighborbased} encodes each input point into a single latent code, and then perform weighted interpolation among a point and its neighboring ones to get the local latent code.

\subsection{Hybrid Priors}

Methods discussed in the preceding sections are organized according to their respectively used, main priors of surface geometry.
To improve the plausibility of reconstructed surfaces, existing methods usually combine multiple priors, e.g., by combining smoothness priors with triangulation or template-based ones \cite{delaunay_tri_smooth1,delaunay_tri_smooth2,delaunay_primitive_smooth}, by imposing additional priors on top of modeling ones \cite{paulsen2009markov,scholkopf2004kernel,williams2021nkf,NeuralPull,phasetrans,NeuralIMLS,splineposenc}, or by learning priors to improve over regularization provided by pre-defined ones \cite{luo2021deepdt,pointtrinet,learningdelaunaysurface,liu2021DeepIMLS,SPFN,parsenet}.
In this section, we focus our discussion on the last case, considering that such hybrid priors are popularly used in the new era of deep learning surface reconstruction.

Given auxiliary sets of training shapes, learning-based priors are combined with triangulation-based prior in \cite{pointtrinet,instrinsic-extrinsic,learningdelaunaysurface} that encourage deep networks to learn particular properties for triangulation from the observed point clouds.
More specifically, PointTriNet \cite{pointtrinet} introduces a framework to generate triangles from observed point clouds directly, where a proposal network suggests candidates of triangles and a classification network predicts whether a proposed candidate should appear in the reconstructed surface or not, and two networks iteratively take effects until the final surface of triangular mesh is generated.
Liu \emph{et al.} \cite{instrinsic-extrinsic} show that connecting those pairs of vertices whose geodesic distance approaches to their Euclidean distance can approximately reconstruct the underlying surface; they construct a network to select candidate triangles satisfying this property.
Delaunay Surface Elements (DSE) \cite{learningdelaunaysurface} introduces small triangulated patches generated by combining Delaunay triangulation with learned logarithmic maps, from which candidate triangles of the output surface are then iteratively selected via their proposed adaptive voting algorithm.
DeepDT \cite{luo2021deepdt} learns deep networks to extract geometric features from observed point clouds, and then integrates the learned features together with graph structural information to vote for the inside/outside labels of Delaunay tetrahedrons; the surface is reconstructed by extracting triangular faces between tetrahedrons of different labels. Gao \etal \cite{gao2020deftet} learn a network to predict the offsets of vertices of tetrahedrons and a second network to predict the occupancy of each tetrahedron, and object surface is generated on the facet that belongs to two tetrahedrons with different occupancies.

Learning-based priors are commonly combined with smoothness priors (e.g., the Laplacian regularization in \cite{wang2018pixel2mesh} and the cosine smoothness in \cite{kato2018neuralrender}).
IMLSNet \cite{liu2021DeepIMLS} trains an implicit network that evaluates signed distances at grids of an octree-based 3D space, where smoothness is regularized by defining the signed distances in a way similar to implicit moving least-squares \cite{RIMLS}.
Xiao \etal \cite{LMIRecon2021} learn a network to implicitly model an indicator function derived from Gauss lemma, which is smooth and linearly approximates the surface; similar to Points2Surf \cite{points2surf}, they learn both local and global features.

There have also been plenty of works \cite{SPFN,parsenet,genova2020local} combining learning-based priors with template-based ones.
Li \emph{et al.} \cite{SPFN} propose to train a deep network to fit geometric primitives according to the observed point clouds; the trained network predicts point-wise features that are fed into a differentiable estimator for algebraic computation of primitive parameters.
ParseNet \cite{parsenet} uses a neural decomposition module to partition an observed point cloud into multiple subsets, each of which is assumed to be a primitive type modeled as an open or close B-spline by a deep network.
Local Deep Implicit Function (LDIF) \cite{genova2020local} represents a surface shape as a set of shape elements, each of which is parameterized by analytic shape variables and latent shape codes; the method learns such variables and latent codes by its SIF encoder and PointNet encoder respectively.
For reconstruction of a large-scale scene surface, RetrievalFuse \cite{siddiqui2021retrievalfuse} retrieves a set of object templates from an auxiliary scene dataset, and learns a network with attention-based refinement to produce the reconstruction.

\section{A Surface Reconstruction Benchmark}
\label{sec:Challenges}

As discussed in Section \ref{sec:review}, a rich set of methods exist that aim to address the ill-posed problem of surface reconstruction from point observations. These methods have their respective merits, yet it is less clear on their advatanges/disandvatanges under different working conditions, due to the lack of a comprehensive surface reconstruction benchmark that identifies and includes the main challenges faced by the studied problem.
In this work, we contribute such a benchmark by both \emph{synthesizing} and \emph{practically scanning} point clouds of object and scene surfaces. We identify the main challenges of surface reconstruction from point clouds obtained by imperfect surface scanning, including \emph{point-wise noise}, \emph{point outliers}, \emph{non-uniform distribution} of points, \emph{misalignment} among point sets obtained by scanning different but overlapped, partial surfaces of an object or scene, and \emph{missing points} of one or several surface patches. An illustration of such challenges is given in \cref{fig:teaser}. We include all the five challenges in synthetic data of the benchmark, and expect that an arbitrary combination of these challenges may appear in any sample of the real-scanned data. The benchmark is organized as the synthetic data of object surfaces, the synthetic data of scene surfaces, and the data of real-scanned surfaces. Tables \ref{table:statistics_summary} and \ref{table:statistics_summary_real} summarize the statistics.
We make the benchmark publicly accessible at \url{https://Gorilla-Lab-SCUT.github.io/SurfaceReconstructionBenchmark}, where we also release the code implementation of our synthetic scanning pipeline to facilitate future research in the community.

\begin{figure*}[hptb]
    \centering
    \includegraphics[width=0.95\textwidth]{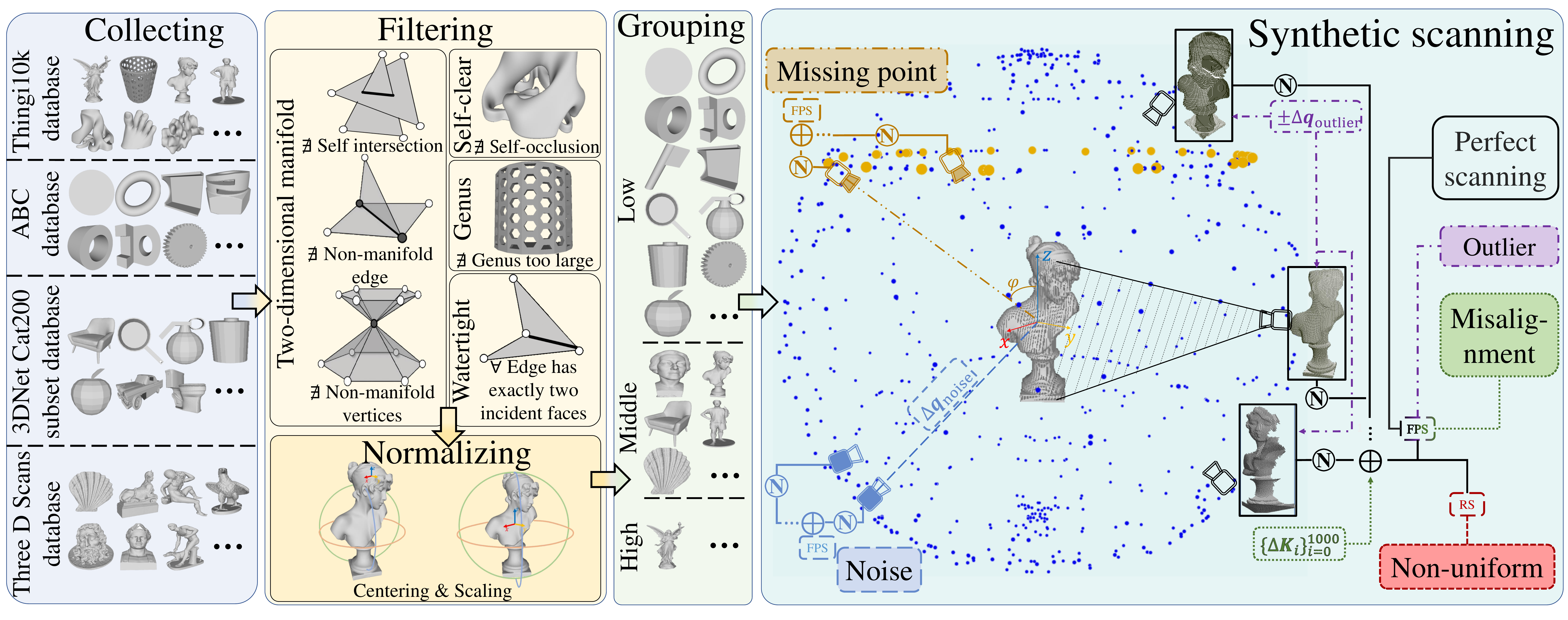}
    \vspace{-0.4cm}
    \caption{
    The pipeline of constructing our synthetic object-level dataset (cf. Section \ref{sec:synthetic_data} for the details).
    We firstly collect CAD models of object surface from four repositories, as shown in the leftmost of the figure. We then filter the collected surfaces by dropping those failing to meet our requirements, and normalize the remaining ones, as shown in the middle of the figure.
    We further organize the normalized object surfaces into three groups of low, middle, and high complexities at a ratio of around $6:3:1$, based on the criterion of (algebraic) surface complexity.
    We finally perform synthetic scanning as shown in the right of the figure, where different challenges possibly encountered in practical scanning are simulated. Blue spots indicate the scanning positions on viewing spheres; after estimating oriented normals (shown as N in the figure), we make registration between different viewpoints of scanning and get the final scanned point cloud, where farthest point sampling (FPS) is used for sampling a fixed number of points per point cloud (except otherwise mentioned cases in which random sampling (RS) is used).
    }
    \label{fig:synthetic_pipeline}
    \vspace{-0.5cm}
\end{figure*}

\subsection{The Synthetic Data of Object Surfaces}\label{sec:synthetic_data}

To prepare synthetic data of object instances in the benchmark, we collect CAD models from existing repositories \cite{thingi10k,3dnetdata,abcdata,Threedscans}, and pre-process them before synthetically scanning their point clouds.
For each CAD model, we simulate the aforementioned five ways of imperfect scanning, in addition to a perfect scanning that gives a clean object point cloud; we in total have six kinds of scanned point clouds for each instance.
We organize the collected object instances into three levels of (algebraic) surface complexity \cite{shapecomplexity}, in order to study how different methods perform on surface reconstruction of varying complexities.

\subsubsection{Data Collection}\label{sec:collect_preprocess}

We collect CAD models of object instances from the existing repositories of Thingi10k \cite{thingi10k}, ABC \cite{abcdata}, 3DNet Cat200 subset \cite{3dnetdata}, and Three D Scans \cite{Threedscans}.
More specifically, we randomly select $3,000$
objects from Thingi10k \cite{thingi10k} (a dataset collected from online-shared 3D printing models),  $3,000$ industrial components from the set of Chunk 0080 in ABC \cite{abcdata}, $3,400$ commodities from 3DNet Cat200 subset \cite{3dnetdata}, and $106$ art sculptures from Three D Scans \cite{Threedscans}. All these instances are represented in the form of triangular mesh.
As illustrated in Fig. \ref{fig:synthetic_pipeline}, some meshes of the collected instances could be non-watertight, with self-occlusion, and/or topologically too complex, and are thus less convenient to be synthetically scanned to simulate practical sensing conditions; we filter out these instances by determining their states of being watertight and 2D-manifold \cite{watertight2manifold}, checking self-occlusion \cite{ambientocc}, and calculating their surface genus \cite{polygonmeshprocessing}.
We finally normalize each mesh of the remaining instances by centering it at the origin and scaling it isotropically to fit into the unit sphere.
We obtain a total of $1,620$ object surface meshes in the benchmark.

\subsubsection{Groups of Varying Surface Complexities}\label{sec:hierarchical}

It is possible that performance of different methods depends on the complexities of object surfaces to be reconstructed. To identify better working conditions for different methods, we intend to divide the above processed object surface meshes into groups of varying complexities.
Surface complexities can be measured under different metrics, among which algebraic complexity and topological complexity are the measures more relevant to surface reconstruction from point clouds \cite{shapecomplexity}; simply put, the former measures the degree of polynomials needed to represent a surface, and the latter can be measured as the surface genus (e.g., the number of holes on the surface).
Given that our instances of surface meshes have been processed to satisfy the conditions of being watertight, 2D manifold, and topologically simpler (equal to or smaller than genus 5), we divide all the $1,620$ instances into three groups of \emph{low-}, \emph{middle-}, and \emph{high-complexity} based on the measure of algebraic complexity.

The algebraic complexity of a surface is usually computed as the highest degree of polynomial functions that approximate/fit local patches on the surface \cite{shapecomplexity}.
However, given fixed budget of approximation errors, it is usually unstable to fit local surface patches with polynomials of high degrees \cite{MLS_Smooth_Denoise}, causing inaccurate prediction of polynomial degrees and thus that of algebraic complexity.
Instead, we take the strategy of fixing the highest function degree and measuring the averaged approximation errors of local surface patches; technical details are given in Appendix \ref{sec:algebraiccomplexity}. We finally obtain the three groups respectively of $972$ instances, $486$ instances, and $162$ instances (at the ratio of around $6:3:1$ for low-, middle-, and high-complexity groups). \cref{tab:filtered_surfaces_composition} summarizes the group statistics.

\begin{table}[hptb]
    \vspace{-0.15cm}
    \centering
    \caption{The benchmark collection of synthetic object instances from existing repositories and their distributions in the three groups of low, middle, and high complexities.}
    \vspace{-0.3cm}
    \label{tab:filtered_surfaces_composition}
    \resizebox{0.45\textwidth}{!}{%
    \begin{tabular}{l|ccc|c}\hline
       Surface complexity & Low & Middle & High & Total \\\hline
        Thingi10k \cite{thingi10k}            & 516    & 230      & 97      & 845   \\
        3DNet Cat200 subset \cite{3dnetdata}  & 144    & 89       & 33      & 266   \\
        ABC \cite{abcdata}                    & 312    & 160      & 6       & 478   \\
        Three D Scans \cite{Threedscans}      & 0      & 7        & 26      & 31    \\\hline
        Our benchmark                                         & 972    & 486      & 162     & 1,620  \\\hline
    \end{tabular}%
    }
    \vspace{-0.4cm}
\end{table}

\subsubsection{Synthetic Point Cloud Scanning}\label{datasysthesizing}

We use Blender Sensor Simulation Toolbox (BlenSor) \cite{blensor} to synthetically scan our collected surface meshes of object instances. We describe in this section our scanning pipeline, including how we implement different ways of imperfect scanning that simulate point cloud sensing happening in practical conditions.

\vspace{0.2cm}\noindent\textbf{Perfect scanning -- }
We first show our way of perfect scanning to give a whole picture of how we conduct our virtual scanning pipline as shown in \cref{fig:synthetic_pipeline}.
As described in Section \ref{sec:collect_preprocess}, our instances of surface meshes have been normalized at the center of a unit sphere in the simulator. To scan an instance, we place a virtual time-of-flight (TOF) camera \cite{tofcamera} on viewing spheres whose radii range from $r_{\textrm{\tiny min}} = 2.5$ to $r_{\textrm{\tiny max}} = 3.5$; a viewpoint on any of the spheres can be specified as the camera extrinsic $\bm{K} = [\bm{R} | \bm{t}] \in \mathbb{R}^{4\times 4}$, where the rotation $\bm{R} \in \mathbb{R}^{4\times 3}$ and translation $\bm{t} \in \mathbb{R}^{4\times 1}$ together specify how the camera is positioned. Denote as $\mathcal{P}_{\bm{K}}$ the point cloud obtained by scanning from the viewpoint $\bm{K}$;
one may transform it into the world coordinate system as $\bm{K}\circ\mathcal{P}_{\bm{K}}$, where $\circ$ denotes an operator\footnote{Note that a point $\bm{p}\in \mathcal{P}_{\bm{K}}$ in the camera coordinate system can be transformed to point $\bm{p}^{\text{\tiny{world}}}\in \mathcal{P}_{\bm{K}}^{\text{\tiny{world}}}$ in the world coordinate system in terms of a homogeneous equation as $[\bm{p}^{\text{\tiny{world}}}; 1]^{\top} = [\bm{p}; 1]^{\top}\bm{K}^{-1}$; here we transform $\mathcal{P}_{\bm{K}}$ to $\mathcal{P}_{\bm{K}}^{\text{\tiny{world}}}$ and write collectively as $\mathcal{P}_{\bm{K}}^{\text{\tiny{world}}} = \bm{K}\circ\mathcal{P}_{\bm{K}}$.}.
In our setting, we sample $1,000$ viewpoints on the spheres of different radii, resulting in a collection of scanned point clouds $\{\bm{K}_1\circ\mathcal{P}_{\bm{K}_1}, \cdots, \bm{K}_{1000}\circ\mathcal{P}_{\bm{K}_{1000}}\}$,  each of which partially covers the surface. \
Then we can register and fuse different partial point clouds to cover the complete surface $\bigcup_{i=1}^{1000} \bm{K}_i\circ\mathcal{P}_{\bm{K}_i}$.
We obtain the final, uniformly distributed point cloud scanning $\mathcal{P}$ by applying Farthest Point Sampling (FPS) \cite{pointnet} to $\bigcup_{i=1}^{1000} \bm{K_i}\circ\mathcal{P}_{\bm{K_i}}$;
$\mathcal{P}$ is set to contain $80$k points for low-complexity surfaces, $120$k points for middle-complexity surfaces, and $160$k points for high-complexity surfaces.

Since some of existing methods require surface normals to preform reconstruction, we compute the oriented surface normals as follows. For any point $\bm{p} \in \mathcal{P}$, we first compute its un-oriented normal $\bar{\bm{n}}_{\bm{p}}$ by performing PCA on the local neighborhood constructed by $k=40$ nearest neighbors of the point; orientation of $\bar{\bm{n}}_{\bm{p}}$ can be simply determined by comparing $\bm{p}$ with the camera position, giving rise to the oriented normal $\bm{n}_{\bm{p}}$.

\vspace{0.2cm}\noindent\textbf{Point-wise noise -- }
Due to sensor noise, ambient noise, reflective nature of the surface, and the incapable precision of the scanning devices, point clouds from practical scanning are inevitably noisy.
In this case, each scanned point is not exactly on the underlying surface, deviating away from the surface in a point-wise, independent manner; and to simulate such noise, we add point-wise perturbations to the points obtained by the aforementioned perfect scanning.
Specifically, for any surface point $\bm{q}$, we generate $\Delta\bm{q}_{\text{\tiny{noise}}} \in \mathbb{R}^3$ by randomly sampling its element values from a Gaussian distribution $\mathcal{N}(0, \sigma_{\textrm{\tiny{noise}}}^2)$ with a truncated values $[-2\sigma_{\text{\tiny{noise}}}, 2\sigma_{\text{\tiny{noise}}}]$.
Given a point $\bm{p} \in \mathcal{P}$, the corresponding noisy point from noisy scanning is obtained as $\bm{p} = \bm{q} + \Delta\bm{q}_{\text{\tiny{noise}}}$.
We set $\sigma_{\text{\tiny{noise}}}$ respectively as $0.001$, $0.003$, and $0.006$ in our benchmark to simulate different severity levels of point-wise noise.
Note that the truncation above is to prevent individual $\{\bm{p} \in \mathcal{P}\}$ from deviating too far away from the surface, which would become point outliers to be discussed shortly.

\vspace{0.2cm}\noindent\textbf{Non-uniform distribution of points -- }
Practical scanning often produces a point cloud whose points are not uniformly distributed over the surface.
For example, the surface patches that are scanned for multiple times (possibly from different viewpoints) would have more points, and a closer scanning position would produce denser points as well.
To simulate such phenomena, we replace the final step of uniformity-promoting FPS in perfect scanning with Random Sampling (RS), and local point densities of the resulting $\mathcal{P}$ would be less uniform over the surface.

\vspace{0.2cm}\noindent\textbf{Point outliers -- }
As mentioned above, outliers of a surface point cloud are defined as those deviating far away from the surface.
They are often caused by impulsive noise of practical scanning.
We simulate such outliers as follows.
Given a point cloud obtained by perfect scanning, we first randomly sample a ratio $r_{\textrm{\tiny{outlier}}}$ of its points, and for any sampled point $\bm{q}$ that is on the surface,
we generate $\Delta\bm{q}_{\textrm{\tiny{outlier}}} \in \mathbb{R}^3$ by randomly sampling its element values from a uniform distribution $\mathcal{U}[a_{\textrm{\tiny{outlier}}}, b_{\textrm{\tiny{outlier}}}]$; the corresponding point outlier $\bm{p}_{\textrm{\tiny{outlier}}} \in \mathcal{P}$ is then obtained as $\bm{p} = \bm{q} \pm \Delta\bm{q}_{\textrm{\tiny{outlier}}}$.
We set $a_{\textrm{\tiny{outlier}}} = 0.01$ to distinguish point outliers from noisy points and set $b_{\textrm{\tiny{outlier}}} = 0.1$ to prevent the outliers from being less relevantly distancing.
The ratio $r_{\textrm{\tiny{outlier}}}$ is respectively set as $0.1\%$, $0.3\%$, and $0.6\%$ for varying numbers of outliers in each obtained $\mathcal{P}$.

\vspace{0.2cm}\noindent\textbf{Misalignment -- }
As described for perfect scanning, a scanning viewpoint is specified by the camera extrinsic $\bm{K} = [\bm{R} | \bm{t}]$; scanning from the viewpoint $\bm{K}$ would produce a point cloud $\mathcal{P}_{\bm{K}}$ that covers the surface partially; a complete point cloud is obtained by scanning from $1,000$ viewpoints and then registering and fusing the obtained point clouds as $\bigcup_{i=1}^{1000} \bm{K}_i\circ\mathcal{P}_{\bm{K}_i}$.
However, misalignment would happen when the camera extrinsics are less accurate.
To simulate such a misalignment, for each viewpoint $\bm{K}$, we generate the perturbation $\Delta\bm{K} = [\Delta\bm{R} | \Delta\bm{t}] \in \mathbb{R}^{4\times 4}$ where
$\Delta\bm{R} \in \mathbb{R}^{4\times 3} $ is obtained by the XYZ Euler angle convention \cite{weisstein2009euler}, i.e., $\Delta\bm{R} = [\bm{R}_{x}(\alpha)\bm{R}_{y}(\beta)\bm{R}_{z}(\gamma); \bm{0}_{3}^{\top}]$ with $\alpha$, $\beta$, and $\gamma$ uniformly sampled from $\mathcal{U}[a_{\textrm{\tiny{rotation}}}, b_{\textrm{\tiny{rotation}}}]$
and those of the translation perturbation $\Delta\bm{t} \in \mathbb{R}^{4\times 1}$ uniformly sampled from $\mathcal{U}[a_{\textrm{\tiny{translation}}}, b_{\textrm{\tiny{translation}}}]$.
The final point cloud $\mathcal{P}$ with misalignment is obtained by applying FPS to $\bigcup_{i=1}^{1000} (\bm{K}_i + \Delta\bm{K}_i)\circ\mathcal{P}_{\bm{K}_i}$.
We set $[a_{\textrm{\tiny{rotation}}}, b_{\textrm{\tiny{rotation}}}]$ as $[-0.5\degree, 0.5\degree]$, $[-1\degree, 1\degree]$, and $[-2\degree, 2\degree]$, and set $[a_{\textrm{\tiny{translation}}}, b_{\textrm{\tiny{translation}}}]$ as $[-0.005, 0.005]$, $[-0.01, 0.01]$, and $[-0.02, 0.02]$, which are respectively for different severities of misalignment.

\vspace{0.2cm}\noindent\textbf{Missing points -- }
Due to surface reflection, self-occlusion, and/or simply insufficient covering of the surface, practical scanning often produces a point cloud that does not cover the whole object surface of interest. Surface reflection depends on a mixed effect of lighting and surface material, and the latter is not included in our collected CAD models. Instead, we take the following simple approach in our benchmark to simulate missing surface points.
Rather than allowing the scanning viewpoint $\bm{K}_i$ to locate on the whole viewing spheres as in perfect scanning (shown as the blue points in \cref{fig:synthetic_pipeline}), we only allow it to locate on a limited number of viewing positions to simulate missing points.
Specifically, we define the viewing positions in a few narrow bands of trajectories with the polar angle of $\varphi\pm\varphi_{\Delta}$ (shown as the yellow points in \cref{fig:synthetic_pipeline}).
For different severities, we respectively set the number of trajectories to be $3$, $2$, and $1$, with the polar angle $\varphi$ set to be $[20\degree, 40\degree, 60\degree]$, $[20\degree, 40\degree]$, and $[20\degree]$ respectively, and with $\varphi_{\Delta} = 3\degree$.
The scanned surface areas approximately cover $99\%$, $94\%$, and $86\%$ of the whole surface for different severities.

\begin{figure*}[hptb]
    \centering
    \includegraphics[width=0.95\textwidth]{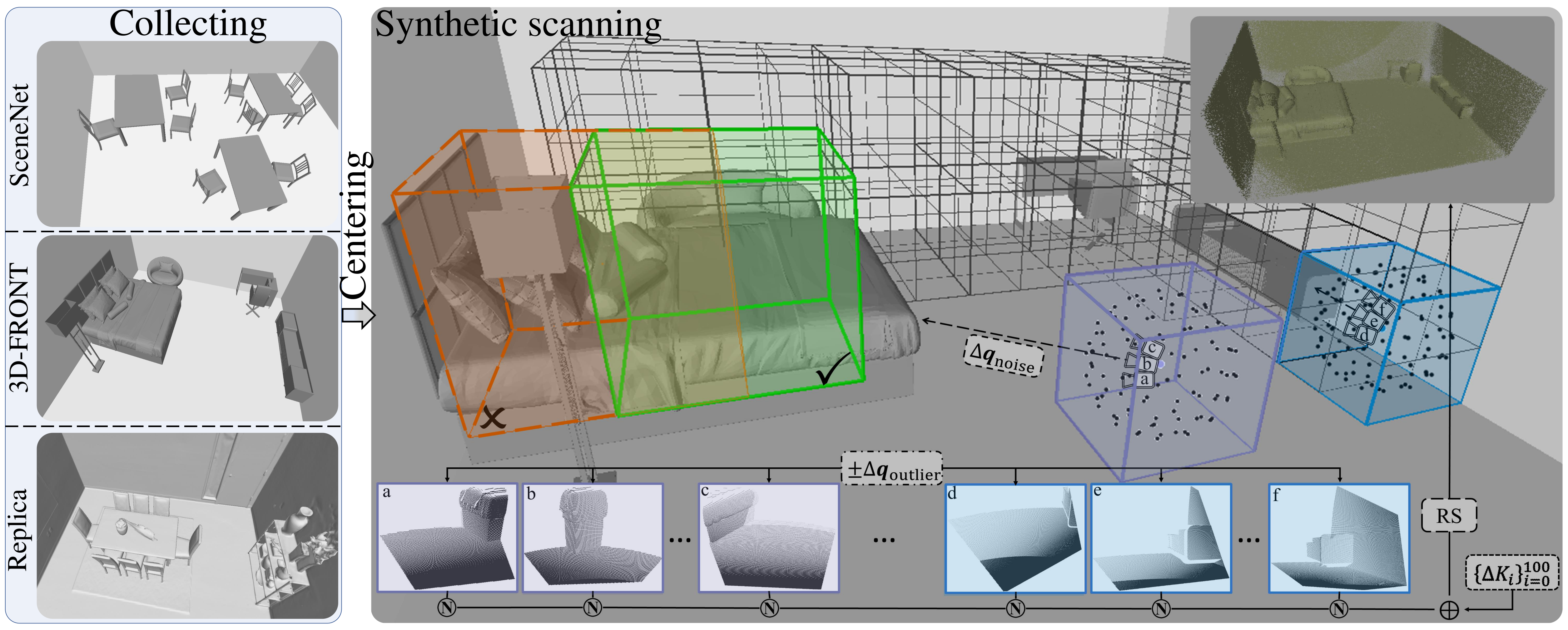}
    \vspace{-0.1in}
    \caption{
    The pipeline of constructing our synthetic scene-level dataset (cf. Section \ref{sec:synthetic_data_scene} for the details).
    We collect CAD surface models of indoor scenes from three datasets, as shown in the left of the figure, and normalize them by centering at the origin in the 3D space.
    We partition the 3D space into overlapped cubes, as shown in the right of the figure; some of the cubes contain certain patches of the scene surface, while others are empty.
    The camera is positioned at centers of those empty cubes (e.g., the purple and blue cubes); for each positioned camera, we randomly sample $100$ viewpoints for the scanning. We include all the five challenges of practical scanning into each scanning of the scene surface, whose individual implementations are similar to those for synthetic object scanning.
    }
    \vspace{-0.35cm}
    \label{fig:synthetic_pipeline_scene}
\end{figure*}

\subsection{The Synthetic Data of Scene Surfaces}\label{sec:synthetic_data_scene}

We adopt a pipeline similar to that presented in Section \ref{sec:synthetic_data} for synthetic scanning of the scene-level data. Since the scale of a scene surface is larger and practical scanning usually produces a point cloud that includes multiple types of imperfections. As such, we include all the five challenges of  point-wise noise, point outliers, non-uniform distribution of points, misalignment, and missing points into our synthetic scanning of a single scene. \cref{fig:synthetic_pipeline_scene} gives the illustration.

\vspace{0.2cm}\noindent\textbf{Data collection -- }
We choose to collect CAD surface models of indoor scenes in the benchmark. To form the collection, we randomly select 4 indoor scenes from SceneNet \cite{scenenet}, 4 from 3D-FRONT \cite{3dfrontdata}, and 2 from Replica \cite{replicadata}.
The scenes from the three datasets are diverse in terms of varying room types, varying room sizes, and the contained different furniture and furnishings.
The collected surfaces are represented in the form of triangular mesh as well. We normalize each scene mesh by centering it at the origin but keeping its original size.

\vspace{0.2cm}\noindent\textbf{Synthetic scanning -- }
We still use BlenSor \cite{blensor} to perform our synthetical scene scanning. In practical scanning of indoor scenes, the scanner is usually placed at a medium distancing from the scene surface; we thus choose the sensor of Kinect V2 \cite{KinectV2_scene} whose working distance ranges from $0.75$m to $2.1$m.
We prepare the scanning by placing each surface mesh of indoor scene in a bounded 3D space, where the scene center has been aligned at the origin and the space size is set to be just enclosing the scene surface (cf. \cref{fig:synthetic_pipeline_scene}).
We firstly partition the 3D space into the volume of $1$m$^3$-sized, $0.5$m$^3$-overlapped cubes; some of the cubes would be empty while others contain certain patches of the scene surface.
We choose the centers of those empty cubes as the positions from which the virtual camera views the scene, and abandon others to ensure that the working distance between the camera and scene surface satisfy the aforementioned Kinect V2 requirements.
From each position, the camera could view a certain patch of the scene surface towards arbitrary directions on a viewing sphere originated at the cube center.
More specifically, for a center of empty cube positioned at $\bm{x} \in \mathbb{R}^{3}$, we make it homogeneous in the form of $\bm{t}_i = [\bm{x}; 1] \in \mathbb{R}^{4}$ and randomly sample $100$ directions to get the camera extrinsics $\{ \bm{K}_{ij} = [\bm{R}_{j} | \bm{t}_i] \}_{j=1}^{100}$, each of which would be used to generate a point cloud $\mathcal{P}_{\bm{K}_{ij}}$ covering a certain patch of the scene surface.
Similar to object scanning, a final point cloud $\mathcal{P}$ is obtained by registering and fusing all the point clouds obtained by the preceding scanning $\{ \bm{K}_{ij}\circ\mathcal{P}_{\bm{K}_{ij}} \}_{i, j}$.
Note that we do not conduct FPS as object-level synthetic scanning does; and consequently, the challenge of non-uniform distribution of points naturally appears here.
Since some methods require surface normals, we compute the oriented normal $\bm{n}_{\bm{p}}$ for all the point $\bm{p} \in \mathcal{P}$ in the same way as object scanning does.
Apart from the natural challenges of non-uniform distribution and missing points (due to self-occlusion), we also simulate the other challenges of point-wise noise, point outliers, and misalignment in the same way as described in Section \ref{datasysthesizing}.
Specifically, following the settings of Kinect V2 camera, we set $\sigma_{\textrm{\tiny{noise}}} = 0.005$m to control the level of point-wise noise, set $a_{\textrm{\tiny{outlier}}} = 0.01$m, $b_{\textrm{\tiny{outlier}}} = 0.1$m, and the ratio $r_{\textrm{\tiny{outlier}}} = 0.4\%$ to control the level of outliers, and set $[a_{\textrm{\tiny{rotation}}}, b_{\textrm{\tiny{rotation}}}]$ as $[-1.5\degree, 1.5\degree]$ and $[a_{\textrm{\tiny{translation}}}, b_{\textrm{\tiny{translation}}}]$ as $[-0.015$m$,0.015$m$]$ to control the level of misalignment.
Such a point cloud $\mathcal{P}$ contains around one million points containing all the five challenges.

\begin{table}[hptb]
    \vspace{-0.15cm}
    \caption{
    Statistics of synthetic data in the benchmark.
    Elements in each triple $\cdot$ / $\cdot$ / $\cdot$ represent the hyper-parameters that control the scanning with three levels of severity; \# denotes the number.
    (cf. Sections \ref{sec:synthetic_data} and \ref{sec:synthetic_data_scene} for specific meanings of the math notations.)
    }
    \vspace{-0.3cm}
    \label{table:statistics_summary}
    \centering
    \resizebox{\linewidth}{!}{
    \begin{tabular}{l|l|c|c|c}
        \hline
        \multicolumn{2}{l|}{} & \multicolumn{2}{c|}{Object level} & Scene level \\
        \hline
        \multicolumn{2}{l|}{normalization} & \multicolumn{2}{c|}{centering+scaling} & centering \\
        \multicolumn{2}{l|}{[$r_{\textrm{\tiny min}}$, $r_{\textrm{\tiny max}}$] (camera distancing)} & \multicolumn{2}{c|}{[$2.5$, $3.5$]} & [$0.75$m, $2.1$m] \\
        \hline
        point-wise noise & $\sigma_{\textrm{\tiny{noise}}}$ & \multicolumn{2}{c|}{$0.001/0.003/0.006$} & $0.005$m \\
        \hline
        \multirow{2}{*}{point outliers} & [$a_{\textrm{\tiny{outlier}}}$, $b_{\textrm{\tiny{outlier}}}$] & \multicolumn{2}{c|}{[$0.01$, $0.1$]} & [$0.01$m, $0.1$m] \\
        ~ & $r_{\textrm{\tiny{outlier}}}$ & \multicolumn{2}{c|}{$0.1\%/0.3\%/0.6\%$} & $0.4\%$ \\
        \hline
        %
        %
        \multirow{2}{*}{\makecell{missing points}} & $\varphi$ & \multicolumn{2}{c|}{$[20\degree, 40\degree, 60\degree] / [20\degree, 40\degree] / [20\degree]$} & \multirow{2}{*}{-} \\
        ~ & $\Delta\varphi$ & \multicolumn{2}{c|}{$3\degree$} & ~ \\
        \hline
        \multirow{2}{*}{\makecell{misalignment}} & [$a_{\textrm{\tiny{rotation}}}$, $b_{\textrm{\tiny{rotation}}}$] & \multicolumn{2}{c|}{$[-0.5\degree$, $0.5\degree]/[-1\degree$, $1\degree]/[-2\degree$,$2\degree]$} & [$-1.5\degree$, $1.5\degree$] \\
        ~ & [$a_{\textrm{\tiny{translation}}}$, $b_{\textrm{\tiny{translation}}}$] & \multicolumn{2}{c|}{[$-0.005$, $0.005]/[-0.01$, $0.01]/[-0.02$, $0.02]$} & [$-0.015$m, $0.015$m] \\ \hline
        \multicolumn{2}{l|}{\#viewpoints} & \multicolumn{2}{c|}{1000} & $100/\textrm{m}^3$ \\\hline
        \multicolumn{2}{l|}{\multirow{3}{*}{\#scanned points}} & low complexity     & $80$k  & ~         \\
        \multicolumn{2}{l|}{}                                    & middle complexity  & $120$k & $1000$k   \\
        \multicolumn{2}{l|}{}                                    & high complexity    & $160$k & ~         \\  \hline
        \multicolumn{2}{l|}{\#surfaces} & \multicolumn{2}{c|}{1620} & 10 \\ \hline
    \end{tabular}
    }
    \vspace{-0.4cm}
\end{table}

\subsection{The Real-scanned Data}

We provide real-scanned data of the benchmark by scanning real object instances via two depth cameras of varying precisions.
To scan an object, we use SHINING 3D Einscan SE \footnote{\url{https://www.einscan.com/desktop-3d-scanners/einscan-se/}} whose precision is of $100$ micrometers to get the input point cloud, and use SHINING 3D OKIO 5M \footnote{{\url{https://www.shining3d.com/solutions/optimscan-5m}}} whose precision is of $5$ micrometers to get the approximate ground-truth.\footnote{Our surface reconstruction evaluation is based on metrics of point set distances (cf. \cref{Exp:metrics}), for which we obtain the ground-truth point clouds by sampling from the corresponding surface meshes. We thus choose to directly use the raw point clouds scanned by SHINING 3D OKIO 5M, instead of converting the scanned point clouds as surface meshes using its in-built software. }

\begin{figure}[hptb]
    \centering
    \includegraphics[width=0.35\textwidth]{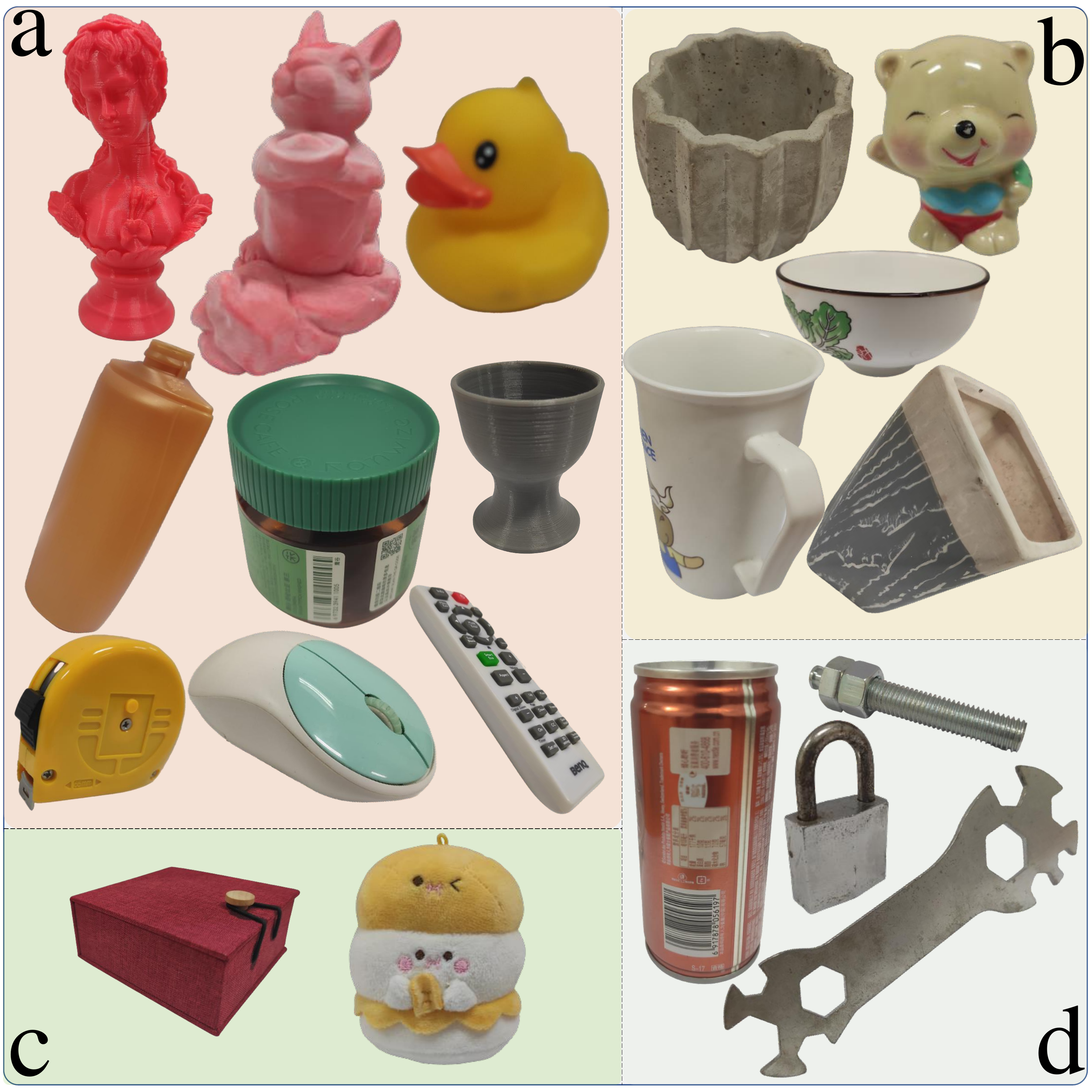}
\vspace{-0.1in}
\caption{
    An illustration of the objects used in our real-scanned dataset.
    The objects are organized by the types of material, where ``a'' shows the objects made of plastic, ``b'' for ceramic, ``c'' for cloth, and ``d'' for metal.
}
\label{fig:scan_obj}
\vspace{-0.2cm}
\end{figure}

\vspace{0.2cm}\noindent\textbf{Data collection -- }
We collect $20$ object instances of varying surface complexities, including commodities, instruments, and artwares, and also of varying materials, including  metal, plastic, ceramic, and cloth; this is to ensure that various sensing imperfections would appear in the obtained point clouds.
\cref{fig:scan_obj} shows these objects.

\begin{wrapfigure}{r}{0.20\textwidth}
    \vspace{-15pt}
    \centering
    \includegraphics[width=0.20\textwidth]{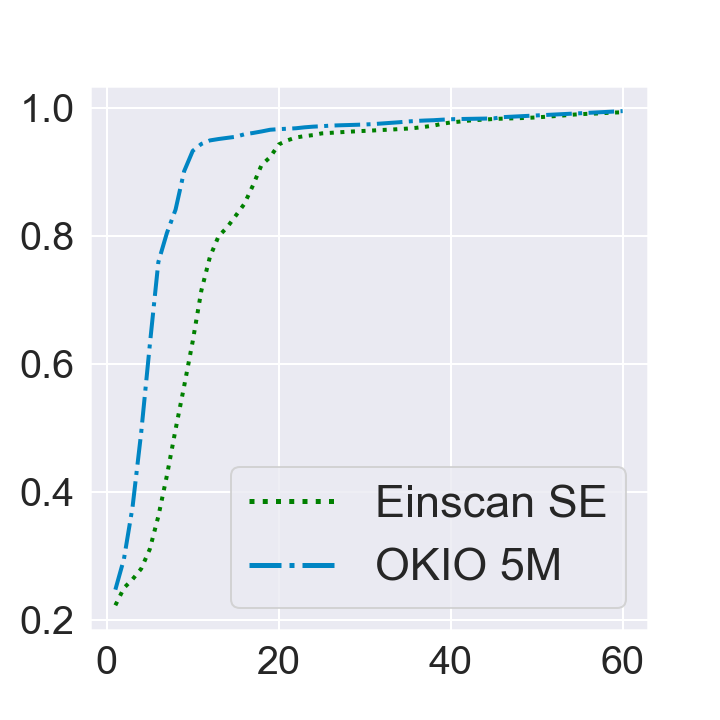}
    \caption{{The relationship between scan quality and scan shots.}}
    \label{CommercialScannerShots}
    \vspace{-10pt}
\end{wrapfigure}

\begin{figure}[hptb]
    \vspace{-0.75cm}
    \centering
    \includegraphics[width=0.40\textwidth]{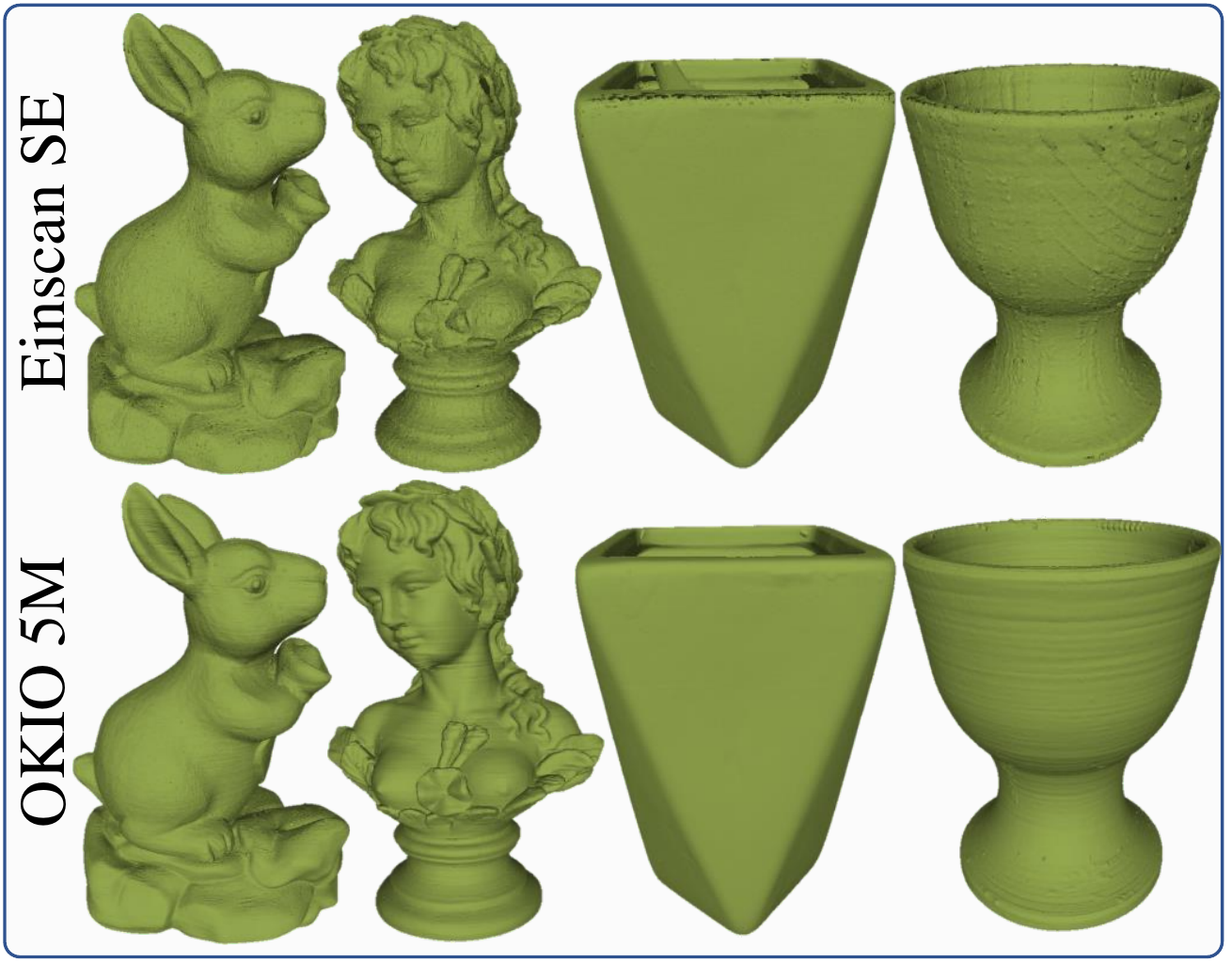}
\vspace{-0.1in}
\caption{
    Examples of real-scanned object pairs.
}
\label{fig:real_scan_pairs}
\vspace{-0.75cm}
\end{figure}

\vspace{0.2cm}\noindent\textbf{Real scanning -- }
In general, better scanning results could be obtained by increasing the numbers of scanning shots from multiple viewpoints, as empirically verified in Fig. \ref{CommercialScannerShots}.
Since qualities of the scanned point clouds start to saturate at around 40 shots for Einscan SE and 20 shots for OKIO 5M, we conduct 40 shots for Einscan SE and 20 shots for OKIO 5M shots when scanning an object.
After scanning an object, we use CloudCompare \cite{girardeau2016cloudcompare} to align different point clouds obtained from different shots.
\cref{table:statistics_summary_real} gives the statistics of our real-scanned data in the benchmark. Some scanned pairs from the two scanners are visualized in \cref{fig:real_scan_pairs}.

\begin{table}[hptb]
    \caption{Statistics of real-scanned data. \# denotes the number.}
    \vspace{-0.1in}
    \label{table:statistics_summary_real}
    \centering
    \begin{tabular}{l|l|c|c}
    \hline
    \multicolumn{2}{l|}{} & Einscan  SE  & OKIO 5M \\ \hline
    \multicolumn{2}{l|}{precision} & $100$ micrometers & $5$ micrometers \\\hline
    \multicolumn{2}{l|}{professional operators} & $\times$ & $\checkmark$ \\\hline
    \multicolumn{2}{l|}{\#resolution} & $1300k$ & $5000k$ \\\hline
    \multicolumn{2}{l|}{\#viewpoints} & $30$-$50$ & $13$-$35$ \\\hline
    \multicolumn{2}{l|}{\#scanned points} & $210k$-$3000k$ & $330k$-$2000k$ \\\hline
    \multicolumn{2}{l|}{\#surfaces} & \multicolumn{2}{c}{20}  \\ \hline
    \multirow{4}{*}{\#material} & metal & \multicolumn{2}{c}{4}  \\
    ~ & plastic & \multicolumn{2}{c}{9}  \\
    ~ & ceramic & \multicolumn{2}{c}{5}  \\
    ~ & cloth & \multicolumn{2}{c}{2}  \\ \hline
    \end{tabular}
\end{table}

\section{Experimental Set-up for Benchmarking Existing Surface Reconstruction Methods}
\label{sec:processing}

With the benchmark prepared in Section \ref{sec:Challenges}, we aim to empirically compare existing surface reconstruction methods, and identify their advantages and disadvantages in different working conditions (e.g., various scanning imperfections considered in our benchmark).
We expect such studies would both provide insights for future research in this area, and guide the use of appropriate methods for practical surface reconstruction from point observations.
Such a comprehensive investigation could be timely for the community, given that a plethora of new methods have been proposed recently, in particular those based on deep learning.

\subsection{Data}
\label{SecExpData}

We conduct the empirical studies using a subset of our collected benchmark, while publicly releasing the whole benchmark to facilitate the community research. More specifically, we randomly sample $22$ synthetic surfaces of object instances, whose distribution in the three groups of low-, middle-, and high-complexity is $12:6:4$; as described in Section \ref{sec:synthetic_data}, for each instance we conduct six ways of synthetic scanning, which produce either a clean point cloud or point clouds with various imperfections, giving rise to a total of $308$ input-output pairs for benchmarking algorithms.
We also use all the $10$ synthetic scene surfaces and all the $20$ real-scanned surfaces for the studies.
For some of existing methods using learning-based priors, we prepare an auxiliary set of training data consisting of surfaces from ShapeNet \cite{shapenet} and ABC \cite{abcdata}, and also the remaining synthetic data of object instances from our benchmark.

\subsection{Pre-processing}
\label{SecPreprocessing}
While existing surface reconstruction methods can be compared using the data prepared in Section \ref{SecExpData}, for most of them, their performance could be greatly improved via some standard pipeline of point cloud pre-processing.
In this work, we compare existing methods both without and with such a pre-processing pipeline. For synthetic data, the pipeline is in the order of outlier removal, de-noising, and point re-sampling; details are given as follows. For real-scanned data, we use the inbuilt pre-processing of different scanners, and use a final step of FPS to re-sample $200,000$ points for each point cloud.

\vspace{0.2cm}\noindent\textbf{Outlier removal -- }
Performance of surface reconstruction degrades severely when extreme outliers exist in a point cloud; fortunately, these outliers are easy to be removed. We use a statistical method \cite{rusu2008towards_statisticaloutlierremoval} to remove extreme point outliers. For a point cloud $\mathcal{P}$, it regards any $\bm{p} \in \mathcal{P}$ as an outlier and remove it when $\bm{p}$ is very far away from its local neighborhood. More precisely, for any $\bm{p} \in \mathcal{P}$, we first compute the averaged distance $\bar{d}_{\bm{p}}$ between $\bm{p}$ and its $k$ nearest neighbors in $\mathcal{P}$; we then compute the mean $m_{\bar{d}}$ and standard deviation $\sigma_{\bar{d}}$ of such distances for all $\{ \bm{p} \in \mathcal{P} \}$; a point $\bm{p}$ is regarded as an outlier when its corresponding $\bar{d}_{\bm{p}} > 5 \cdot \sigma_{\bar{d}}$. We set $k = 35$ in this work for outlier removal.

\vspace{0.2cm}\noindent\textbf{De-noising -- }
The inevitable existence of point-wise noise influences surface reconstruction as well. For an input point cloud $\mathcal{P}$, we choose to suppress such noise using Jets smoothing \cite{Jet_smooth_Denoise}, which smoothes out the point cloud without sacrificing its surface curvatures.
It works by first fitting a parametric surface patch to a local neighborhood $\mathcal{N}$ of $k$ points in $\mathcal{P}$, and then projecting $\{\bm{p} \in \mathcal{N}\}$ onto the fitted surface patch. We set $k = 18$ in this work for point-wise de-noising.

\vspace{0.2cm}\noindent\textbf{Point re-sampling -- }
Empirical results show that surface reconstruction benefits from more uniform distribution of points, even when reducing the number of points contained in $\mathcal{P}$ \cite{yang20173d}.
For the synthetic data of object or scene surfaces, we simply use farthest point sampling \cite{pointnet} as the method to re-sample a more uniform distribution of points; we preserve $40\%$ of original points during re-sampling.

\subsection{Evaluation Metrics}
\label{Exp:metrics}

We quantitatively compare reconstruction results from different methods using the popular metrics of Chamfer Distance (CD) \cite{CD}, F-score \cite{largesaclebenchmark}, and Normal Consistency Score (NCS) \cite{occupancy}; Appendix \ref{sec:Evaluation} specifies their computations.
We also propose a neural metric, termed Neural Feature Similarity (NFS), focusing on perceptual consistency between each reconstruction and the ground-truth; intuitively speaking, NFS compares the similarity of two shapes in the deep feature space, and thus depends more on the high-level semantic information consistent with human perception \cite{deepfeatureperceptualmetric}; details are given in Appendix \ref{sec:Evaluation} as well.

CD and F-score are used for measuring the overall similarity between two shapes;
NCS is more useful for measuring the nuance of two similar shapes by measuring their consistency of surface normals;
NFS measures semantic difference related to human perception.

\subsection{Methods and Implementation Details}\label{SecCompMethodAndImplementations}

In Section \ref{sec:learning_prior}, we have categorized existing methods according to what geometric priors they have respectively used to regularize the reconstruction.
It is less feasible to study and empirically compare all the existing methods; instead, we take the strategy of selecting representative ones from each method group of geometric priors, assuming that our studies and conclusions would generalize in the same groups of existing methods. More specifically, we adopt the most representative Greedy Delaunay (GD) \cite{greedydelaunay} and BPA \cite{BPA} as the methods to be studied for triangulation-based prior;
for priors of surface smoothness, we adopt SPSR \cite{kazhdan2013screened} using the first manner of surface smoothness (cf. \cref{eq:psr}) and RIMLS \cite{RIMLS} using both two manners of surface smoothness (cf. \cref{eq:L_first_order_ex_regu} and constraining $\mathcal{H}_{\bm{f}}$);
for modeling priors, we adopt SALD \cite{sald} that is able to reconstruct surfaces from un-oriented point clouds, and IGR \cite{gropp2020implicitgeometricregularizationforlearningshape} that can do so from oriented ones; for learning-based priors, we adopt the global, semantic learning methods of OccNet \cite{occupancy} and DeepSDF \cite{deepsdf}, and also the local learning methods of LIG \cite{LIG} and Points2Surf \cite{points2surf};
we consider three methods that use hybrid priors, including Delaunay Surface Elements (DSE) \cite{learningdelaunaysurface} that combines triangulation-based prior with learning-based prior, IMLSNet \cite{liu2021DeepIMLS} that combines surface smoothness prior with learning-based prior, and ParseNet \cite{parsenet} that combines template-based prior with learning-based prior;
we do not consider methods using template-based priors \emph{only}, given that their performance largely depends on whether there would exist a good match between a surface to be reconstructed and the assumed templates (e.g., the assumed geometric primitives or the templates that can be retrieved in an auxiliary dataset), and that even the one with learning-based templates \cite{parsenet} fail to reconstruct surfaces of certain complexities.

Our implementations of the more classical, learning-free methods are based on established libraries; for example, we implement GD using the CGAL library \cite{cgal}, and directly execute BPA \cite{BPA}, SPSR \cite{kazhdan2013screened}, and RIMLS \cite{RIMLS} with proper parameter tunings in MeshLab \cite{meshlab_LocalChapterEvents:ItalChap:ItalianChapConf2008:129-136}.
For those learning-based methods, we use the implementation codes publicly released by the authors when they are available, again with proper tuning of hyper-parameters for individual surfaces to be reconstructed.
For those without code releasing, we re-implement their algorithms and tune the respective hyper-parameters as the optimal ones.

In \cref{sec:experiments}, we report and discuss the results \emph{that are obtained with the pre-processing mentioned in \cref{SecPreprocessing}}. We note that those without pre-processing are of similar comparative qualities, and we put them in Appendix \ref{sec:more_exp_wo_pre}.

\section{Main Results}
\label{sec:experiments}

We first summarize our key empirical findings, before presenting details of our series of experiments; insights are drawn subsequently.
\begin{itemize}
    \item While many challenges of surface reconstruction from point clouds can be more or less tackled by using different regularization/priors of surface geometry, the challenges of \emph{misalignment}, \emph{missing points}, and \emph{outliers} have been less addressed and remain unsolved.
    \item Data-driven solutions using deep learning have recently shown great promise for surface modeling and reconstruction, including their potential to deal with various data imperfections, however, our systematic experiments suggest that they struggle in generalizing to reconstruction of complex shapes; it is surprising that some classical methods such as SPSR \cite{kazhdan2013screened} perform even better in terms of generalization and robustness.
    \item Use of surface normals is a key to success of surface reconstruction from raw, observed point clouds, even when the surface normals are estimated less accurately; in many cases, the reconstruction result improves as long as the interior and exterior of a surface can be identified in the 3D space.
    \item There exist inconsistencies between different evaluation metrics, and in many cases, good quantitative results do not translate as visually pleasant ones. For example, quantitative results measured by CD and F-score are not much affected by the challenge of misalignment; however, the reduced scores of NCS and NFS suggest that the recovered surfaces might be less pleasant to human perception.
\end{itemize}

\begin{table*}[hptb]
\centering
\caption{Quantitative results on the testing synthetic data of object surfaces.
Comparisons are made on data of perfect scanning and those of all the five challenges of imperfect scanning specified in Section \ref{datasysthesizing}; for those challenges with varying levels of severity, we use the data of middle-level severity for the comparison
(cf. Appendix \ref{sec:all_exp_syn_obj} for the overall results).
Results of the \textbf{best} and \uline{\textbf{second best}} methods are highlighted in each column. Comparative methods are also grouped according to what priors of surface geometry they have used (cf. Section \ref{sec:review} for the grouping and Section \ref{SecCompMethodAndImplementations} for how these representative methods are selected).
}
\vspace{-0.1in}
\label{tab:syn_obj_whole_chellenges_mid_whole_mectrics}
\resizebox{0.91\textwidth}{!}{%
\begin{tabular}{c|l|cccccc|cccccc}
\hline
\multirow{2}{*}{Prior} & \multirow{2}{*}{Method} & \multicolumn{6}{c|}{CD $ (\times10^{-4}) \downarrow$} & \multicolumn{6}{c}{F-score $(\%) \uparrow$} \\\cline{3-14}
~ & ~ & \makecell{Perfect\\scanning} & \makecell{Non-uniform\\distribution} & \makecell{Point-wise\\noise} & \makecell{Point\\outliers} & \makecell{Missing\\points} & \makecell{Mis-\\alignment}& \makecell{Perfect\\scanning} & \makecell{Non-uniform\\distribution} & \makecell{Point-wise\\noise} & \makecell{Point\\outliers} & \makecell{Missing\\points} & \makecell{Mis-\\alignment} \\\hline
\multirow{2}{*}{\makecell{Triangulation-\\based}}   & GD \cite{greedydelaunay}                                                & \textbf{14.24}               & \textbf{14.87}               & 18.20                        & 123.19                       & 64.15                        & 21.15                        & \textbf{99.66}               & \uline{\textbf{99.07}}       & 98.91                        & 54.59                        & 87.72                        & 97.41                    \\
                                                    & BPA \cite{BPA}                                                          & 14.89                        & 17.13                        & 18.58                        & \textbf{14.66}               & \uline{\textbf{62.55}}       & 20.88                        & 98.70                        & 97.02                        & 98.51                        & \uline{\textbf{99.31}}       & \uline{\textbf{87.85}}       & 97.24                    \\\cline{1-2}
\multirow{2}{*}{\makecell{Smoothness}}              & SPSR \cite{kazhdan2013screened}                                         & 14.47                        & 15.36                        & \textbf{16.05}               & \uline{\textbf{14.71}}       & 225.66                       & \textbf{17.24}               & 99.59                        & 99.02                        & \textbf{99.46}               & \textbf{99.65}               & 76.91                        & \textbf{99.27}           \\
                                                    & RIMLS \cite{RIMLS}                                                      & 15.73                        & 16.74                        & \uline{\textbf{17.17}}       & 126.40                       & 65.36                        & 21.12                        & 99.27                        & 98.76                        & \uline{\textbf{99.24}}       & 57.78                        & \textbf{87.92}               & 97.53                    \\\cline{1-2}
\multirow{2}{*}{\makecell{Modeling}}                & SALD \cite{sald}                                                        & 15.10                        & \uline{\textbf{14.96}}       & 18.77                        & 53.65                        & \textbf{55.63}               & 20.09                        & 99.45                        & \textbf{99.10}               & 98.94                        & 88.05                        & 85.78                        & 98.09                    \\
                                                    & IGR \cite{gropp2020implicitgeometricregularizationforlearningshape}     & 18.40                        & 18.33                        & 18.57                        & 43.60                        & 151.53                       & 20.72                        & 97.54                        & 97.79                        & 97.75                        & 91.11                        & 70.49                        & 96.46                    \\\cline{1-2}
\multirow{2}{*}{\makecell{Learning\\Semantics}}     & OccNet \cite{occupancy}                                                 & 201.96                       & 210.80                       & 205.21                       & 225.85                       & 231.65                       & 212.51                       & 31.03                        & 29.90                        & 29.77                        & 23.75                        & 24.52                        & 29.19                    \\
                                                    & DeepSDF \cite{deepsdf}                                                  & 229.18                       & 227.42                       & 230.40                       & 511.36                       & 378.58                       & 232.16                       & 17.79                        & 18.81                        & 17.08                        & 4.15                         & 14.06                        & 17.05                    \\\cline{1-2}
\multirow{2}{*}{\makecell{Local\\Learning}}         & LIG \cite{LIG}                                                          & 23.09                        & 24.03                        & 22.05                        & 115.38                       & 70.98                        & 24.30                        & 96.20                        & 95.32                        & 96.99                        & 76.69                        & 82.43                        & 95.01                    \\
                                                    & Points2Surf \cite{points2surf}                                          & 17.18                        & 18.81                        & 18.48                        & 83.91                        & 102.18                       & 20.36                        & 98.14                        & 97.72                        & 97.39                        & 72.94                        & 76.46                        & 96.49                    \\\cline{1-2}
                                                    & DSE \cite{learningdelaunaysurface}                                      & \uline{\textbf{14.26}}       & 15.34                        & 17.89                        & 100.37                       & 68.88                        & \uline{\textbf{20.06}}       & \uline{\textbf{99.64}}       & 98.84                        & 99.17                        & 52.21                        & 87.71                        & \uline{\textbf{98.20}}   \\
\multirow{2}{*}{\makecell{Hybird}}                  & IMLSNet \cite{liu2021DeepIMLS}                                          & 22.56                        & 23.17                        & 22.67                        & 99.95                        & 74.35                        & 23.77                        & 94.82                        & 94.51                        & 94.36                        & 64.55                        & 80.01                        & 94.14                    \\
                                                    & ParseNet \cite{parsenet}                                                & 162.94                       & 161.14                       & 135.84                       & 176.38                       & 195.98                       & 136.86                       & 40.52                        & 38.82                        & 44.13                        & 37.21                        & 41.28                        & 46.60                    \\\hline\hline
\multirow{2}{*}{Prior} & \multirow{2}{*}{Method} & \multicolumn{6}{c|}{NCS $ (\times10^{-2}) \uparrow$} & \multicolumn{6}{c}{NFS $ (\times10^{-2}) \uparrow$} \\\cline{3-14}
~ & ~ & \makecell{Perfect\\scanning} & \makecell{Non-uniform\\distribution} & \makecell{Point-wise\\noise} & \makecell{Point\\outliers} & \makecell{Missing\\points} & \makecell{Mis-\\alignment}& \makecell{Perfect\\scanning} & \makecell{Non-uniform\\distribution} & \makecell{Point-wise\\noise} & \makecell{Point\\outliers} & \makecell{Missing\\points} & \makecell{Mis-\\alignment} \\\hline
\multirow{2}{*}{\makecell{Triangulation-\\based}}   & GD \cite{greedydelaunay}                                                & 98.57                        & 98.05                        & 87.58                        & 92.17                        & \uline{\textbf{95.17}}       & 84.96                        & 95.22                        & 95.19                        & 84.63                        & 37.47                        & \uline{\textbf{82.93}}       & 80.69                   \\
                                                    & BPA \cite{BPA}                                                          & 98.37                        & 97.89                        & 91.68                        & \uline{\textbf{98.07}}       & 94.42                        & 90.12                        & 94.10                        & 93.60                        & 87.49                        & \uline{\textbf{93.24}}       & 80.36                        & 81.96                   \\\cline{1-2}
\multirow{2}{*}{\makecell{Smoothness}}              & SPSR \cite{kazhdan2013screened}                                         & 98.58                        & \uline{\textbf{98.38}}       & \uline{\textbf{97.03}}       & \textbf{98.56}               & 89.99                        & \uline{\textbf{96.24}}       & \textbf{96.38}               & \uline{\textbf{96.22}}       & \textbf{94.98}               & \textbf{96.31}               & 69.34                        & \textbf{94.28}          \\
                                                    & RIMLS \cite{RIMLS}                                                      & 98.19                        & 97.77                        & 95.23                        & 83.42                        & 93.27                        & 92.48                        & 95.01                        & 94.02                        & 92.67                        & 62.01                        & 79.23                        & 87.12                   \\\cline{1-2}
\multirow{2}{*}{\makecell{Modeling}}                & SALD \cite{sald}                                                        & \textbf{98.67}               & \textbf{98.52}               & 96.42                        & 95.32                        & \textbf{95.19}               & 94.73                        & \uline{\textbf{96.11}}       & \textbf{96.65}               & 89.72                        & 51.74                        & \textbf{84.01}               & 88.26                   \\
                                                    & IGR \cite{gropp2020implicitgeometricregularizationforlearningshape}     & 97.62                        & 97.59                        & \textbf{97.52}               & 96.47                        & 90.61                        & \textbf{97.04}               & 94.71                        & 94.22                        & \uline{\textbf{94.52}}       & 84.63                        & 68.14                        & \uline{\textbf{92.54}}  \\\cline{1-2}
\multirow{2}{*}{\makecell{Learning\\Semantics}}     & OccNet \cite{occupancy}                                                 & 79.55                        & 79.30                        & 79.64                        & 78.55                        & 78.43                        & 79.58                        & 47.33                        & 46.55                        & 46.03                        & 42.11                        & 42.46                        & 45.80                   \\
                                                    & DeepSDF \cite{deepsdf}                                                  & 78.65                        & 79.11                        & 77.78                        & 73.40                        & 74.52                        & 78.12                        & 39.94                        & 40.91                        & 39.29                        & 16.65                        & 31.59                        & 39.26                   \\\cline{1-2}
\multirow{2}{*}{\makecell{Local\\Learning}}         & LIG \cite{LIG}                                                          & 96.49                        & 95.79                        & 94.22                        & 91.66                        & 89.56                        & 92.70                        & 91.58                        & 90.66                        & 89.55                        & 66.21                        & 74.98                        & 84.34                   \\
                                                    & Points2Surf \cite{points2surf}                                          & 95.24                        & 95.09                        & 94.62                        & 87.87                        & 86.36                        & 94.48                        & 93.45                        & 93.23                        & 92.59                        & 63.30                        & 68.53                        & 91.59                   \\\cline{1-2}
                                                    & DSE \cite{learningdelaunaysurface}                                      & \uline{\textbf{98.60}}       & 97.86                        & 87.79                        & 77.34                        & 94.40                        & 86.20                        & 94.50                        & 94.75                        & 83.53                        & 42.32                        & 79.62                        & 76.63                   \\
\multirow{2}{*}{\makecell{Hybird}}                  & IMLSNet \cite{liu2021DeepIMLS}                                          & 96.13                        & 95.98                        & 96.02                        & 87.45                        & 90.48                        & 95.87                        & 90.61                        & 90.20                        & 89.97                        & 52.82                        & 74.59                        & 89.19                   \\
                                                    & ParseNet \cite{parsenet}                                                & 77.71                        & 76.89                        & 80.31                        & 75.46                        & 75.83                        & 80.48                        & 38.54                        & 37.71                        & 49.30                        & 35.98                        & 38.40                        & 45.73                   \\\hline

\end{tabular}%
}
\end{table*}

\begin{table}[hptb]
\centering
\vspace{-0.2cm}
\caption{Quantitative results on the testing synthetic data of scene surfaces. Results of the \textbf{best} and \uline{\textbf{second best}} methods are highlighted in each column. Comparative methods are grouped according to what priors of surface geometry they have used (cf. Section \ref{sec:review} for the grouping and Section \ref{SecCompMethodAndImplementations} for how these representative methods are selected).  ``-'' indicates the method cannot produce reasonable results due to their limited generalization.
}
\vspace{-0.1in}
\label{tab:syn_scene}
\resizebox{\linewidth}{!}{%
\begin{tabular}{c|l|c|c|c|c}
    \hline
    Prior & Method & \makecell{CD\\$(\times10^{-3}) \downarrow$} & \makecell{F-score\\$(\%) \uparrow$} & \makecell{NCS\\$(\times10^{-2}) \uparrow$} & \makecell{NFS\\$(\times10^{-2}) \uparrow$} \\ \hline
    \multirow{2}{*}{\makecell{Triangulation-\\based}}   & GD \cite{greedydelaunay}                                              & 33.85                        & 75.95                        & 62.08                        & 41.92                       \\
                                                        & BPA \cite{BPA}                                                        & 45.82                        & 53.46                        & 58.25                        & 44.19                       \\\cline{1-2}
    \multirow{2}{*}{\makecell{Smoothness}}              & SPSR \cite{kazhdan2013screened}                                       & \textbf{30.47}               & \textbf{83.22}               & 83.74                        & \uline{\textbf{63.03}}      \\
                                                        & RIMLS \cite{RIMLS}                                                    & 41.45                        & 74.56                        & 69.93                        & 38.13                       \\\cline{1-2}
    \multirow{2}{*}{\makecell{Modeling}}                & SALD \cite{sald}                                                      & 32.43                        & 79.03                        & \textbf{91.58}               & 52.79                       \\
                                                        & IGR \cite{gropp2020implicitgeometricregularizationforlearningshape}   & \uline{\textbf{31.41}}       & \uline{\textbf{81.63}}       & \uline{\textbf{91.26}}       & \textbf{67.58}              \\\cline{1-2}
    \multirow{2}{*}{\makecell{Learning\\Semantics}}     & OccNet \cite{occupancy}                                               & 93.12                        & 37.75                        & 85.98                        & 50.34                       \\
                                                        & DeepSDF \cite{deepsdf}                                                &   -                          &  -                           &  -                           &   -                         \\\cline{1-2}
    \multirow{2}{*}{\makecell{Local\\Learning}}         & LIG \cite{LIG}                                                        & 41.40                        & 78.03                        & 88.12                        & 59.97                       \\
                                                        & Points2Surf \cite{points2surf}                                        & 36.24                        & 76.14                        & 83.60                        & 61.82                       \\\cline{1-2}
                                                        & DSE \cite{learningdelaunaysurface}                                    & 32.97                        & 77.53                        & 57.99                        & 41.56                       \\
    \multirow{2}{*}{\makecell{Hybird}}                  & IMLSNet \cite{liu2021DeepIMLS}                                        & 35.52                        & 78.05                        & 87.17                        & 61.98                       \\
                                                        & ParseNet \cite{parsenet}                                              &   -                          &  -                           &  -                           &   -                         \\\hline
\end{tabular}%
}
\vspace{-0.06in}
\end{table}

\begin{table}[hptb]
\vspace{-0.06in}
\centering
\caption{
Quantitative results on the real-scanned data. Results of the \textbf{best} and \uline{\textbf{second best}} methods are highlighted in each column. Comparative methods are grouped according to what priors of surface geometry they have used (cf. Section \ref{sec:review} for the grouping and Section \ref{SecCompMethodAndImplementations} for how these representative methods are selected). }
\vspace{-0.1in}
\label{tab:real_obj}
\resizebox{\linewidth}{!}{%
\begin{tabular}{c|l|c|c|c|c}
    \hline
    Prior & Method & \makecell{CD\\$(\times10^{-2}) \downarrow$} & \makecell{F-score\\$(\%) \uparrow$} & \makecell{NCS\\$(\times10^{-2}) \uparrow$} & \makecell{NFS\\$(\times10^{-2}) \uparrow$} \\ \hline
    \multirow{2}{*}{\makecell{Triangulation-\\based}}   & GD \cite{greedydelaunay}                                              & 31.72                  & 87.51                  & 88.86                  & 82.20                        \\
                                                        & BPA \cite{BPA}                                                        & 40.37                  & 80.95                  & 87.56                  & 68.69                        \\\cline{1-2}
    \multirow{2}{*}{\makecell{Smoothness}}              & SPSR \cite{kazhdan2013screened}                                       & \textbf{31.05}         & \textbf{87.74}         & \uline{\textbf{94.94}} & \textbf{89.38}               \\
                                                        & RIMLS \cite{RIMLS}                                                    & 32.80                  & 87.05                  & 91.97                  & 85.19                        \\\cline{1-2}
    \multirow{2}{*}{\makecell{Modeling}}                & SALD \cite{sald}                                                      & \uline{\textbf{31.13}} & \uline{\textbf{87.72}} & 94.68                  & 86.86                        \\
                                                        & IGR \cite{gropp2020implicitgeometricregularizationforlearningshape}   & 32.70                  & 87.18                  & \textbf{95.99}         & \uline{\textbf{89.10}}       \\\cline{1-2}
    \multirow{2}{*}{\makecell{Learning\\Semantics}}     & OccNet \cite{occupancy}                                               & 232.71                 & 17.11                  & 80.96                  & 39.70                        \\
                                                        & DeepSDF \cite{deepsdf}                                                & 263.92                 & 19.83                  & 77.95                  & 40.95                        \\\cline{1-2}
    \multirow{2}{*}{\makecell{Local\\Learning}}         & LIG \cite{LIG}                                                        & 48.75                  & 83.76                  & 92.57                  & 81.48                        \\
                                                        & Points2Surf \cite{points2surf}                                        & 48.93                  & 80.89                  & 89.52                  & 81.83                        \\\cline{1-2}
                                                        & DSE \cite{learningdelaunaysurface}                                    & 32.16                  & 86.88                  & 87.20                  & 76.81                        \\
    \multirow{2}{*}{\makecell{Hybird}}                  & IMLSNet \cite{liu2021DeepIMLS}                                        & 38.46                  & 82.44                  & 93.31                  & 85.30                        \\
                                                        & ParseNet \cite{parsenet}                                              & 149.96                 & 38.92                  & 81.51                  & 45.67                        \\\hline
\end{tabular}%
}
\vspace{-0.08in}
\end{table}

Quantitative results of comparative methods are given in Tables \ref{tab:syn_obj_whole_chellenges_mid_whole_mectrics}, \ref{tab:syn_scene}, and \ref{tab:real_obj}, which are respectively for synthetic data of object surfaces, synthetic data of scene surfaces, and real-scanned data. Qualitative results are presented in the following sections accompanying our discussions.

\subsection{The Remaining Challenges}
\label{SecInsightMissingDataMisalignment}

\begin{figure}[hptb]
    \begin{minipage}{0.49\linewidth}
        \centering
        \includegraphics[width=0.99\textwidth]{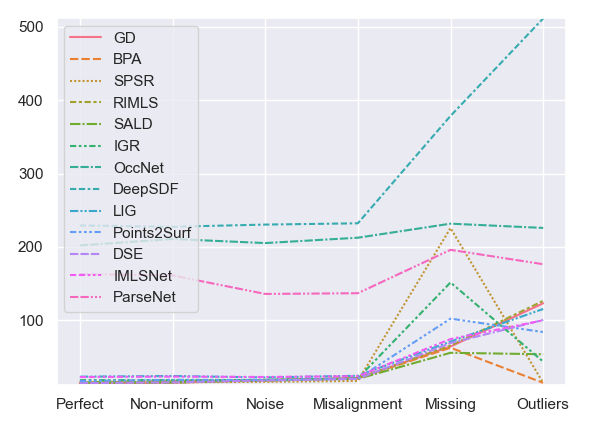}
        \subcaption{CD}
    \end{minipage}
    \begin{minipage}{0.49\linewidth}
        \centering
        \includegraphics[width=0.99\textwidth]{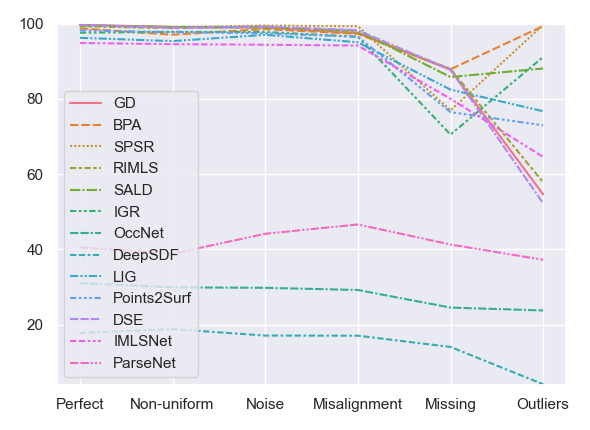}
        \subcaption{F-score}
    \end{minipage}\\
    \vspace{-0.05in}
    \begin{minipage}{0.49\linewidth}
        \includegraphics[width=0.99\textwidth]{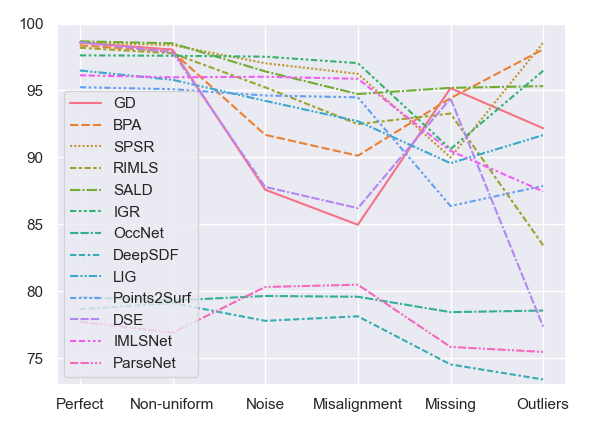}
        \subcaption{NCS}
    \end{minipage}
    \begin{minipage}{0.49\linewidth}
        \includegraphics[width=0.99\textwidth]{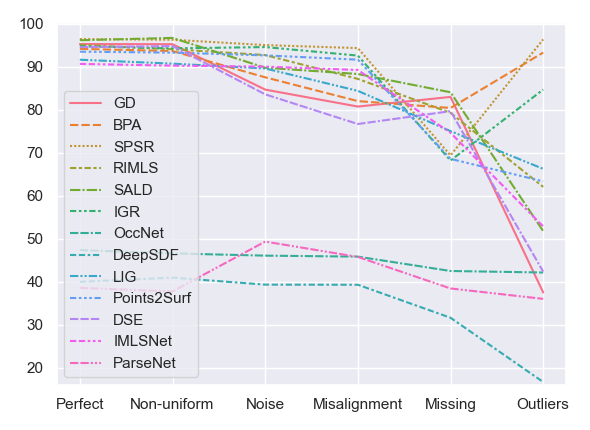}
        \subcaption{NFS}
    \end{minipage}
    \vspace{-0.1cm}
    \caption{
        Plotting of quantitative results in \cref{tab:syn_obj_whole_chellenges_mid_whole_mectrics}.
    }
    \vspace{-0.6cm}
    \label{fig:result_middlevel}
\end{figure}

\begin{figure*}[hptb]
    \centering
    \vspace{-0.6cm}
    \begin{minipage}{0.94\linewidth}
        \centering
        \includegraphics[width=0.94\textwidth]{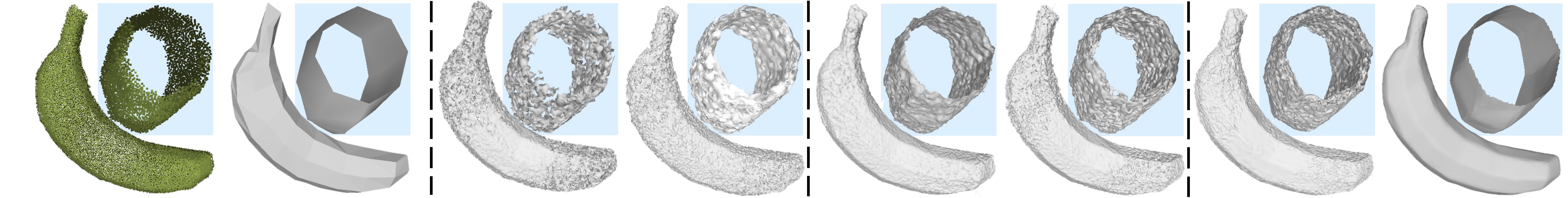}
        \scalebox{0.67}{ \begin{tabular}{p{70pt}<{\centering}p{70pt}<{\centering}p{75pt}<{\centering}p{75pt}<{\centering}p{70pt}<{\centering}p{70pt}<{\centering}p{70pt}<{\centering}p{70pt}<{\centering}} \small Input PC & \    \small GT & \   \small GD \cite{greedydelaunay} & \     \small BPA \cite{BPA} &  \       \small SPSR \cite{kazhdan2013screened} &   \       \small RIMLS \cite{RIMLS}  &  \    \small SALD \cite{sald} &  \       \small IGR \cite{gropp2020implicitgeometricregularizationforlearningshape}    \\          \end{tabular}}
    \end{minipage}
    \vspace{-0.03in}

    \begin{minipage}{0.94\linewidth}
        \centering
        \includegraphics[width=0.94\textwidth]{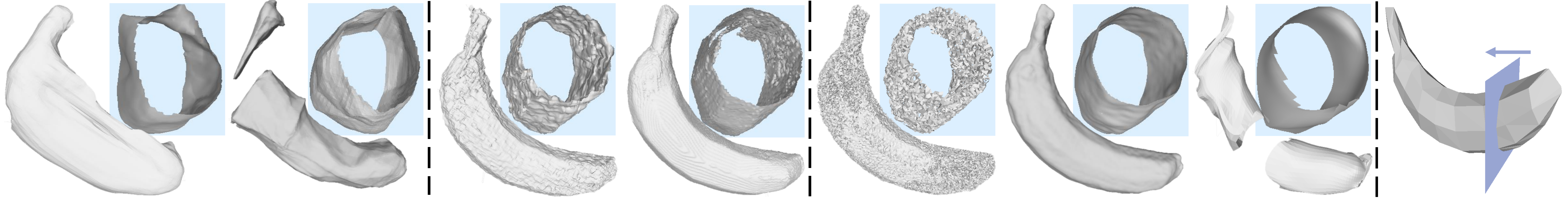}
        \scalebox{0.67}{\begin{tabular}{p{70pt}<{\centering}p{70pt}<{\centering}p{75pt}<{\centering}p{75pt}<{\raggedright}p{70pt}<{\centering}p{70pt}p{70pt}<{\centering}p{70pt}<{\centering}}  \small OccNet \cite{occupancy}  &  \       \small DeepSDF \cite{deepsdf}  &  \    \small LIG \cite{LIG} & \       \small Points2Surf \cite{points2surf} &  \         \small DSE \cite{learningdelaunaysurface} & \ \small IMLSNet \cite{liu2021DeepIMLS}  &  \       \small ParseNet \cite{parsenet} &  \       \small View Direction \\            \end{tabular}}
    \end{minipage}
    \vspace{-0.2cm}
    \caption{
        An example of qualitative results from different methods when dealing with the challenge of \textit{misalignment}.
    }
    \label{fig:misalignment}
\end{figure*}

\begin{figure*}[hptb]
    \centering
    \vspace{-0.6cm}
    \begin{minipage}{0.94\linewidth}
        \centering
        \includegraphics[width=0.94\textwidth]{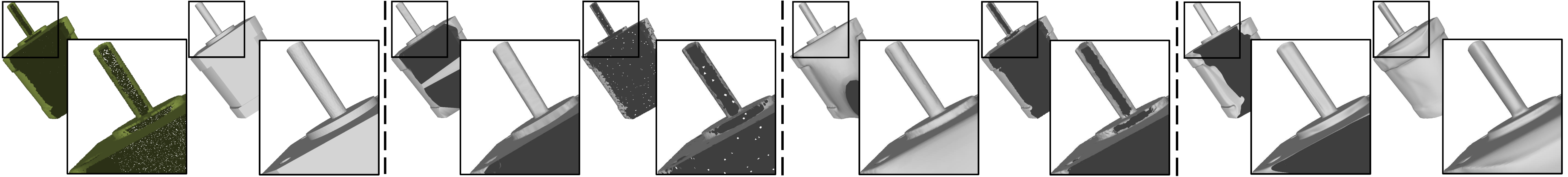}
        \scalebox{0.67}{ \begin{tabular}{p{70pt}<{\centering}p{70pt}<{\centering}p{75pt}<{\centering}p{75pt}<{\centering}p{70pt}<{\centering}p{70pt}<{\centering}p{70pt}<{\centering}p{70pt}<{\centering}} \small Input PC & \    \small GT & \   \small GD \cite{greedydelaunay} & \     \small BPA \cite{BPA} &  \       \small SPSR \cite{kazhdan2013screened} &   \       \small RIMLS \cite{RIMLS}  &  \    \small SALD \cite{sald} &  \       \small IGR \cite{gropp2020implicitgeometricregularizationforlearningshape}    \\          \end{tabular}}
    \end{minipage}
    \vspace{-0.03in}
    \begin{minipage}{0.94\linewidth}
        \centering
        \includegraphics[width=0.94\textwidth]{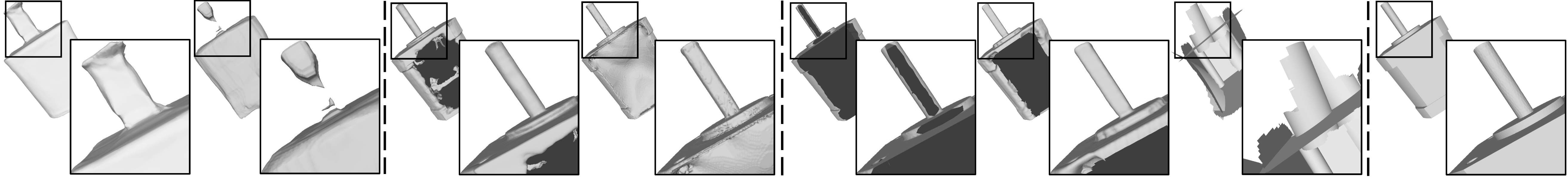}
        \scalebox{0.67}{\begin{tabular}{p{70pt}<{\centering}p{70pt}<{\centering}p{75pt}<{\centering}p{75pt}<{\raggedright}p{70pt}<{\centering}p{70pt}p{70pt}<{\centering}p{70pt}<{\centering}}  \small OccNet \cite{occupancy}  &  \       \small DeepSDF \cite{deepsdf}  &  \    \small LIG \cite{LIG} & \       \small Points2Surf \cite{points2surf} &  \         \small DSE \cite{learningdelaunaysurface} & \ \small IMLSNet \cite{liu2021DeepIMLS}  &  \       \small ParseNet \cite{parsenet} &  \       \small GT \\            \end{tabular}}
    \end{minipage}
    \vspace{-0.2cm}
    \caption{
    An example of qualitative results from different methods when dealing with the challenge of \textit{missing points}.
    }
    \label{fig:missing}
\end{figure*}

\begin{figure*}[hptb]
    \centering
    \vspace{-0.6cm}
    \begin{minipage}{0.94\textwidth}
        \centering
        \includegraphics[width=0.94\textwidth]{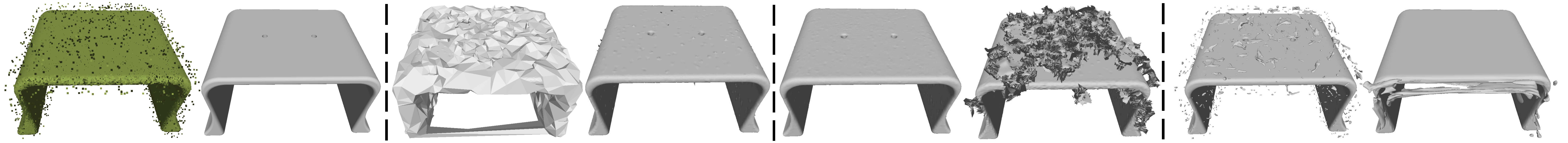}
        \scalebox{0.67}{ \begin{tabular}{p{70pt}<{\centering}p{70pt}<{\centering}p{75pt}<{\centering}p{75pt}<{\centering}p{70pt}<{\centering}p{70pt}<{\centering}p{70pt}<{\centering}p{70pt}<{\centering}} \small Input PC & \    \small GT & \   \small GD \cite{greedydelaunay} & \     \small BPA \cite{BPA} &  \       \small SPSR \cite{kazhdan2013screened} &   \       \small RIMLS \cite{RIMLS}  &  \    \small SALD \cite{sald} &  \       \small IGR \cite{gropp2020implicitgeometricregularizationforlearningshape}    \\          \end{tabular}}
    \end{minipage}
    \vspace{-0.03in}
    \begin{minipage}{0.94\textwidth}
        \centering
        \includegraphics[width=0.94\textwidth]{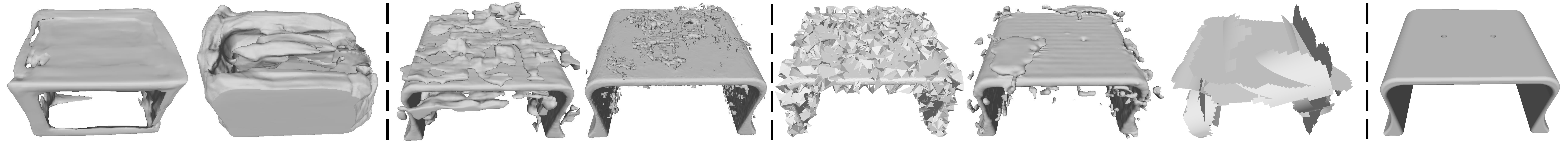}
        \scalebox{0.67}{\begin{tabular}{p{70pt}<{\centering}p{70pt}<{\centering}p{75pt}<{\centering}p{75pt}<{\raggedright}p{70pt}<{\centering}p{70pt}p{70pt}<{\centering}p{70pt}<{\centering}}  \small OccNet \cite{occupancy}  &  \       \small DeepSDF \cite{deepsdf}  &  \    \small LIG \cite{LIG} & \       \small Points2Surf \cite{points2surf} &  \         \small DSE \cite{learningdelaunaysurface} & \ \small IMLSNet \cite{liu2021DeepIMLS}  &  \       \small ParseNet \cite{parsenet} &  \       \small GT \\            \end{tabular}}
    \end{minipage}
    \vspace{-0.2cm}
    \caption{
        An example of qualitative results from different methods when dealing with the challenge of \textit{point outliers}.
    }
\label{fig:outlier}
\end{figure*}

For ease of analysis, we plot in \cref{fig:result_middlevel} the quantitative results in \cref{tab:syn_obj_whole_chellenges_mid_whole_mectrics} under the metrics of CD, F-score, NCS, and NFS. As presented in \cref{Exp:metrics}, the four metrics focus on different measure perspectives. By diagnosing the comparative methods using these measures and investigating their capabilities to cope with different challenges of imperfect scanning, \cref{fig:result_middlevel} helps in identifying the remaining challenges.

\cref{fig:result_middlevel} shows that, under all the metrics, the challenge of \emph{non-uniform distribution of points} is relatively easy to be tackled by almost all the methods, except those learning semantics or geometric primitives (i.e., DeepSDF \cite{deepsdf}, OccNet \cite{occupancy}, and ParseNet \cite{parsenet}), achieving similar results as those on data of perfect scanning. The discussion on why some learning-based methods fail to generalize is given in \cref{sec:learning_based_leanring_free}. For the challenge of \emph{point-wise noise}, though the overall shape structures (measured by CD and F-Score) can be roughly recovered by most of the methods (again, except some learning-based ones), the reconstructions might be short of surface details, as verified by the reduced scores of NCS and NFS, especially for triangulation-based methods such as GD \cite{greedydelaunay}, BPA \cite{BPA}, and the hybrid one of DSE \cite{learningdelaunaysurface} that combines the triangulation-based prior. This is intuitive since methods based on triangulation of points rely heavily on cleanness of input points.

There exists a similar but severer phenomenon for the challenge of \emph{misalignment}. As indicated by \cref{fig:result_middlevel}, when measured by CD and F-score, most of the methods give reasonably good results, suggesting that the overall shape structures have been recovered. However, the reduced scores of NCS and NFS suggest that some of the recovered surfaces might be less pleasant to human perception. In fact, as shown by the example in \cref{fig:misalignment}, most methods fail in reconstructing the surface on the misaligned areas, generating thickened or even multiple layers of the local surface. Misalignment is a practical issue in 3D scanning, especially when using hand-held, consumer scanners. \cref{fig:result_middlevel} and \cref{fig:misalignment} show that methods using smoothness and/or modeling priors have the advantage in handling misalignment.

\cref{fig:result_middlevel} also shows that the two challenges of \emph{missing points} and \emph{point outliers} are much more difficult to be handled. When an input point cloud has missing points, as shown in \cref{fig:missing}, most of the methods ignore reconstruction of the missing surface areas, resulting in incomplete surfaces. The implicit methods (e.g., IGR \cite{gropp2020implicitgeometricregularizationforlearningshape}, OccNet \cite{occupancy}, and Points2Surf \cite{points2surf}) tend to generate watertight surfaces by filling the holes with concave/convex hulls; however, such envelopes may not represent the surface correctly, possibly making the reconstruction even poorer under all the evaluation metrics (cf. \cref{fig:result_middlevel}).

As for \emph{point outliers}, although a pre-processing step of outlier removal has already been adopted for all the comparative methods (cf. \cref{SecPreprocessing}), performance of different methods still varies drastically. \cref{fig:outlier} gives an example; the results depend on whether the respective methods have their inbuilt mechanisms of outlier removal. For example, BPA \cite{BPA} requires the sizes of its triangular faces to satisfy certain conditions, making it naturally suitable for handling outliers; methods using a global implicit field (e.g., SPSR \cite{SPSS} and IGR \cite{gropp2020implicitgeometricregularizationforlearningshape}) ignore the outliers implicitly, and are thus capable of handling point outliers as well.

Summarizing the above analyses gives us the following empirical findings: (1) the challenge of \emph{missing points} remains unsolved by all the comparative methods; (2) for the challenges of \emph{misalignment} and \emph{point outliers}, most of the methods (except few ones such as SPSR \cite{SPSS}) give unsatisfactory results; (3) there exist inconsistencies between different evaluation metrics, and in many cases, good quantitative results do not translate as visually pleasant ones; (4) methods that learn semantics or pre-defined shape patterns may fail to generalize even on clean data of perfect scanning, when the testing point clouds do not fall in the learned data domains; we will discuss more on this issue shortly.

\subsection{Optimization-based, Learning-free Methods \emph{Versus} Learning-based, Data-driven Ones}\label{sec:learning_based_leanring_free}

\begin{table*}[hptb]
\centering
\caption{
Comparison between optimization-based, learning-free methods and learning-based, data-driven methods on the testing synthetic data of object surfaces. 
We report quantitative results for the imperfect scanning of \emph{point-wise noise} at three levels of severity; results are in the format of ``$\cdot$ / $\cdot$ / $\cdot$'', where the most left one is the absolute value under each evaluation metric, and the right two ones are those relative to the most left one. 
The \textbf{best} and \uline{\textbf{second best}} methods are highlighted in each column. 
}
\vspace{-0.1in}
\label{tab:noise}
\resizebox{0.94\textwidth}{!}{
\begin{tabular}{l|c|r@{ /}r@{ / }c|r@{ /}r@{ / }c|r@{ /}r@{ / }c|r@{ /}r@{ / }c}
\hline
\multirow{2}{*}{\makecell{Algorithms}} & \multirow{2}{*}{\makecell{Priors}} & \multicolumn{3}{c|}{CD $ (\times10^{-4}) \downarrow$} & \multicolumn{3}{c|}{F-score $(\%) \uparrow$} & \multicolumn{3}{c|}{NCS $ (\times10^{-2}) \uparrow$} & \multicolumn{3}{c}{NFS $ (\times10^{-2}) \uparrow$} \\\cline{3-14}
~ & ~ & \emph{\makecell{low-}} & \emph{\makecell{middle-}}  & \emph{\makecell{high-\\level}} &\emph{\makecell{low-}} & \emph{\makecell{middle-}} & \emph{\makecell{high-\\level}} & \emph{\makecell{low-}} & \emph{\makecell{middle-}} & \emph{\makecell{high-\\level}} & \emph{\makecell{low-}} & \emph{\makecell{middle-}} & \emph{\makecell{high-\\level}} \\\hline
GD \cite{greedydelaunay}                                               &   \multirow{6}{*}{\makecell{learning-\\free}}         & 16.52                        & 1.68                          & 13.21                         & 99.09                        & -0.18                      & -7.44                         & 94.00                        & -6.42                       & -25.68                      & 91.57                       & -6.94                      & -35.27                   \\ 
BPA \cite{BPA}                                                         &                                                       & 16.59                        & 1.99                          & 12.71                         & 98.63                        & -0.12                      & -9.04                         & 95.36                        & -3.68                       & -17.39                      & 90.71                       & -3.22                      & -25.56                   \\ 
SPSR \cite{kazhdan2013screened}                                        &                                                       & \uline{\textbf{15.50}}       & 0.55                          & 2.66                          & \uline{\textbf{99.51}}       & -0.05                      & \uline{\textbf{-0.35}}        & \uline{\textbf{97.84}}       & -0.81                       & -3.95                       & \uline{\textbf{95.60}}      & -0.62                      & -3.61                    \\               
RIMLS \cite{RIMLS}                                                     &                                                       & 16.13                        & 1.04                          & 9.75                          & 99.36                        & -0.12                      & -4.67                         & 97.36                        & -2.13                       & -11.23                      & 94.19                       & -1.52                      & -14.43                   \\ 
SALD \cite{sald}                                                       &                                                       & \textbf{15.33}               & 3.44                          & 12.26                         & \textbf{99.54}               & -0.60                      & -7.06                         & \textbf{98.07}               & -1.65                       & -9.16                       & \textbf{95.66}              & -5.94                      & -28.48                   \\                                 
IGR \cite{gropp2020implicitgeometricregularizationforlearningshape}    &                                                       & 18.21                        & 0.36                          & 0.99                          & 97.87                        & -0.12                      & \uline{\textbf{-0.35}}        & 97.64                        & -0.12                       & -0.54                       & 94.37                       & 0.15                       & -0.70                    \\ \hline  
OccNet \cite{occupancy}                                                &   \multirow{7}{*}{\makecell{learning-\\based}}        & 209.04                       & -3.83                         & 0.97                          & 29.85                        & -0.08                      & -1.93                         & 79.43                        & 0.21                        & -0.38                       & 46.17                       & -0.14                      & -1.13                    \\
DeepSDF \cite{deepsdf}                                                 &                                                       & 241.28                       & \uline{\textbf{-10.88}}       & \uline{\textbf{-1.65}}        & 16.70                        & \uline{\textbf{0.38}}      & -0.71                         & 78.01                        & -0.23                       & \uline{\textbf{0.51}}       & 38.55                       & 0.74                       & \uline{\textbf{0.15}}    \\
LIG \cite{LIG}                                                         &                                                       & 23.96                        & -1.91                         & 2.31                          & 94.50                        & \textbf{2.49}              & -0.70                         & 93.96                        & \uline{\textbf{0.26}}       & -5.70                       & 86.71                       & \uline{\textbf{2.84}}      & -5.68                    \\
Points2Surf \cite{points2surf}                                         &                                                       & 17.74                        & 0.74                          & 4.89                          & 98.02                        & -0.63                      & -3.81                         & 94.97                        & -0.35                       & -1.53                       & 92.83                       & -0.24                      & -2.62                    \\               
DSE \cite{learningdelaunaysurface}                                     &                                                       & 16.07                        & 1.82                          & 11.54                         & 99.40                        & -0.23                      & -6.65                         & 94.43                        & -6.64                       & -23.04                      & 90.05                       & -6.52                      & -32.97                   \\                                                                    
IMLSNet \cite{liu2021DeepIMLS}                                         &                                                       & 22.64                        & 0.03                          & 1.09                          & 94.41                        & -0.05                      & \textbf{-0.01}                & 96.08                        & -0.06                       & -0.46                       & 90.13                       & -0.16                      & -0.47                    \\
ParseNet \cite{parsenet}                                               &                                                       & 154.11                       & \textbf{-18.27}               & \textbf{-14.14}               & 44.80                        & -0.67                      & -2.14                         & 78.55                        & \textbf{1.76}               & \textbf{2.60}               & 46.21                       & \textbf{3.09}              & \textbf{2.79}            \\\hline                                        
\end{tabular}%
}
\end{table*}

\begin{figure*}[htbp]
    \vspace{-0.2in}
    \begin{minipage}{0.05\linewidth}
        \centering
        \scalebox{1.0}{\begin{tabular}{p{50pt}}
            \rotatebox{90}{\small learning-free} \\[45pt]
        \end{tabular}}
    \end{minipage}\begin{minipage}{0.94\linewidth}
        \centering
        \includegraphics[width=0.98\textwidth]{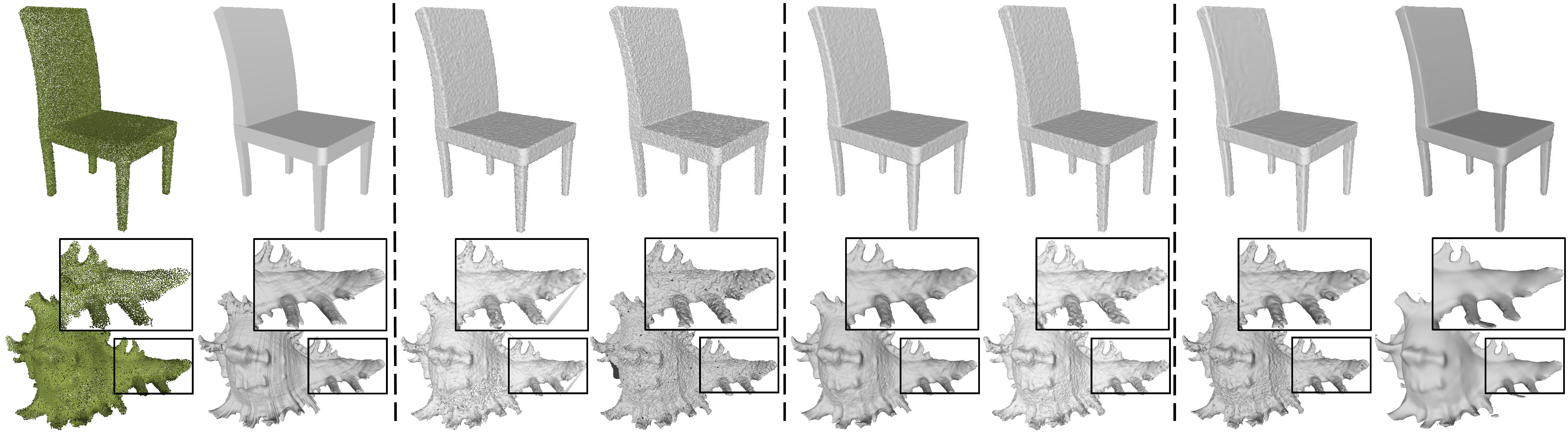}
        \scalebox{0.70}{ \begin{tabular}{p{70pt}<{\centering}p{70pt}<{\centering}p{75pt}<{\centering}p{75pt}<{\centering}p{70pt}<{\centering}p{70pt}<{\centering}p{70pt}<{\centering}p{70pt}<{\centering}} \small Input PC & \    \small GT & \   \small GD \cite{greedydelaunay} & \     \small BPA \cite{BPA} &  \       \small SPSR \cite{kazhdan2013screened} &   \       \small RIMLS \cite{RIMLS}  &  \    \small SALD \cite{sald} &  \       \small IGR \cite{gropp2020implicitgeometricregularizationforlearningshape}    \\          \end{tabular}}
    \end{minipage}
    \begin{minipage}{0.05\linewidth}
        \centering
        \scalebox{1.0}{\begin{tabular}{p{50pt}}
            \rotatebox{90}{\small learning-based} \\[45pt]
        \end{tabular}}
    \end{minipage}
    \begin{minipage}{0.94\linewidth}
        \centering
        \includegraphics[width=0.98\textwidth]{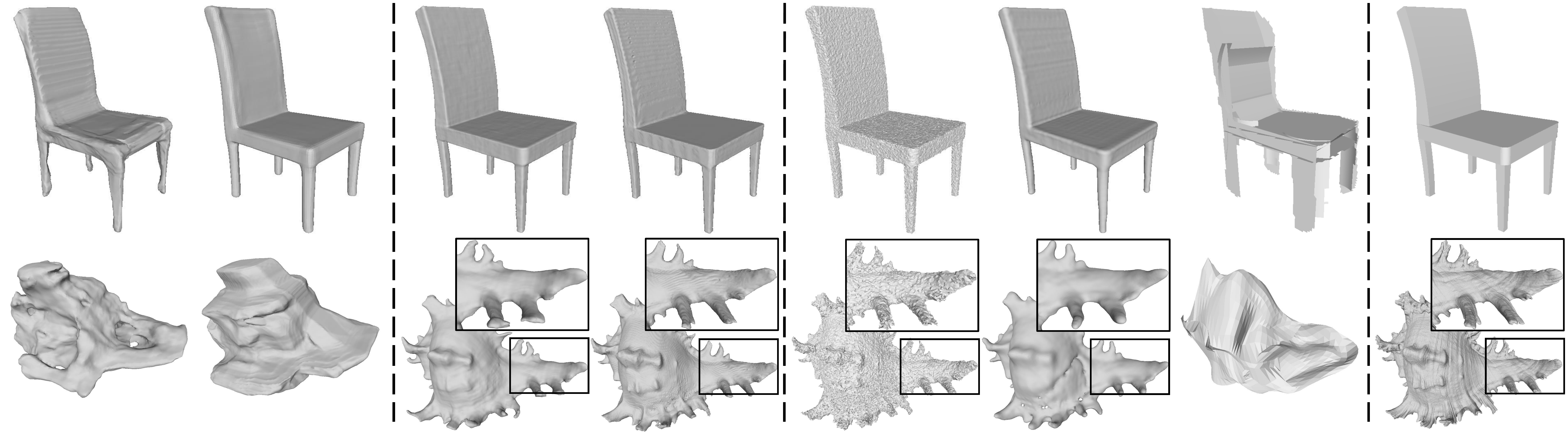}
        \vspace{-0.05in}
        \scalebox{0.70}{\begin{tabular}{p{70pt}<{\centering}p{70pt}<{\centering}p{75pt}<{\centering}p{75pt}<{\raggedright}p{70pt}<{\centering}p{70pt}p{70pt}<{\centering}p{70pt}<{\centering}}  \small OccNet \cite{occupancy}  &  \       \small DeepSDF \cite{deepsdf}  &  \    \small LIG \cite{LIG} & \       \small Points2Surf \cite{points2surf} &  \         \small DSE \cite{learningdelaunaysurface} & \ \small IMLSNet \cite{liu2021DeepIMLS}  &  \       \small ParseNet \cite{parsenet} &  \       \small GT \\            \end{tabular}}
    \end{minipage}
    \vspace{-0.1cm}
    \caption{Qualitative results respectively from optimization-based, learning-free methods and learning-based, data-driven methods on the testing synthetic data of object surfaces. The input point clouds have imperfect scanning of \emph{point-wise noise}. The example of \emph{chair} is of popular semantic categories, for which our auxiliary training set contains rich instances of the same category from ShapeNet \cite{shapenet}; the other one is non-semantic.
    }
    \label{fig:exp_wnoi_sem_nonsem}
\end{figure*}

\begin{figure*}[htbp]
    \vspace{-0.2in}
    \begin{minipage}{0.05\linewidth}
        \centering
        \scalebox{0.7}{\begin{tabular}{p{50pt}}
            \rotatebox{90}{\small learning-free} \\[20pt]
        \end{tabular}}
    \end{minipage}
    \begin{minipage}{0.95\linewidth}
        \centering
        \includegraphics[width=0.99\textwidth]{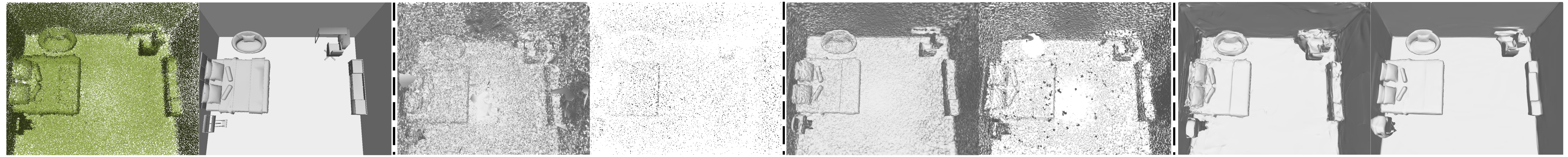}
        \scalebox{0.73}{ \begin{tabular}{p{70pt}<{\centering}p{70pt}<{\centering}p{75pt}<{\centering}p{75pt}<{\centering}p{70pt}<{\centering}p{70pt}<{\centering}p{70pt}<{\centering}p{70pt}<{\centering}} \small Input PC & \    \small GT & \   \small GD \cite{greedydelaunay} & \     \small BPA \cite{BPA} &  \       \small SPSR \cite{kazhdan2013screened} &   \       \small RIMLS \cite{RIMLS}  &  \    \small SALD \cite{sald} &  \       \small IGR \cite{gropp2020implicitgeometricregularizationforlearningshape}    \\          \end{tabular}}
    \end{minipage}
    \begin{minipage}{0.05\linewidth}
        \centering
        \scalebox{0.7}{\begin{tabular}{p{50pt}}
            \rotatebox{90}{\small learning-based} \\[20pt]
        \end{tabular}}
    \end{minipage}
    \begin{minipage}{0.95\linewidth}
        \centering
        \includegraphics[width=0.99\textwidth]{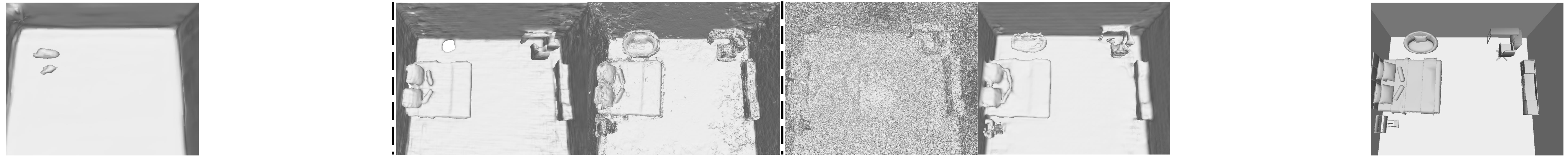}
        \vspace{-0.05in}
        \scalebox{0.73}{\begin{tabular}{p{70pt}<{\centering}p{70pt}<{\centering}p{75pt}<{\centering}p{75pt}<{\raggedright}p{70pt}<{\centering}p{70pt}p{70pt}<{\centering}p{70pt}<{\centering}}  \small OccNet \cite{occupancy}  &  \       \small DeepSDF \cite{deepsdf}  &  \    \small LIG \cite{LIG} & \       \small Points2Surf \cite{points2surf} &  \         \small DSE \cite{learningdelaunaysurface} & \ \small IMLSNet \cite{liu2021DeepIMLS}  &  \       \small ParseNet \cite{parsenet} &  \       \small GT \\            \end{tabular}}
    \end{minipage}
    \vspace{-0.1cm}
    \caption{
        Qualitative results respectively from optimization-based, learning-free methods and learning-based, data-driven methods on the testing synthetic data of scene surfaces. The scene is a bedroom instance from the 3D-FRONT \cite{3dfrontdata}.}
    \label{fig:exp_scene}
\end{figure*}

In this section, we investigate the behaviors of comparative methods by organizing them into two groups of \emph{optimization-based, learning-free methods} and \emph{learning-based, data-driven ones}. The former group includes SPSR \cite{kazhdan2013screened}, RIMLS \cite{RIMLS}, SALD \cite{sald}, IGR \cite{gropp2020implicitgeometricregularizationforlearningshape}, and we also include GD \cite{greedydelaunay} and BPA \cite{BPA} into the group for a complete coverage of the studied methods; the latter group includes OccNet \cite{occupancy}, DeepSDF \cite{deepsdf}, LIG \cite{LIG}, Points2Surf \cite{points2surf}, DSE \cite{learningdelaunaysurface}, IMLSNet \cite{liu2021DeepIMLS}, and ParseNet \cite{parsenet}.
By doing so, we aim to investigate how the two groups perform in terms of generalizing to complex shapes, where we pay special attention to methods of global, semantic learning (i.e., OccNet \cite{occupancy} and DeepSDF \cite{deepsdf}) whose advantages may only be manifested when the categories of object surfaces exist in the auxiliary training set.
Note that our testing data of synthetic object surfaces include 22 randomly selected objects, as described in Section \ref{SecExpData}, which contain both those belonging to popular semantic categories (e.g., \emph{chair}) and those without clearly defined semantics; for the former case, our auxiliary training set contains rich shape instances of same categories from ShapeNet \cite{shapenet}. In this section, we also study robustness of the two groups of methods against imperfect scanning at varying levels of severity.

Quantitative results in \cref{tab:noise} show that on reconstruction of synthetic object surfaces, optimization-based, learning-free methods generalize better under different evaluation metrics, since our testing shapes contain both semantic ones and non-semantic, complex ones; we have consistent observations from examples in \cref{fig:exp_wnoi_sem_nonsem} --- while learning-based, data-driven methods are good at reconstructing an \emph{chair} surface, they fail in generalizing to non-semantic shapes.
\cref{tab:noise} also show that learning-based methods are good in terms of robustness against higher levels of data imperfections.
We observe similar phenomena on reconstruction of real-scanned data (cf. Appendix \ref{sec:more_exp_real} for more details), by re-organizing \cref{tab:real_obj} according to the two method groups.

To further investigate the behaviors of learning from auxiliary data, we conduct experiments of synthetic scene surface reconstruction. Results in \cref{tab:syn_scene} (after re-organization of method groups) and \cref{fig:exp_scene} show that among the learning-based methods, LIG \cite{LIG}, Points2Surf \cite{points2surf} and IMLSNet \cite{liu2021DeepIMLS} are capable of handling reconstruction of scene-level surfaces via local modeling and aggregation, while semantic learning methods of OccNet and DeepSDF fail, as expected.

We summarize the above analyses as follows: (1) learning-based, data-driven methods are able to reconstruct object surfaces when the training set contains object instances of the same semantic categories, and they show a certain degree of robustness against data imperfections; however, these methods fail to generalize when the condition is not satisfied; (2) for reconstruction of scene-level surfaces, local learning methods succeed by local modeling and aggregation, and in contrast, global, semantic learning methods fail to do so; (3) some optimization-based, learning-free methods (e.g., SPSR \cite{kazhdan2013screened}) perform surprisingly well in robustness and generalization for both object-level and scene-level surfaces.

\begin{table}[hptb]
    \centering
    \caption{Comparison on testing synthetic data of object surfaces among methods without using surface normals ($\times$), methods using surfaces normals ($\checkmark$), and methods using surface normals only during learning ($*$). We report quantitative results for the imperfect scanning of \emph{point-wise noise} at the middle level of severity; results are in the format of ``$\cdot$ / $\cdot$'', where the left one is obtained assuming the availability of ground-truth camera poses, and the right one is obtained without knowing the camera poses. The \textbf{best} and \uline{\textbf{second best}} methods are highlighted in each column.
    }
    \vspace{-0.1in}
    \label{tab:oriented_normal_comparing}
    \resizebox{\linewidth}{!}{
    \begin{tabular}{l|c|c|c|c|c} \hline
    \multirow{1}{*}{Algorithms} & \multirow{1}{*}{Normals} & \multicolumn{1}{c|}{CD $(\times10^{-4}) \downarrow$} & \multicolumn{1}{c|}{F-score $(\%) \uparrow$} & \multicolumn{1}{c|}{NCS $(\times10^{-2}) \uparrow$} & \multicolumn{1}{c}{NFS $(\times10^{-2}) \uparrow$} \\\hline
    GD \cite{greedydelaunay}                                                &  \multirow{3}{*}{$\times$}      & 18.20                        / {\ \ \ \ \ \ \ }  & 98.91                        / {\ \ \ \ \ \ \ }  & 87.58                        / {\ \ \ \ \ \ \ } & 84.63                    / {\ \ \ \ \ \ \ }       \\
    SALD \cite{sald}                                                        &                                 & 18.77                        / {\ \ \ \ \ \ \ }  & 98.94                        / {\ \ \ \ \ \ \ }  & 96.42                        / {\ \ \ \ \ \ \ } & 89.72                    / {\ \ \ \ \ \ \ }       \\
    DSE \cite{learningdelaunaysurface}                                      &                                 & 17.89                        / {\ \ \ \ \ \ \ }  & 99.17                        / {\ \ \ \ \ \ \ }  & 87.79                        / {\ \ \ \ \ \ \ } & 83.53                    / {\ \ \ \ \ \ \ }       \\ \hline
    BPA \cite{BPA}                                                          &  \multirow{7}{*}{$\checkmark$}  & 18.58                        / 18.61             & 98.51                        / 98.56             & 91.68                        / 91.79            & 87.49                    / 85.08                  \\
    SPSR \cite{kazhdan2013screened}                                         &                                 & \textbf{16.05}               / 60.42             & \textbf{99.46}               / 91.45             & \uline{\textbf{97.03}}       / 94.68            & \textbf{94.98}           / 84.34                  \\
    RIMLS \cite{RIMLS}                                                      &                                 & \uline{\textbf{17.17}}       / 18.58             & \uline{\textbf{99.24}}       / 97.47             & 95.23                        / 94.06            & 92.67                    / 86.63                  \\
    IGR \cite{gropp2020implicitgeometricregularizationforlearningshape}     &                                 & {\ } 18.57                   / 262.40            & 97.75                        / 85.98             & \textbf{97.52}               / 95.28            & \uline{\textbf{94.52}}   / 79.61                  \\
    OccNet \cite{occupancy}                                                 &                                 & 205.21                       / 214.12            & 29.77                        / 28.83             & 79.64                        / 79.01            & 46.03                    / 45.12                  \\
    DeepSDF \cite{deepsdf}                                                  &                                 & 230.40                       / 569.41            & 17.08                        / 15.64             & 77.78                        / 77.09            & 39.29                    / 33.78                  \\
    LIG \cite{LIG}                                                          &                                 & 22.05                        / 34.78             & 96.99                        / 89.60             & 94.22                        / 91.36            & 89.55                    / 80.02                  \\
    ParseNet \cite{parsenet}                                                &                                 & 135.84                       / 197.28            & 44.13                        / 35.67             & 80.31                        / 76.19            & 49.30                    / 41.65                  \\ \hline
    Points2Surf \cite{points2surf}                                          &  \multirow{2}{*}{$*$}           & 18.48                        / {\ \ \ \ \ \ \ }  & 97.39                        / {\ \ \ \ \ \ \ }  & 94.62                        / {\ \ \ \ \ \ \ } & 92.59                    / {\ \ \ \ \ \ \ }       \\
    IMLSNet \cite{liu2021DeepIMLS}                                          &                                 & 22.67                        / {\ \ \ \ \ \ \ }  & 94.36                        / {\ \ \ \ \ \ \ }  & 96.02                        / {\ \ \ \ \ \ \ } & 89.97                    / {\ \ \ \ \ \ \ }       \\ \hline
    \end{tabular}%
    }
    \vspace{-0.5cm}
    \end{table}

\subsection{The Importance of Orientations of Surface Normals}\label{sec:without_or_need_normal}

Some of our studied methods compute surface normals from observed point clouds, and use the computed normals for surface reconstruction, including BPA \cite{BPA}, SPSR \cite{kazhdan2013screened}, RIMLS \cite{RIMLS}, IGR \cite{gropp2020implicitgeometricregularizationforlearningshape}, OccNet \cite{occupancy}, DeepSDF \cite{deepsdf}, LIG \cite{LIG}, and ParseNet \cite{parsenet}; a few other methods (e.g., Points2Surf \cite{points2surf} and IMLSNet \cite{liu2021DeepIMLS}) compute surface normals on training data, and then train models to estimate surface normals when reconstructing testing surfaces. To investigate how these methods benefit from surface normal computation/estimation, we conduct experiments on our testing data with scanning imperfections, since robustness of these methods would be tested when less accurate surface normals are computed from imperfectly scanned data. Assume that the relative pose of a camera w.r.t. an observed point cloud is given; for any observed point, we compute its \emph{oriented} surface normal by performing PCA on its local neighborhood of points (cf. \cref{datasysthesizing} for the details); as such, the computed surface normals may not be precise but their \emph{inward or outward orientations} must be correct.

\cref{tab:oriented_normal_comparing} gives the quantitative results for synthetic data of object surfaces; under different evaluation metrics, the additional computation or estimation of surface normals does help in improving surface reconstruction. When the relative pose of camera w.r.t. observed surface points is not available, the computed surface normals would be \emph{wrongly oriented}  at some local surface neighborhoods, since principal directions of PCA on local neighborhoods of noisy points are less reliable. To investigate the importance of inward or outward orientations of surface normals, we also report experiments in \cref{tab:oriented_normal_comparing} where each result on the right of the ``/'' symbol is obtained without knowing the relative camera poses; compared with results on the left side of ``/'' that are obtained assuming the ground-truth camera poses, results on the right side drops drastically. We also conduct experiments on our synthetic data of scene surfaces and real-scanned data, and observe similar phenomena; these results are presented in Appendix \ref{sec:more_exp_normals}.

We have following empirical findings based on the above analyses: (1) computation or estimation of oriented surface normals help in surface reconstruction from point clouds; (2) compared with precisions of surface normals, it is more important to have the correct inward or outward orientations of surface normals.

\section{Conclusion}\label{sec:discuss}
In this paper, we have reviewed both the classical and the more recent deep learning-based methods for surface reconstruction from point clouds; we have organized our reviews by categorizing these methods according to what priors of surface geometry they had used to regularize their solutions.
To better understand the respective strengths and limitations of existing methods, we contribute a large-scale benchmarking dataset consisting of both synthetic and real-scanned data, which provides various sensing imperfections that are commonly encountered in practical 3D scanning.
We conduct thorough empirical studies on the constructed benchmark, evaluating the robustness and generalization of different methods.
Our studies help identify the remaining challenges faced by existing methods, and we expect that our studies would be useful for guiding the directions in future research.

\bibliographystyle{IEEEtran}
\bibliography{IEEEabrv,paper}

\clearpage
\appendices
\section{Technical Details for Computing the Algebraic Surface Complexity}
\label{sec:algebraiccomplexity}
As discussed in \cref{sec:hierarchical}, we have divided all the object surfaces into different groups according to their algebraic complexities measured by the averaged approximation errors, which are computed using a highest-degree fixed function to fit the local surface patches. We present the details in this section.

Assume that the underlying surface $\mathcal{S}^{*}$ of a given point cloud $\mathcal{P}$ is $\mathcal{C}^{\infty}$ smooth.
By differential geometry, if we approximate the surface $\mathcal{S}_{\bm{p}}^{*}$ locally at a point $\bm{p}$ with a polynomial $\mathcal{S}_{\bm{p}}(m)$ of degree $m$, we have the following error bound \cite{MLS_Smooth_Denoise}
\begin{equation}
    \|\mathcal{S}_{\bm{p}}^{*} - \mathcal{S}_{\bm{p}}(m)\| \leq C(\|{\mathcal{S}_{\bm{p}}^{*}}^{(m+1)}\|)\cdot \omega_{\bm{p}}^{m+1},
    \nonumber
\end{equation}
where $\omega_{\bm{p}}$ denotes the width of local 2D domain centered at $\bm{p}$ for $\mathcal{S}_{\bm{p}}(m)$ and $C(\|{\mathcal{S}_{\bm{p}}^{*}}^{(m+1)}\|)$ is a constant depending on the $(m+1)^{th}$ derivatives of the local surface.
Given a surface $\mathcal{S}^{*}$, we can define $\{\bm{p}_i\}$ and $\{\omega_{\bm{p}_i}\}$ to cover the whole surface by $\{\mathcal{S}_{\bm{p}_i}(m)\}$ with no holes.
And in this way, the union of $\{\mathcal{S}_{\bm{p}_i}(m)\}$ approximates $\mathcal{S}^{*}$ with the following error bound
\begin{equation}\label{eq:int_errorgap}
    \int_{\mathcal{S}^{*}} \|\mathcal{S}_{\bm{p}_i}^{*} - \mathcal{S}_{\bm{p}_i}(m)\|  \leq \int_{\mathcal{S}^{*}} C(\|{\mathcal{S}_{\bm{p}_i}^{*}}^{(m+1)}\|)\cdot \omega_{\bm{p}_i}^{m+1}.
\end{equation}
Considering that the surface $\mathcal{S}^{*}$ is represented as a triangular mesh $\mathcal{G}_{\mathcal{S}^{*}}$, a discrete version of \cref{eq:int_errorgap} can be written as
\begin{equation}\label{eq:dis_errorgap}
    \sum_{i=1}^{n} \|\mathcal{S}_{\bm{v}_i}^{*} - \mathcal{S}_{\bm{v}_i}(m)\|  \leq \sum_{i=1}^{n} C(\|{\mathcal{S}_{\bm{v}_i}^{*}}^{(m+1)}\|)\cdot \omega_{\bm{v}_i}^{m+1},
\end{equation}
where ${\{\bm{v}_{i}\}}_{i=1}^{n}$ are the vertices of $\mathcal{G}_{\mathcal{S}^{*}}$.
It is practically reasonable to set $m=1$ to have a piecewise linear surface approximation, since our visual system is insensitive to surface smoothness beyond second order \cite{lin2020cad}.
Considering a fixed number $n$ of vertices for all meshes, which is the case in our benchmark,; it is clear that comparing algebraic complexities of different surfaces is to compare their respective constants $C(\|{\mathcal{S}_{\bm{v}}^{*}}^{(2)}\|)$, which depend on the curvatures of local surfaces. We can compute the curvature as the integral of local, squared normal curvatures
\begin{equation}
    \begin{split}
        &\qquad\frac{1}{\pi}\int_0^{\pi}{\kappa_{n}^2(\bm{v}, \theta)d\theta} \\
        &=\frac{1}{\pi}\int_0^{\pi} (\kappa_{1}(\bm{v})\cos^2(\theta) + \kappa_{2}(\bm{v})\sin^2(\theta))^2d\theta \\
        &=\frac{3}{8}\kappa_{1}^2(\bm{v}) + \frac{2}{8}\kappa_{1}(\bm{v})\kappa_{2}(\bm{v}) + \frac{3}{8}\kappa_{2}^2(\bm{v}) \\
        &=\frac{3}{2}(\frac{\kappa_{1}(\bm{v})+\kappa_{2}(\bm{v})}{2})^2 - \frac{1}{2}\kappa_{1}(\bm{v})\kappa_{2}(\bm{v}) \\
        &=\frac{3}{2} \varkappa^2(\bm{v})-\frac{1}{2} \varrho(\bm{v}),
    \end{split}
\end{equation}
where $\kappa_{n}(\bm{v}, \theta)$ denotes the normal curvature at $\bm{v}$ in the direction of $\theta$ using Euler's famous formula \cite{3dshapeanalysis}; $\kappa_{1}(\bm{v})$ and $\kappa_{2}(\bm{v})$ denote the principal curvatures; $\varkappa(\bm{v})$ and $\varrho(\bm{v})$ denote the mean and Gaussian curvatures at $\bm{v}$, respectively.
For $\mathcal{S}^{*}$, we take the expectation of the integral of local, squared normal curvatures over points on the surface as its algebraic complexity
\begin{equation}\label{eq:final_errorgap}
    \mathbb{E}_{{\{\bm{v}_{i}\}}_{i=1}^{n} \in \mathcal{G}_{\mathcal{S}^{*}}}[\frac{3}{2} \varkappa^2(\bm{v})-\frac{1}{2} \varrho(\bm{v})] .
\end{equation}
Note that a higher value of \cref{eq:final_errorgap} indicates a more complicated surface.
We finally divide all the object surfaces in our benchmark into three groups respectively of 972, 486, and 162 instances (at the ratio of around 6 : 3 : 1 for low-, middle-, and high-complexity groups) according to their computed surface complexities. 

\section{Details of Evaluation Metrics}\label{sec:Evaluation}
We use three popular metrics of Chamfer Distance (CD) \cite{CD}, F-score \cite{largesaclebenchmark}, and Normal Consistency Score (NCS) \cite{occupancy}, and also a newly proposed neural metric, termed Neural Feature Similarity (NFS), to quantitatively compare surface meshes obtained from different methods. Details are presented as follows.

\subsection{Popular Evaluation Metrics}
\noindent\textbf{Chamfer Distance -- }
Chamfer Distance (CD) is a metric between two point sets; in our case, it is defined over two point sets $\mathcal{P}=\{\bm{p}_i\}_{i=1}^{n_\mathcal{P}}$ and $\mathcal{Q}=\{\bm{q}_i\}_{i=1}^{n_\mathcal{Q}}$ sampled from the reconstructed surface $\mathcal{G}_{\mathcal{S}}$ and ground-truth surface $\mathcal{G}_{\mathcal{S}^{*}}$, respectively.
It is a symmetric metric computed as
\begin{eqnarray}
\frac{1}{2n_{\mathcal{P}}} \sum_{\bm{p} \in \mathcal{P}} \min_{\bm{q} \in \mathcal{Q}} \|\bm{p}-\bm{q}\|_2 + \frac{1}{2n_{\mathcal{Q}}} \sum_{\bm{q} \in \mathcal{Q}} \min_{\bm{p} \in \mathcal{P}} \|\bm{p}-\bm{q}\|_2 .
    \notag
\end{eqnarray}
We set $n_{\mathcal{P}}=n_{\mathcal{Q}}=200$k and $n_{\mathcal{P}}=n_{\mathcal{Q}}=1500$k for evaluation on the synthetic data of object surfaces and scene surfaces, respectively.

\vspace{0.1cm}
\noindent\textbf{F-score -- }
Similar to CD, F-score is also a metric between two point sets. Different from CD, F-score acts more like an $l_0$-norm measurement, since it adopts an indication function $\bm{1}_{\text{condition}}[\cdot]$ (shortened as $\bm{1}_{c}[\cdot]$) to measure the precision and recall between two point sets, and it also takes the harmonic mean between the precision and recall, which means that it will be dominated by the minimum of either precision or recall.
Therefore, though it still specializes in measuring the overall similarity between two shapes, it is more sensitive to extreme cases in which either precision or recall is bad.
Specifically, let $\Phi(\bm{p}; \mathcal{Q}) = \min_{\bm{q} \in \mathcal{Q}} \|\bm{p}-\bm{q}\|_2$ and $\Phi(\bm{q}; \mathcal{P}) = \min_{\bm{p} \in \mathcal{P}} \|\bm{p}-\bm{q}\|_2$; the precision can be defined as $D(\mathcal{P}; \mathcal{Q}, \tau) = \frac{1}{n_{\mathcal{P}}} \sum_{\bm{p}\in\mathcal{P}}{\bm{1}_{c}[\Phi(\bm{p}; \mathcal{Q})<\tau]}$, and the recall can be defined as $D(\mathcal{Q}; \mathcal{P}, \tau) = \frac{1}{n_{\mathcal{Q}}} \sum_{\bm{q}\in\mathcal{Q}}{\bm{1}_{c}[\Phi(\bm{q}; \mathcal{P})<\tau]}$.
The overall F-score is computed as
\begin{equation}
\frac{2D(\mathcal{P}; \mathcal{Q}, \tau)D(\mathcal{Q}; \mathcal{P}, \tau)}{D(\mathcal{P}; \mathcal{Q}, \tau)+D(\mathcal{Q}; \mathcal{P}, \tau)} \times 100,
    \notag
\end{equation}
where $\tau$ is a hyper-parameter controlling the sensitivity of F-score.
We set $\tau=0.005$ with $n_{\mathcal{P}}=n_{\mathcal{Q}}=200$k and $\tau=0.03$ with $n_{\mathcal{P}}=n_{\mathcal{Q}}=1500$k for evaluation on the synthetic data of object surfaces and scene surfaces, respectively, and set $n_{\mathcal{P}}=n_{\mathcal{Q}}$ to be the number of points in the obtained point clouds by high-precision, real scanning (i.e., the ground-truth surfaces of our real-scanned data), with $\tau=0.5$.

\noindent\textbf{Normal Consistency Score -- }
Normal Consistency Score (NCS) is a metric measuring the consistency between two vector fields. As such, it measures the difference of higher order information between different surfaces, and is more sensitive to subtle shape differences. It can be computed as
\begin{equation}
\begin{split}
\frac{1}{2} \left(  \frac{1}{n_{\mathcal{P}}} \sum_{\bm{p}\in\mathcal{P}} \left| \left<\bm{n}_{\mathcal{T}_{\bm{p}}}, \bm{n}_{\mathcal{T}_{\Phi(\bm{p}; \mathcal{Q})}}\right>\right| + \frac{1}{n_{\mathcal{Q}}} \sum_{\bm{q}\in\mathcal{Q}} \left|\left<\bm{n}_{\mathcal{T}_{\bm{q}}}, \bm{n}_{\mathcal{T}_{\Phi(\bm{q}; \mathcal{P})}}\right>\right| \right),
    \end{split}
    \notag
\end{equation}
where $\bm{n}_{\mathcal{T}_{\bm{p}}}$ denotes the normal estimated from the continuous triangular facet $\mathcal{T}_{\bm{p}}$ at point $\bm{p}$.
We use $\bm{n}_{\mathcal{T}_{\bm{p}}}$ instead of the $\bm{n}_{\bm{p}}$ estimated from the discrete point cloud to reduce the error brought by discrete point sampling.
We set $n_{\mathcal{P}}=n_{\mathcal{Q}}=200$k and $n_{\mathcal{P}}=n_{\mathcal{Q}}=1500$k for evaluation on the synthetic data of object surfaces and scene surfaces respectively, and set $n_{\mathcal{P}}=n_{\mathcal{Q}}$ to be the number of points in the obtained point clouds by high-precision, real scanning (i.e., the ground-truth surfaces of our real-scanned data).

\subsection{The Proposed Neural Evaluation Metric}
\noindent\textbf{Neural Feature Similarity -- }
The aforementioned three metrics are popularly used in existing surface reconstruction evaluation; however, they may fail to capture the shape differences that are more consistent with our human perception.
Inspired by \cite{deepfeatureperceptualmetric}, we propose to compare the similarity between two shapes in the deep feature space, which depends more on the high-level semantic information and the resulting comparison would be more consistent with human perception. More specifically, given the deep features for different surfaces, we use cosine similarity to compute the proposed Neural Feature Similarity (NFS)
\begin{equation}
\frac{\bm{g}_{\bm{\theta}}(\mathcal{P}) \cdot \bm{g}_{\bm{\theta}}(\mathcal{Q})}{\|\bm{g}_{\bm{\theta}}(\mathcal{P})\|_{2}\|\bm{g}_{\bm{\theta}}(\mathcal{Q})\|_{2}},
    \notag
\end{equation}
where the feature-extracting network $\bm{g}_{\bm{\theta}}$ takes in a point cloud $\mathcal{P}$ and outputs a feature vector.

We train the network $\bm{g}_{\bm{\theta}}$ in a self-supervised manner, such that features of point clouds sampled from the same surface are as invariant as possible, and features of point clouds sampled from different surfaces are different. We use the following self-supervision training objective
\begin{equation}
\begin{split}
\sum_{i,i'} \left| \frac{\bm{g}_{\bm{\theta}}(\mathcal{P}_i) \cdot \bm{g}_{\bm{\theta}}(\mathcal{P}_{i'})}{\|\bm{g}_{\bm{\theta}}(\mathcal{P}_{i})\|_{2}\|\bm{g}_{\bm{\theta}}(\mathcal{P}_{i'})\|_{2}}  - 1 \right|  + \sum_{i,j} \left| \frac{\bm{g}_{\bm{\theta}}(\mathcal{P}_i) \cdot \bm{g}_{\bm{\theta}}(\mathcal{P}_j)}{\|\bm{g}_{\bm{\theta}}(\mathcal{P}_i)\|_{2}\|\bm{g}_{\bm{\theta}}(\mathcal{P}_j)\|_{2}} \right| ,
\end{split}
\end{equation}
where the index pair $\{i, i'\}$ means that the point clouds $\{\mathcal{P}_i, \mathcal{P}_{i'}\}$ are sampled from the same surface, and $\{i, j\}$ for those from different surfaces.
To further improve the accuracy, we take the aligned local surface patches as the input, instead of the whole surface, and take the average of the differences overs all the local surfaces to be the final result. We adopt a $6$-layer, MLP-based auto-decoder (with the activation function of Leaky ReLU, and $256$ channels per layer) as $\bm{g}_{\bm{\theta}}(\cdot)$, and each local surface patch is represented as a $256$ dimensional latent vector.
We use Adam \cite{Adam} with the learning rate initially set as $0.0001$ and decreased by half every $200$ epochs; we train for a total of $1000$ epochs.

\section{More Results for the Synthetic Data of Object Surfaces} \label{sec:all_exp_syn_obj}
In this section, we present the overall results on the testing synthetic data of object surfaces as in \cref{tab:syn_obj_all}.

\begin{table*}[hptb]
\centering
\caption{
Quantitative results on the testing synthetic data of object surfaces.
For those challenges with varying levels of severity, we report results in the format of ``$\cdot$ / $\cdot$ / $\cdot$'', which, from left to right, are results for low-, middle-, and high-level severities of each challenge.
Results of the \textbf{best} and \uline{\textbf{second best}} methods are highlighted in each column.
}
\vspace{-0.1in}
\label{tab:syn_obj_all}
\resizebox{\textwidth}{!}{%
\begin{tabular}{l|ccr@{ /}r@{ / }cr@{ /}r@{ / }cr@{ /}r@{ / }cr@{ /}r@{ / }c}
\hline
\multirow{2}{*}{Algorithms} & \multicolumn{14}{c}{CD $ (\times10^{-4}) \downarrow$}  \\\cline{2-15}
~ & \makecell{Perfect\\scanning} & \makecell{Non-uniform\\distrubution} & \multicolumn{3}{c}{\makecell{Point-wise noise\\\emph{low- /middle- /high-level}}} & \multicolumn{3}{c}{\makecell{Point outliers\\\emph{low- /middle- /high-level}}} & \multicolumn{3}{c}{\makecell{Missing points\\\emph{low- /middle- /high-level}}} & \multicolumn{3}{c}{\makecell{Misalignment\\\emph{low- /middle- /high-level}}} \\\hline
GD \cite{greedydelaunay}                                                & \textbf{14.24}               & \textbf{14.87}               & 16.52                        & 18.20                        & 29.73                        & 103.77                       & 123.19                       & 155.98                       & \uline{\textbf{44.33}}       & 64.15                        & \uline{\textbf{110.74}}       & 16.68                        & 21.15                        & 39.49                   \\
BPA \cite{BPA}                                                          & 14.89                        & 17.13                        & 16.59                        & 18.58                        & 29.30                        & \uline{\textbf{14.41}}       & \textbf{14.66}               & \textbf{16.15}               & 50.99                        & \uline{\textbf{62.55}}       & 121.59                        & 16.61                        & 20.88                        & 42.46                   \\
\hline
\end{tabular}%
}
\end{table*}

\section{More Results for Studying the Importance of Orientations of Surface Normals} \label{sec:more_exp_normals}

\begin{table}[hptb]
\centering
\caption{Comparison on the testing synthetic data of scene surfaces among methods without using surface normals ($\times$), methods using surfaces normals ($\checkmark$), and methods using surface normals only during learning ($*$).
    Results are in the format of ``$\cdot$ / $\cdot$'', where the left one is obtained assuming the availability of ground-truth camera poses, and the right one is obtained without knowing the camera poses.
    The \textbf{best} and \uline{\textbf{second best}} methods are highlighted in each column.
}
\vspace{-0.1in}
\label{tab:oriented_normal_comparing_scene}
\resizebox{\linewidth}{!}{
\begin{tabular}{l|c|c|c|c|c} \hline
\multirow{1}{*}{Algorithms} & \multirow{1}{*}{Normals} & \multicolumn{1}{c|}{CD $(\times10^{-3}) \downarrow$} & \multicolumn{1}{c|}{F-score $(\%) \uparrow$} & \multicolumn{1}{c|}{NCS $(\times10^{-2}) \uparrow$} & \multicolumn{1}{c}{NFS $(\times10^{-2}) \uparrow$} \\\hline
GD \cite{greedydelaunay}                                                &  \multirow{3}{*}{$\times$}      & 33.85                     / {\ \ \ \ \ \ \ }  & 75.95                   / {\ \ \ \ \ \ \ }  & 62.08                   / {\ \ \ \ \ \ \ } & 41.92                    / {\ \ \ \ \ \ \ }       \\
SALD \cite{sald}                                                        &                                 & 32.43                     / {\ \ \ \ \ \ \ }  & 79.03                   / {\ \ \ \ \ \ \ }  & \textbf{91.58}          / {\ \ \ \ \ \ \ } & 52.79                    / {\ \ \ \ \ \ \ }       \\
DSE \cite{learningdelaunaysurface}                                      &                                 & 32.97                     / {\ \ \ \ \ \ \ }  & 77.53                   / {\ \ \ \ \ \ \ }  & 57.99                   / {\ \ \ \ \ \ \ } & 41.56                    / {\ \ \ \ \ \ \ }       \\ \hline
BPA \cite{BPA}                                                          &  \multirow{7}{*}{$\checkmark$}  & 45.82                     / 57.08             & 53.46                   / 47.57             & 58.25                   / 44.04            & 44.19                    / 32.01                  \\
SPSR \cite{kazhdan2013screened}                                         &                                 & {\ }\textbf{30.47}        / 142.65            & \textbf{83.22}          / 68.06             & 83.74                   / 61.31            & \uline{\textbf{63.03}}   / 41.69                  \\
RIMLS \cite{RIMLS}                                                      &                                 & 41.45                     / 55.79             & 74.56                   / 65.14             & 69.93                   / 52.16            & 38.13                    / 26.55                  \\
IGR \cite{gropp2020implicitgeometricregularizationforlearningshape}     &                                 & {\ }\uline{\textbf{31.41}}/ 352.04            & \uline{\textbf{81.63}}  / 63.37             & \uline{\textbf{91.26}}  / 67.33            & \textbf{67.58}           / 42.20                  \\
OccNet \cite{occupancy}                                                 &                                 & {\ }93.12                 / 120.85            & 37.75                   / 32.42             & 85.98                   / 64.12            & 50.34                    / 36.75                  \\
DeepSDF \cite{deepsdf}                                                  &                                 &  -                        / -                 &       -                 / -                 &       -                 / -                &        -                 / -                      \\
LIG \cite{LIG}                                                          &                                 & 41.40                     / 81.22             & 78.03                   / 64.12             & 88.12                   / 64.53            & 59.97                    / 39.92                  \\
ParseNet \cite{parsenet}                                                &                                 &  -                        / -                 &       -                 / -                 &       -                 / -                &        -                 / -                      \\ \hline
Points2Surf \cite{points2surf}                                          &  \multirow{2}{*}{$*$}           & 36.24                     / {\ \ \ \ \ \ \ }  & 76.14                   / {\ \ \ \ \ \ \ }  & 83.60                   / {\ \ \ \ \ \ \ } & 61.82                    / {\ \ \ \ \ \ \ }       \\
IMLSNet \cite{liu2021DeepIMLS}                                          &                                 & 35.52                     / {\ \ \ \ \ \ \ }  & 78.05                   / {\ \ \ \ \ \ \ }  & 87.17                   / {\ \ \ \ \ \ \ } & 61.98                    / {\ \ \ \ \ \ \ }       \\ \hline
\end{tabular}%
}
\end{table}

\begin{table}[hptb]
\centering
\caption{
    Comparison on the testing real-scanned data among methods without using surface normals ($\times$), methods using surfaces normals ($\checkmark$), and methods using surface normals only during learning ($*$) .
    Results are in the format of ``$\cdot$ / $\cdot$'', where the left one is obtained assuming the availability of ground-truth camera poses, and the right one is obtained without knowing the camera poses.
    The \textbf{best} and \uline{\textbf{second best}} methods are highlighted in each column.
}
\vspace{-0.1in}
\label{tab:oriented_normal_comparing_real}
\resizebox{\linewidth}{!}{
\begin{tabular}{l|c|c|c|c|c} \hline
\multirow{1}{*}{Algorithms} & \multirow{1}{*}{Normals} & \multicolumn{1}{c|}{CD $(\times10^{-2}) \downarrow$} & \multicolumn{1}{c|}{F-score $(\%) \uparrow$} & \multicolumn{1}{c|}{NCS $(\times10^{-2}) \uparrow$} & \multicolumn{1}{c}{NFS $(\times10^{-2}) \uparrow$} \\\hline
GD \cite{greedydelaunay}                                                &  \multirow{3}{*}{$\times$}      & 31.72                    / {\ \ \ \ \ \ \ }  & 87.51                   / {\ \ \ \ \ \ \ }  & 88.86                    / {\ \ \ \ \ \ \ } & 82.20                    / {\ \ \ \ \ \ \ }       \\
SALD \cite{sald}                                                        &                                 & \uline{\textbf{31.13}}   / {\ \ \ \ \ \ \ }  & \uline{\textbf{87.72}}  / {\ \ \ \ \ \ \ }  & 94.68                    / {\ \ \ \ \ \ \ } & 86.86                    / {\ \ \ \ \ \ \ }       \\
DSE \cite{learningdelaunaysurface}                                      &                                 & 32.16                    / {\ \ \ \ \ \ \ }  & 86.88                   / {\ \ \ \ \ \ \ }  & 87.20                    / {\ \ \ \ \ \ \ } & 76.81                    / {\ \ \ \ \ \ \ }       \\ \hline
BPA \cite{BPA}                                                          &  \multirow{7}{*}{$\checkmark$}  & 40.37                    / 40.94             & 80.95                   / 80.49             & 87.56                    / 87.17            & 68.69                    / 66.80                  \\
SPSR \cite{kazhdan2013screened}                                         &                                 & {\ } \textbf{31.05}      / 126.89            & \textbf{87.74}          / 80.67             & \uline{\textbf{94.94}}   / 92.64            & \textbf{89.38}           / 79.17                  \\
RIMLS \cite{RIMLS}                                                      &                                 & 32.80                    / 36.39             & 87.05                   / 85.50             & 91.97                    / 90.84            & 85.19                    / 79.64                  \\
IGR \cite{gropp2020implicitgeometricregularizationforlearningshape}     &                                 & {\ } 32.70               / 467.06            & 87.18                   / 76.68             & \textbf{95.99}           / 93.79            & \uline{\textbf{89.10}}   / 75.04                  \\
OccNet \cite{occupancy}                                                 &                                 & 232.71                   / 252.81            & 17.11                   / 16.57             & 80.96                    / 80.92            & 39.70                    / 38.52                  \\
DeepSDF \cite{deepsdf}                                                  &                                 & 263.92                   / 552.25            & 19.83                   / 15.16             & 77.95                    / 77.66            & 40.95                    / 35.21                  \\
LIG \cite{LIG}                                                          &                                 & 48.75                    / 79.89             & 83.76                   / 76.38             & 92.57                    / 89.46            & 81.48                    / 72.21                  \\
ParseNet \cite{parsenet}                                                &                                 & 149.96                   / 217.29            & 38.92                   / 31.06             & 81.51                    / 77.30            & 45.67                    / 38.48                  \\ \hline
Points2Surf \cite{points2surf}                                          &  \multirow{2}{*}{$*$}           & 48.93                    / {\ \ \ \ \ \ \ }  & 80.89                   / {\ \ \ \ \ \ \ }  & 89.52                    / {\ \ \ \ \ \ \ } & 81.83                    / {\ \ \ \ \ \ \ }       \\
IMLSNet \cite{liu2021DeepIMLS}                                          &                                 & 38.46                    / {\ \ \ \ \ \ \ }  & 82.44                   / {\ \ \ \ \ \ \ }  & 93.31                    / {\ \ \ \ \ \ \ } & 85.30                    / {\ \ \ \ \ \ \ }       \\ \hline
\end{tabular}%
}
\end{table}

In this section, we show more results of scene surfaces and real-scanned data for studying the importance of orientations of surface normals. \cref{tab:oriented_normal_comparing_scene} shows qualitative results of scene surfaces that compare methods using surfaces normals, those without using surfaces normals, and those using surface normals only during learning. Those on real-scanned data are shown in \cref{tab:oriented_normal_comparing_real}.

\section{Experimental Results Without Data Pre-processing} \label{sec:more_exp_wo_pre}

\begin{table*}[hptb]
\centering
\caption{
Quantitative results on the testing synthetic data of object surfaces when the input point clouds are NOT pre-processed (cf. \cref{SecPreprocessing} for details). For those challenges with varying levels of severity, we report results in the format of ``$\cdot$ / $\cdot$ / $\cdot$'', which, from left to right, are the results for low-, middle-, and high-level severities of each challenge.
Results of the \textbf{best} and \uline{\textbf{second best}} methods are highlighted in each column.
}
\vspace{-0.1in}
\label{tab:syn_obj_wo_preprocess}
\resizebox{\textwidth}{!}{%
\begin{tabular}{l|ccr@{ /}r@{ / }cr@{ /}r@{ / }cr@{ /}r@{ / }cr@{ /}r@{ / }c}
\hline
\multirow{2}{*}{Algorithms} & \multicolumn{14}{c}{CD $ (\times10^{-4}) \downarrow$}  \\\cline{2-15}
~ & \makecell{Perfect\\scanning} & \makecell{Non-uniform\\distrubution} & \multicolumn{3}{c}{\makecell{Point-wise noise\\\emph{low- /middle- /high-level}}} & \multicolumn{3}{c}{\makecell{Point outliers\\\emph{low- /middle- /high-level}}} & \multicolumn{3}{c}{\makecell{Missing points\\\emph{low- /middle- /high-level}}} & \multicolumn{3}{c}{\makecell{Misalignment\\\emph{low- /middle- /high-level}}} \\\hline
GD \cite{greedydelaunay}                                                & \textbf{14.35}               & \textbf{14.87}               & 21.39                        & 34.53                        & 46.66                        & 210.20                       & 581.89                       & 233.24                       & \uline{\textbf{33.66}}       & \uline{\textbf{65.23}}       & \uline{\textbf{97.21}}       & 21.65                        & 40.52                        & 56.94                     \\
BPA \cite{BPA}                                                          & 18.60                        & 32.68                        & 23.76                        & 49.42                        & 55.63                        & \uline{\textbf{17.81}}       & \uline{\textbf{18.94}}       & \uline{\textbf{25.73}}       & 83.17                        & 78.99                        & 126.46                       & 56.64                        & 36.90                        & 68.30                     \\
\hline
\end{tabular}%
}
\end{table*}

\begin{table}[hptb]
\centering
\caption{
Quantitative results on the testing synthetic data of scene surfaces when the input point clouds are Not preprocessed (cf. \cref{SecPreprocessing} for details).
Results of the \textbf{best} and \uline{\textbf{second best}} methods are highlighted in each column.
``-'' indicates that the method cannot produce reasonable results due to their limited generalization.
}
\vspace{-0.1in}
\label{tab:syn_scene_wo_preprocess}
\resizebox{\linewidth}{!}{%
\begin{tabular}{c|l|c|c|c|c}
    \hline
    Prior & Algorithm & \makecell{CD\\$(\times10^{-3}) \downarrow$} & \makecell{F-score\\$(\%) \uparrow$} & \makecell{NCS\\$(\times10^{-2}) \uparrow$} & \makecell{NFS\\$(\times10^{-2}) \uparrow$} \\ \hline
    \multirow{2}{*}{\makecell{Triangulation-\\based}}   & GD \cite{greedydelaunay}                                              & 34.50                        & 71.33                        & 56.78                        & 38.77                    \\
                                                        & BPA \cite{BPA}                                                        & 49.01                        & 55.45                        & 53.52                        & 39.87                    \\\cline{1-2}
    \multirow{2}{*}{\makecell{Smoothness}}              & SPSR \cite{kazhdan2013screened}                                       & \textbf{30.63}               & \textbf{83.03}               & 83.50                        & \uline{\textbf{62.46}}   \\
                                                        & RIMLS \cite{RIMLS}                                                    & 42.53                        & 69.74                        & 64.83                        & 33.82                    \\\cline{1-2}
    \multirow{2}{*}{\makecell{Modeling}}                & SALD \cite{sald}                                                      & 32.87                        & 75.15                        & \textbf{91.29}               & 51.36                    \\
                                                        & IGR \cite{gropp2020implicitgeometricregularizationforlearningshape}   & \uline{\textbf{32.82}}       & \uline{\textbf{81.37}}       & \uline{\textbf{90.32}}       & \textbf{66.71}           \\
\hline
\end{tabular}%
}
\end{table}

We have reported and discussed the results in \cref{SecPreprocessing} that are obtained \textit{with} the pre-processing pipeline.
In this section, we report results \textit{without} the pre-processing pipeline.
The results with and without pre-processing are of similar comparative qualities, which are given in \cref{tab:syn_obj_wo_preprocess} and \cref{tab:syn_scene_wo_preprocess} respectively for synthetic data of object surfaces and synthetic data of scene surfaces.
Note that the results of real-scanned data are not involved in this section as the scanners we used automatically pre-process the point clouds with their inbuilt pre-processing methods.

\section{More Results for the Real-scanned Data} \label{sec:more_exp_real}

\begin{figure*}[htbp]
    \begin{minipage}{1.0\linewidth}
        \centering
        \includegraphics[width=0.99\textwidth]{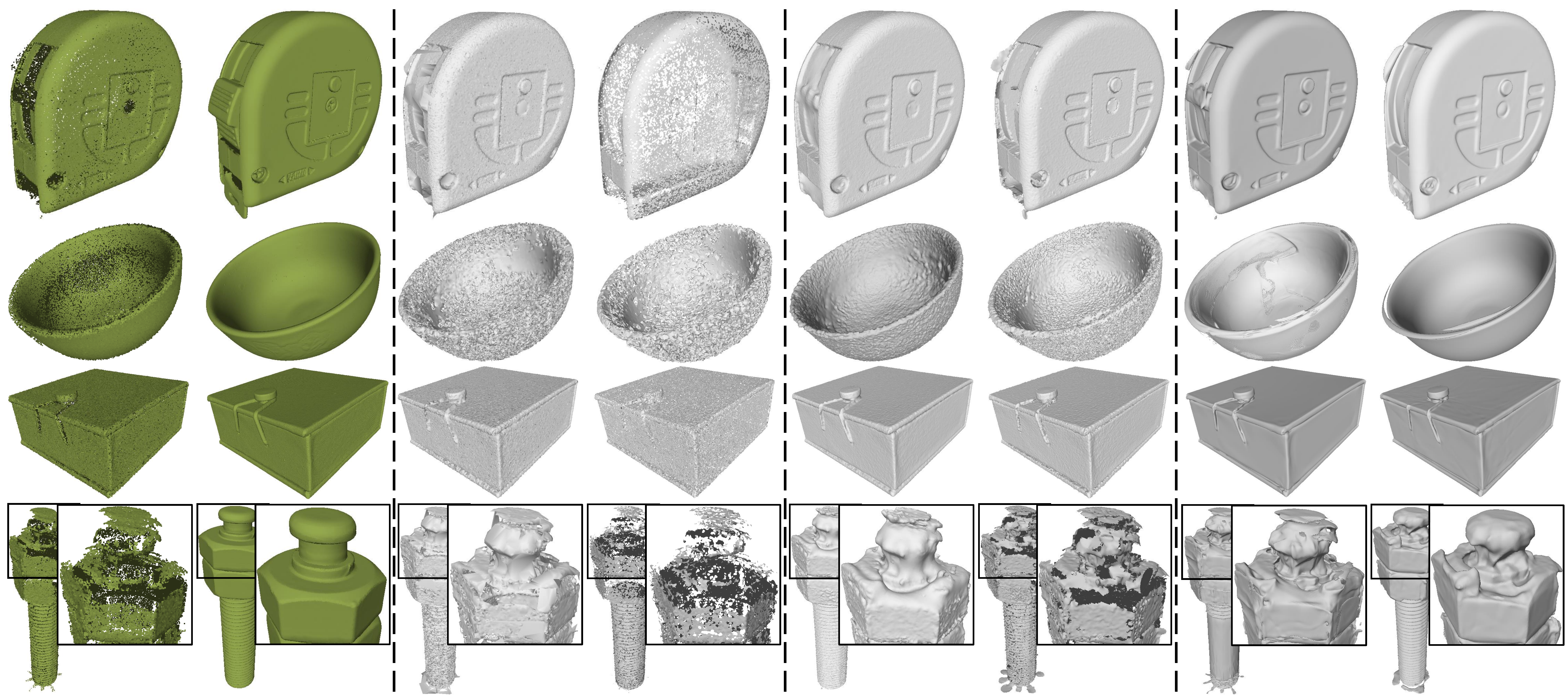}
        \vspace{0.07in}
        \scalebox{0.75}{ \begin{tabular}{p{70pt}<{\centering}p{70pt}<{\centering}p{75pt}<{\centering}p{75pt}<{\centering}p{70pt}<{\centering}p{70pt}<{\centering}p{70pt}<{\centering}p{70pt}<{\centering}} \small Input PC & \    \small GT & \   \small GD \cite{greedydelaunay} & \     \small BPA \cite{BPA} &  \       \small SPSR \cite{kazhdan2013screened} &   \       \small RIMLS \cite{RIMLS}  &  \    \small SALD \cite{sald} &  \       \small IGR \cite{gropp2020implicitgeometricregularizationforlearningshape}    \\          \end{tabular}}
    \end{minipage}
    \vspace{-0.02in}
    \begin{minipage}{1.0\linewidth}
        \centering
        \includegraphics[width=0.99\textwidth]{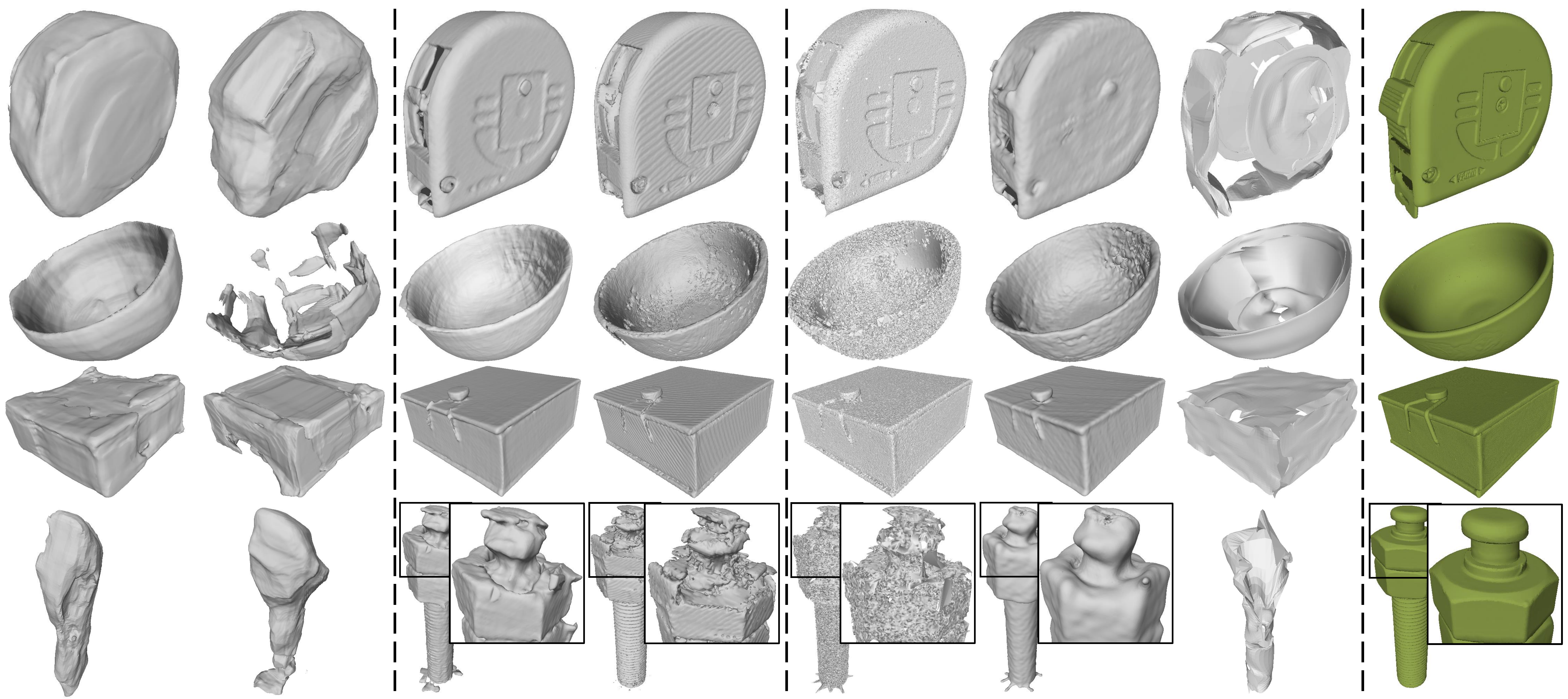}
        \vspace{0.07in}
        \scalebox{0.75 }{\begin{tabular}{p{70pt}<{\centering}p{70pt}<{\centering}p{75pt}<{\centering}p{75pt}<{\raggedright}p{70pt}<{\centering}p{70pt}p{70pt}<{\centering}p{70pt}<{\centering}}  \small OccNet \cite{occupancy}  &  \       \small DeepSDF \cite{deepsdf}  &  \    \small LIG \cite{LIG} & \       \small Points2Surf \cite{points2surf} &  \         \small DSE \cite{learningdelaunaysurface} & \ \small IMLSNet \cite{liu2021DeepIMLS}  &  \       \small ParseNet \cite{parsenet} &  \       \small GT \\            \end{tabular}}
    \end{minipage}
    \caption{Additional qualitative results on the real-scanned data.}
    \label{fig:exp_real}
\end{figure*}

\cref{fig:exp_real} shows more qualitative results on the real-scanned data. The results suggest that quality of a scanned point cloud depends heavily on the surface material. Other observations are consistent with those in \cref{sec:experiments}.

\end{document}